\documentclass[journal, twocolumn]{IEEEtran}

\usepackage[pagebackref=false,breaklinks=false,letterpaper=false,colorlinks,bookmarks=false]{hyperref}
\usepackage{makecell}
\usepackage{overpic}
\usepackage{multirow}
\usepackage{bigstrut}
\usepackage{graphicx}
\usepackage{threeparttable}

\def\ie{\emph{i.e.}}
\def\eg{\emph{e.g.}}
\def\etc{\emph{etc}}
\def\etal{\textit{et al.}}

\newcommand{\rev}[1]{\textcolor{red}{#1}}
\newcommand{\blu}[1]{\textcolor{blue}{#1}}
\newcommand{\gre}[1]{\textcolor{green}{#1}}
\newcommand{\secref}[1]{$\S$ \ref{#1}}

\newcommand{\rotate}[1]{ \rotatebox{90}{#1}}

\usepackage{amsmath,graphicx}
\usepackage{color}

\usepackage{lineno,hyperref}
\usepackage{amssymb}
\usepackage{amsmath,bm}
\usepackage{booktabs}
\usepackage{color}
\usepackage{amsmath,graphicx}
\usepackage{color}
\usepackage{booktabs}
\usepackage{color}
\usepackage{multirow}
\usepackage{makecell}

\usepackage[ruled,linesnumbered]{algorithm2e}
\usepackage{amsmath,graphicx}
\usepackage{color}

\usepackage{lineno,hyperref}
\usepackage{amssymb}
\usepackage{amsmath,bm}
\usepackage{booktabs}
\usepackage{color}

\usepackage{multirow}
\usepackage{makecell}
\usepackage{bm}

\usepackage{cite}

\ifCLASSINFOpdf
\else
\fi

\begin{document}
	
	\title{RGB-D Salient Object Detection: A Survey}
	
	\author{Tao Zhou, Deng-Ping Fan, Ming-Ming Cheng, Jianbing Shen, and Ling Shao
		\thanks{Corresponding author: Ding-Ping Fan (dengpingfan@mail.nankai.edu.cn).}
		\thanks{T. Zhou, D.-P. Fan, J. Shen, and L. Shao are with Inception Institute of Artificial Intelligence, Abu Dhabi, UAE.  
		}
			\thanks{M.-M. Cheng is with CS, Nankai University, Tianjin 300350, China.}}

	\maketitle

	\begin{abstract}
		
Salient object detection (SOD), which simulates the human visual perception system to locate the most attractive object(s) in a scene, has been widely applied to various computer vision tasks. Now, with the advent of depth sensors, depth maps with affluent spatial information that can be beneficial in boosting the performance of SOD can easily be captured. Although various RGB-D based SOD models with promising performance have been proposed over the past several years, an in-depth understanding of these models and the challenges in this field remains lacking. In this paper, we provide a comprehensive survey of RGB-D based SOD models from various perspectives, and review related benchmark datasets in detail. Further, considering the fact that light fields can also provide depth maps, we review SOD models and popular benchmark datasets from this domain as well. Moreover, to investigate the SOD ability of existing models, we carry out a comprehensive evaluation and conduct an attribute-based evaluation of several representative RGB-D based SOD models. Finally, we discuss several challenges and open directions of RGB-D based SOD for future research. All collected models, benchmark datasets, source code links, datasets constructed for attribute-based evaluation, and codes for evaluation have been made publicly available at \href{https://github.com/taozh2017/RGBD-SODsurvey}{https://github.com/taozh2017/RGBD-SODsurvey}.

	\end{abstract}
	

\section{Introduction}
\label{sec:introduction}

\begin{figure}[t]
    \centering
    \includegraphics[width=0.99\linewidth]{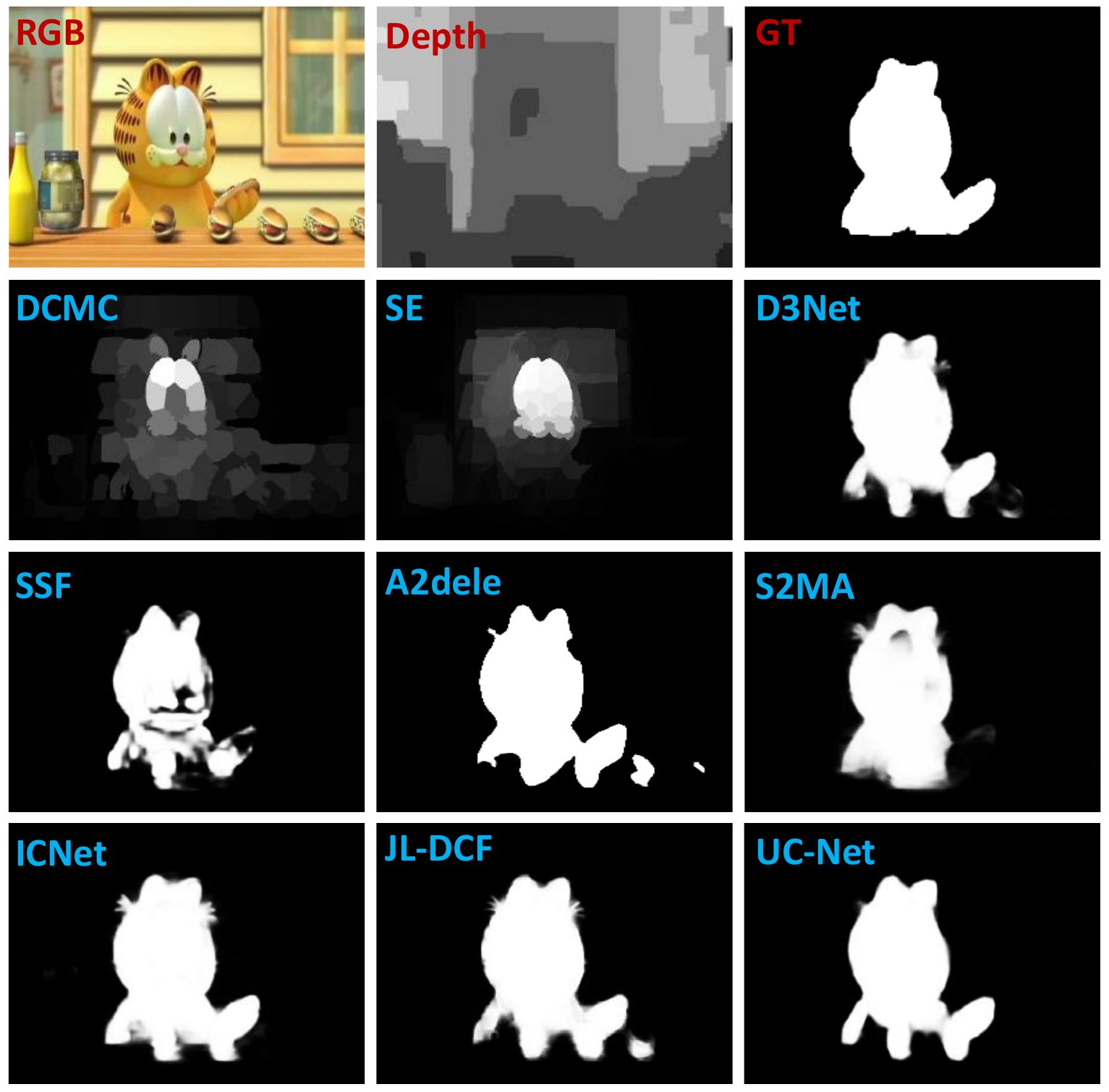} \vspace {-0.5cm}
    \caption{RGB-D based salient object prediction on a sample image using two classic non-deep models (\ie, DCMC \cite{cong2016saliency} and SE \cite{guo2016salient}) and seven state-of-the-art deep models (\ie, D$^3$Net \cite{fan2019rethinking}, SSF \cite{zhang2020}, A2dele \cite{piao2020}, S$^2$MA \cite{liu2020}, ICNet \cite{li2020icnet}, JL-DCF \cite{fu2020jl}, and UC-Net \cite{zhang2020uc}). 
    }\label{fig_00}
\end{figure}

Salient object detection (SOD) aims to locate the most visually prominent object(s) in a given scene~\cite{fan2018salient}. SOD plays a key role in a range of real-world applications, such as stereo matching~\cite{nie2019multi}, 
image understanding \cite{zhu2014unsupervised}, co-saliency detection~\cite{deng2020re}, action recognition \cite{rapantzikos2009dense}, video detection and segmentation \cite{fan2019shifting,wang2017saliency,song2018pyramid,wang2017video}, semantic segmentation \cite{shimoda2016distinct,zeng2019joint}, medical image segmentation~\cite{fan2020pra,fan2020inf,wu2020jcs}, object tracking \cite{mahadevan2009saliency,hong2015online}, person re-identification \cite{zhao2016person,martinel2015kernelized}, camouflaged object detection~\cite{fan2020camouflaged}, image retrieval~\cite{liu2013model}, \etc. Although significant progress has been made in the SOD field over the past several years \cite{zhao2019egnet,tu2016real,xia2017and,hou2007saliency,yan2013hierarchical,yang2013saliency,li2016deep,li2016deep,zhang2016co,zhang2017amulet,zhang2017learning,wang2017stagewise,Li_2018_ECCV,wangwen2019salient,su2019selectivity,zhao2019pyramid}, there is still room for improvement when faced with challenging factors, such as complicated background or different lighting conditions in the scenes. One way to overcome these challenges is to employ depth maps, which provide complementary spatial information for RGB images and have become easier to capture due to the large availability of depth sensors (\eg, Microsoft Kinect).

\begin{figure*}[t]
\centering
\begin{overpic}[width=0.99\linewidth]{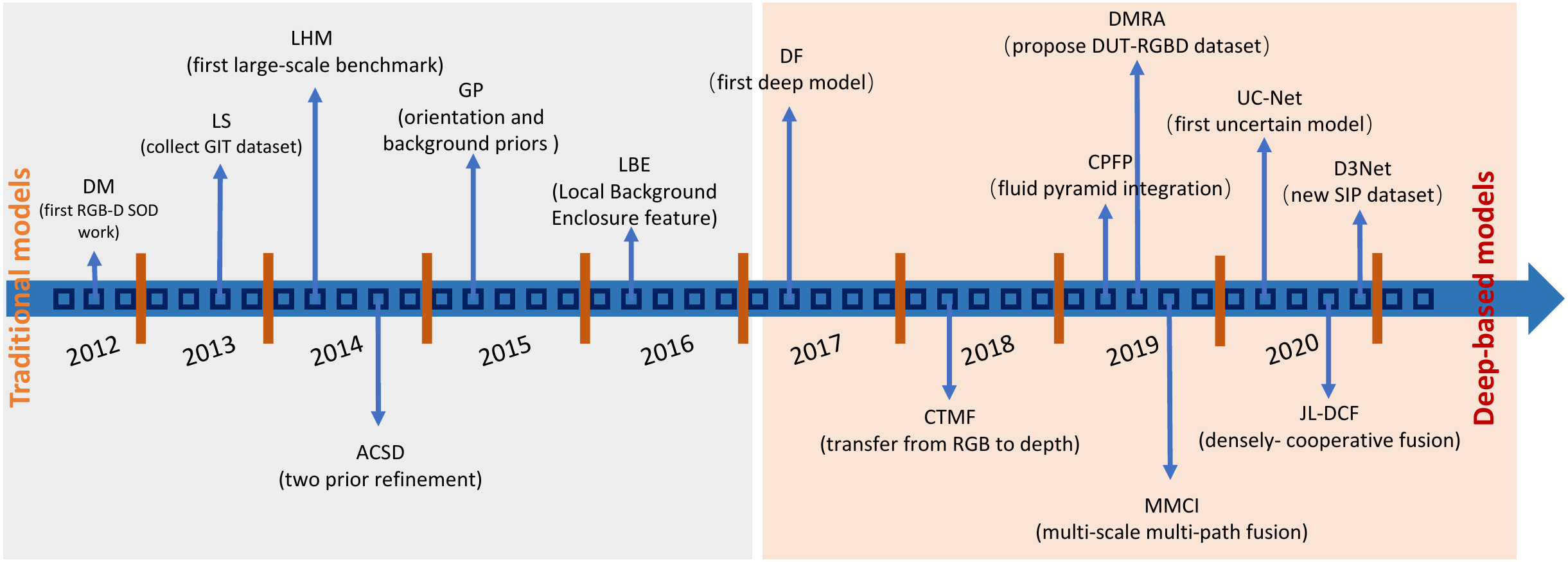}
\put(7.7,23.7){\footnotesize \cite{lang2012depth}}  
\put(15.2,27.7){\footnotesize \cite{ciptadi2013depth}}  
\put(21.3,33.2){\footnotesize \cite{peng2014rgbd}}      
\put(25.8,6.6){\footnotesize \cite{ju2014depth}}        
\put(31.1,29.9){\footnotesize \cite{ren2015exploiting}} 
\put(41.7,25.0){\footnotesize \cite{feng2016local}}     
\put(51.7,31.9){\footnotesize \cite{qu2017rgbd}}        
\put(62.4,9.0){\footnotesize \cite{han2017cnns}}        
\put(72.8,25.3){\footnotesize \cite{zhao2019contrast}}  
\put(74.6,34.4){\footnotesize \cite{piao2019depth}}     
\put(76.6,3.4){\footnotesize \cite{chen2019multi}}      
\put(83.4,29.3){\footnotesize \cite{zhang2020uc}}       
\put(86.9,9.3){\footnotesize \cite{fu2020jl}}           
\put(89.0,24.9){\footnotesize \cite{fan2019rethinking}}           

\end{overpic}\vspace{-0.25cm}

\caption{ A brief chronology of RGB-D based SOD. The first early RGB-D based SOD work was the DM \cite{lang2012depth} model, proposed in 2012. Deep learning techniques have been widely applied to RGB-D based SOD since 2017. More details can be found in \secref{sec:models}.}\label{fig_0}

\end{figure*}
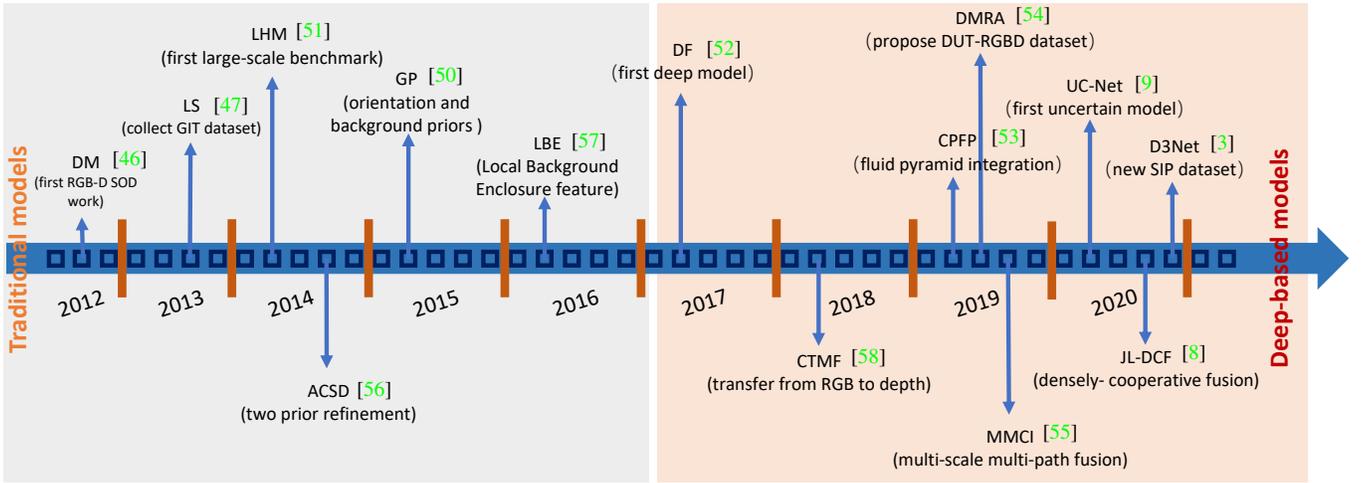

Recently, RGB-D based SOD has gained increasing attention and various methods have been developed \cite{fan2019rethinking,chen2019cnn}. Early RGB-D based SOD models tended to extract handcrafted features and then fuse RGB image and depth maps. For example, Lang \etal~\cite{lang2012depth}, the first work on RGB-D based SOD, utilized Gaussian mixture models to model the distribution of depth-induced saliency. Ciptadi \etal~\cite{ciptadi2013depth} extracted 3D layout and shape features from depth measurements. Besides, several methods \cite{desingh2013depth,cheng2014depth,cheng2014depth,ren2015exploiting} measure depth contrast using the depth difference between different regions. In \cite{peng2014rgbd}, a multi-contextual contrast model including local, global, and background contrast was developed to detect salient objects using depth maps. More importantly, however, this work also provided the first large-scale RGB-D dataset for SOD. Despite the effectiveness achieved by traditional methods using handcrafted features, they tend to suffer from a limited generalization ability for low-level features and lack the high-level reasoning required for complex scenes. To address these limitations, several deep learning-based RGB-D SOD methods \cite{fan2019rethinking} have been developed, showing improved performance. DF \cite{qu2017rgbd} was the first model to introduce deep learning technology into the RGB-D based SOD task. More recently, various deep learning-based models \cite{zhao2019contrast,piao2019depth,chen2019multi,zhang2020uc,fu2020jl,li2020icnet,liu2020} have focused on exploiting effective multi-modal correlations and multi-scale/level information to boost SOD performance. To more clearly describe the progress in the RGB-D based SOD field, we provide a brief chronology in Fig.~\ref{fig_0}.

In this paper, we provide a comprehensive survey on RGB-D based SOD, aiming to thoroughly cover various aspects of the models for this task and provide insightful discussions on the challenges and open directions for future work. We also review another related topic, \ie, light field SOD, in which the light field can provide more information (including focal stack, all-focus images, and depth maps) to boost the performance of salient object detection. Further, we provide a comprehensive comparison to evaluate existing RGB-D based SOD models and discuss their main advantages.

\subsection{Related Reviews and Surveys}

There are several surveys that are closely related to salient object detection. For example,  Borji \etal~\cite{borji2015salient} provided a 
quantitative evaluation of 35 state-of-the-art non-deep saliency detection methods. Cong \etal~\cite{cong2018review} reviewed several different saliency detection models, including RGB-D based SOD, co-saliency detection, and video SOD. Zhang \etal~ \cite{zhang2018review} provided an overview of co-saliency detection and reviewed its history, and summarized several benchmark algorithms in this field. Han \etal~ \cite{han2018advanced} reviewed the recent progress in SOD, including models, benchmark datasets, and evaluation metrics, as well as discussed the underlying connection among general object detection, SOD, and category-specific object detection. Nguyen \etal~\cite{nguyen2018attentive} reviewed various works related to saliency applications and provided insightful discussions on the role of saliency in each. Borji \etal~\cite{borji2014salient} provided a comprehensive review of recent progress in SOD and discussed some related works, including generic scene segmentation, saliency for fixation prediction, and object proposal generation. Fan \etal~\cite{fan2018salient} provided a comprehensive evaluation of several state-of-the-art CNNs-based SOD models, and proposed a high quality SOD dataset, termed \textbf{SOC} (details can be found at: \href{http://dpfan.net/socbenchmark/}{http://dpfan.net/socbenchmark/}). Zhao \etal~\cite{zhao2019object} reviewed various deep learning-based object detection models and algorithms in detail, as well as various specific tasks, including SOD works. Wang \etal~\cite{wang2019salient} focused on reviewing deep learning-based SOD models. Different from previous SOD surveys, in this paper, we focus on reviewing the existing RGB-D based SOD models and benchmark datasets.

\subsection{Contributions}
Our main contributions are summarized as follows:

\begin{itemize}

\item We provide the first systematic review of RGB-D based SOD models from different perspectives. We summarize existing RGB-D SOD models into traditional or deep methods, fusion-wise methods, single-stream/multi-stream methods, and attention-aware methods.

\item We review nine RGB-D datasets that are commonly used in this field, and provide details for each dataset. Moreover, we provide a comprehensive as well as an attribute-based evaluation of several representative RGB-D based SOD models. 

\item We supply the first collection and review of the related light field SOD models and benchmark datasets.

\item We thoroughly investigate several challenges for RGB-D based SOD, and the relation between SOD and other topics, shedding light on potential directions for future research.

\end{itemize}

\subsection{Organization}

In \secref{sec:models}, we review existing RGB-D based models in terms of different aspects. In \secref{sec:dataset}, we summarize and provide details for current benchmark datasets for RGB-D salient object detection. In \secref{sec:lightfield}, we conduct a comprehensive review of light field SOD models and benchmark datasets. In \secref{sec:evaluation}, we provide a comprehensive and attribute-based evaluation of several representative RGB-D based models. We then discuss challenges and open directions of this field in \secref{sec:challenge}. Finally, we conclude this paper in \secref{sec:conclusion}.

\section{RGB-D based SOD Models}
\label{sec:models}

Over the past few years, several RGB-D based SOD methods have been developed and obtained promising performance. These models are summarized in Tables ~\ref{tab:001}, \ref{tab:002}, \ref{tab:003} and \ref{tab:004}. The complete benchmark can be found at \href{http://dpfan.net/d3netbenchmark/}{http://dpfan.net/d3netbenchmark/}.
To review these RGB-D based SOD models in detail,we introduce them from different perspectives as follows. (1) \textbf{Traditional/deep models}: they are viewed from the perspective of feature extraction, that is using the manual features or  deep features. It is convenient for follow-up researchers to grasp the historical development trends of RGB-D SOD models. (2) \textbf{Fusion-wise models}: it is critical to effectively fuse RGB and depth images in this task, thus we review different fusion strategies to understand their effectiveness. (3) \textbf{Single-stream/multi-stream models}: we consider this problem from the perspective of model parameters. Single stream can save parameters, but the final result may not be optimal, and multiple streams may require more parameters. Thus, it is helpful to understand the balance between the amount of calculation and accuracy of different models. (4) \textbf{Attention-aware models}: attention mechanisms have widely been applied in various visual tasks including SOD. We review related works on RGB-D SOD to analyze how do different models use attention. Thus, it is an alternative to design attention modules for future works.

\renewcommand\arraystretch{1.0}
\begin{table*}[t!]
    \centering
    
    \caption{Summary of RGB-D based SOD methods (published from 2012 to 2016).}
    \scriptsize
    \vspace {-2.5mm}
    \label{tab:001}
    \setlength{\tabcolsep}{3.5pt}
    \begin{tabular}{|p{0.4cm}|p{0.6cm}<{\centering}|r|p{1.0cm}<{\centering}|p{1.8cm}|p{1.2cm}|p{9.0cm}|}
    
        \hline
         \# & Year & Method & Pub.  & Training Set & Backbone & Description \\ \hline\hline
    
    1 & 2012 & DM \cite{lang2012depth}  & ECCV    & Without & Without & Models the correlation between saliency and depth by approximating the joint density using Gaussian mixture models \\ \hline 

    2 & 2012 & RCM \cite{zhang2012depth} & ICCSE  & Without  & Without & Develops a region contrast based SOD model with depth cues  \\ \hline 
    
    3 & 2013 & LS \cite{ciptadi2013depth} & BMVC  & Without  & Without & Extends the dissimilarity framework to model the joint interaction between  depth cues and RGB images \\ \hline
          
    4 & 2013 & RC \cite{desingh2013depth} & BMVC  & Withoutt  & Without & Derives RGB-D saliency by formulating a 3D saliency model based on the region contrast of the scene and fuses it using SVM \\ \hline 

    5 & 2013 & SOS \cite{lei2013evaluation} & NEURO  & Without  & Without & Incorporates depth cues for salient object segmentation by suppressing background regions \\ \hline 

    6 & 2014 & SRDS \cite{fan2014salient} & ICDSP  & Without  & Without & Integrates depth and depth weighted color contrast with spatial compactness of color distribution  \\ \hline  
       
    7 & 2014 & LHM \cite{peng2014rgbd} & ECCV  & Without  & Without & Uses a multi-stage RGB-D algorithm to combine both depth and appearance cues to segment salient objects  \\ \hline   
    
    8 & 2014 & DESM \cite{cheng2014depth} & ICIMCS  & Without  & Without & Combines three saliency cues: color contrast, spatial bias, and depth contrast  \\ \hline  

    9 & 2014 & ACSD \cite{ju2014depth} & ICIP  & Without  & Without & Measures a point's saliency by how much it stands out from the surroundings, and has two priors (regions nearer to viewers are more salient and salient objects tend to be located at the center) \\ \hline 

    10 & 2015 & GP \cite{ren2015exploiting} & CVPRW  & Without & Without & Explores orientation and background priors for detecting salient objects, and uses PageRank and MRFs to optimize the saliency maps \\ \hline  
    
    11 & 2015 & SFP \cite{guo2015salient} & ICIMCS  & Without  & Without & Develops a RGB-D based SOD approach using saliency fusion and propagation\\ \hline  

    12 & 2015 & DIC \cite{tang2016depth} & TVC  & Without  & Without & Fuses the saliency maps from color and depth to generate a noise-free salient patch, and utilizes random walk algorithm to infer the object boundary \\ \hline  

    13 & 2015 & SRD \cite{jiang2015salient} & ICRA  & Without  & Without & Designs a graph-based segmentation to identify homogeneous regions using color and depth cues \\ \hline  

    14 & 2015 & MGMR \cite{xue2015rgb} & ICIP  & Without  & Without & Designs a mutual guided manifold ranking strategy to achieve SOD  \\ \hline  
    
    15 & 2015 & SF \cite{zhu2015selective} & CAC  & Without  & Without & Proposes to automatically select discriminative features using decision trees for better performance   \\ \hline  
    
    16 & 2016 & PRC \cite{du2016improving} & ACCESS  & Without  & Without & Saliency fusion and progressive region classification are used to optimize depth-aware saliency models \\ \hline 
    
    17 & 2016 & LBE \cite{feng2016local} & CVPR  & Without  & Without & Uses a local background enclosure to capture the spread of angular directions \\ \hline 
    
    18 & 2016 & SE \cite{guo2016salient} & ICME  & Without  & Without & Utilizes cellular automata to propagate the initial saliency map and then generate the final saliency prediction result \\ \hline 
    
    19 & 2016 & DCMC \cite{cong2016saliency} & SPL  & Without  & Without & Develops a new measure to evaluate the reliability of depth maps for reducing the influence of poor-quality depth maps on saliency detection.
    \\ \hline  

    20 & 2016 & BF \cite{wang2016rgb} & ICPR  & Without  & Without & Fuses contrasting features from RGB and depth images with a Bayesian framework
    \\ \hline  
    
    21 & 2016 & DCI \cite{sheng2016saliency} & ICASSP  & Without  & Without & Adopts the original depth map to subtract the fitted surface for generating a contrast increased map \\ \hline    

    22 & 2016 & DSF \cite{song2016depth} & ICASSP  & Without  &Without  & Develops a multi-stage depth-aware saliency model for SOD \\ \hline    

    23 & 2016 & GM \cite{wang2016visual} & ACCV    & Without  &Without  & Combines color and depth-based contrast features using a generative mixture model \\ \hline

    \hline
    \end{tabular}
\end{table*}

\renewcommand\arraystretch{1.1}
\begin{table*}[t!]
    \centering
    
    \caption{Summary of RGB-D based SOD methods (published from 2017 to 2018).}
    \scriptsize
    \vspace {-2.5mm}
    \label{tab:002}
    \setlength{\tabcolsep}{3.5pt}
    \begin{tabular}{|p{0.4cm}|p{0.6cm}<{\centering}|r|p{1.0cm}<{\centering}|p{2.0cm}|p{1.2cm}|p{9.0cm}|}
    
        \hline
         \# & Year & Method & Pub.  & Training Set & Backbone & Description \\ \hline\hline
      
    24 & 2017 & HOSO \cite{feng2017hoso} & DICTA    & Without & Without & Combines surface orientation distribution contrast with color and depth contrast  \\ \hline 

    25 & 2017 & M{$^3$}Net \cite{chen2017m} & IROS  & NLPR(0.65K), NJUD(1.4K) & VGG-16 & Designs a multi-path multi-modal fusion strategy to integrate RGB and depth images in a task-motivated and adaptive way \\ \hline

    26 & 2017 & MFLN \cite{chen2017rgb} & ICCVS  & NLPR(0.65K), NJUD(1.4K) & AlexNet & Leverages a CNN to learn high-level representations for depth maps, and uses a multi-modal fusion network to integrate RGB and depth representations for RGB-D based SOD \\ \hline 
    
    27 & 2017 & BED \cite{shi2017learning} & ICCVW  & NLPR(0.6K), NJUD(1.2K) & GoogleNet& Uses a CNN to integrate top-down and bottom-up information for RGB-D based SOD, and uses a mid-level feature representation to capture background enclosure\\ \hline 

    28 & 2017 & CDCP \cite{zhu2017innovative} & ICCVW  & Without & Without & Proposes a novel RGB-D SOD algorithm using a center dark channel prior to boost performance \\ \hline 
    
    29 & 2017 & TPF \cite{zhu2017three} & ICCVW  & Without & Without & Leverages stereopsis to generate optical flow, which can provide an additional cue (depth cue) for producing the final detection result \\ \hline    

    30 & 2017 & MFF \cite{wang2017rgb} & SPL  & Without & Without & Uses a multistage fusion framework to integrate multiple visual priors from the RGB image and depth cue for SOD  \\ \hline 
    
    31 & 2017 & MDSF \cite{song2017depth} & TIP  & NLPR(0.5K), NJUD(1.5K) & Without & Proposes a RGB-D SOD framework via a multi-scale discriminative saliency fusion strategy, and utilizes bootstrap learning to achieve the SOD task \\ \hline
    
    32 & 2017 & DF \cite{qu2017rgbd} & TIP  & NLPR(0.75K), NJUD(1.0K) & Without & Feeds RGB and depth features into a CNN architecture to derive the saliency confidence value, and uses Laplacian propagation to produce the final detection result \\ \hline 

    33 & 2017 & MCLP \cite{cong2017iterative} & TCYB & Without & Without & Utilizes the additional depth maps and employs the existing RGB saliency map as an initialization using a refinement-cycle model to obtain the final co-saliency map \\ \hline 
    
    34 & 2018 & ISC \cite{imamoglu2018integration} & SIVP & Without & Without & Fuses salient features using both bottom-up and top-down saliency cues \\ \hline 
    
    35 & 2018 & HSCS \cite{cong2018hscs} & TMM  & Without & Without & Utilizes a hierarchical sparsity reconstruction and energy function refinement for RGB-D based co-saliency detection \\ \hline  
    
    36 & 2018 & ICS \cite{cong2017co} & TIP  &  Without & Without & Exploits the constraint correlation among multiple images and introduces depth maps into the co-saliency model  \\ \hline     

    37 & 2018 & CTMF \cite{han2017cnns} & TCYB  & NLPR(0.65K), NJUD(1.4K) & VGG-16 & Transfers the structure of the deep color network to be applicable for the depth modality and fuses both modalities to produce the final saliency map  \\ \hline  

    38 & 2018 & PCF \cite{chen2018progressively} & CVPR  & NLPR(0.65K), NJUD(1.4K) & VGG-16 & Designs the first multi-scale fusion architecture and a novel complementarity-aware fusion module to fuse both cross-modal and cross-level features \\ \hline  

    39 & 2018 & SCDL \cite{huang2018rgbd} & ICDSP  & NLPR(0.75K), NJUD(1.0K) & VGG-16 & Designs a new loss function to increase the spatial coherence of salient objects  \\ \hline 

    40 & 2018 & ACCF \cite{chen2018attention} & IROS  & NLPR(0.65K), NJUD(1.4K)  & VGGNet & Adaptively selects complementary features from different modalities at each level, and then performs more informative cross-modal cross-level combinations\\ \hline   

    41 & 2018 & CDB \cite{liang2018stereoscopic} & NEURO  &Without & Without & Utilizes a contrast prior and depth-guided-background prior to construct a 3D stereoscopic saliency model \\ \hline  

    \hline
    \end{tabular}
\end{table*}

\subsection{Traditional/Deep Models}

\textbf{Traditional Models}. With depth cues, several useful attributes, such as boundary cues, shape attributes, surface normals, etc., can be explored to boost the identification of salient objects in complex scenes. Over the past several years, many traditional RGB-D models based on handcrafted features have been developed \cite{ciptadi2013depth,desingh2013depth,fan2014salient,peng2014rgbd,cheng2014depth,ju2014depth,guo2015salient,ren2015exploiting,tang2016depth,du2016improving,feng2016local,guo2016salient,cong2016saliency,chen2017rgb,shi2017learning,zhu2017innovative,liang2018stereoscopic}. For example, the early work \cite{ciptadi2013depth} focused on modeling the interaction between layout and shape features generated from the RGB image and depth map. Besides, the representative work \cite{peng2014rgbd} developed a novel multi-stage RGB-D model, and constructed the first large-scale RGB-D benchmark dataset, termed NLPR.

\textbf{Deep Models}. However, the above-mentioned methods suffer from unsatisfactory SOD performance due to the limited expression ability of handcrafted features. To address this, several studies have turned to deep neural networks (DNNs) to fuse RGB-D data \cite{shi2017learning,qu2017rgbd,huang2018rgbd,chen2018attention,liu2019salient,zhu2019pdnet,chen2019multi,piao2019saliency,chen2019three,chen2019discriminative,cong2019going,wang2019adaptive,zhao2019contrast,piao2019depth,zhou2020attention,liu2020cross,liang2020cocnn,li2020asif,huang2020triple,li2020icnet,chen2020improved,fu2020jl,piao2020,zhang2020,zhang2020uc}. These models can learn high-level representations to explore complex correlations across RGB images and depth cues for improving SOD performance. We review some representative works in detail as follows.

$\bullet$ \textbf{DF} \cite{qu2017rgbd} develops a novel convolutional neural network (CNN) to integrate different low-level saliency cues into hierarchical features, for effectively locating salient regions in RGB-D images. This was the first CNN-based model for the RGB-D SOD task. However, it utilizes a shallow architecture to learn the saliency map. 

$\bullet$ \textbf{PCF} \cite{chen2018progressively} presents a complementarity-aware fusion module to integrate cross-modal and cross-level feature representations. It can effectively exploit complementary information by explicitly using cross-modal/level connections and modal/level-wise supervision to decrease fusion ambiguity.

$\bullet$ \textbf{CTMF} \cite{han2017cnns} employs a computational model to identify salient objects from RGB-D scenes, utilizing CNNs to learn high-level representations for RGB images and depth cues, while simultaneously exploiting the complementary relationships and joint representation. Besides, this model transfers the structure of the model from the source domain (\emph{i.e.}, RGB images) to be applicable to the target domain (\emph{i.e.}, depth maps). 

$\bullet$ \textbf{CPFP} \cite{zhao2019contrast} proposes a contrast-enhanced network to produce an enhanced map, and presents a fluid pyramid integration module to effectively fuse cross-modal information in a hierarchical manner. Besides, considering the fact that depth cues tend to suffer from noise, a feature-enhanced module is proposed to learn an enhanced depth cue for boosting the SOD performance. It is worth noting that this is an effective solution.

$\bullet$ \textbf{UC-Net} \cite{zhang2020uc} proposes a probabilistic RGB-D based SOD network via conditional variational autoencoders (VAEs) to model human annotation uncertainty. It generates multiple saliency maps for each input image by sampling in the learned latent space. This was the first work to investigate uncertainty in RGB-D based SOD, and was inspired by the data labeling process. This method leverages the diverse saliency maps to improve the final SOD performance. 

\begin{figure*}[t]
    \vspace {-4mm}
    \centering
    \includegraphics[width=0.99\linewidth]{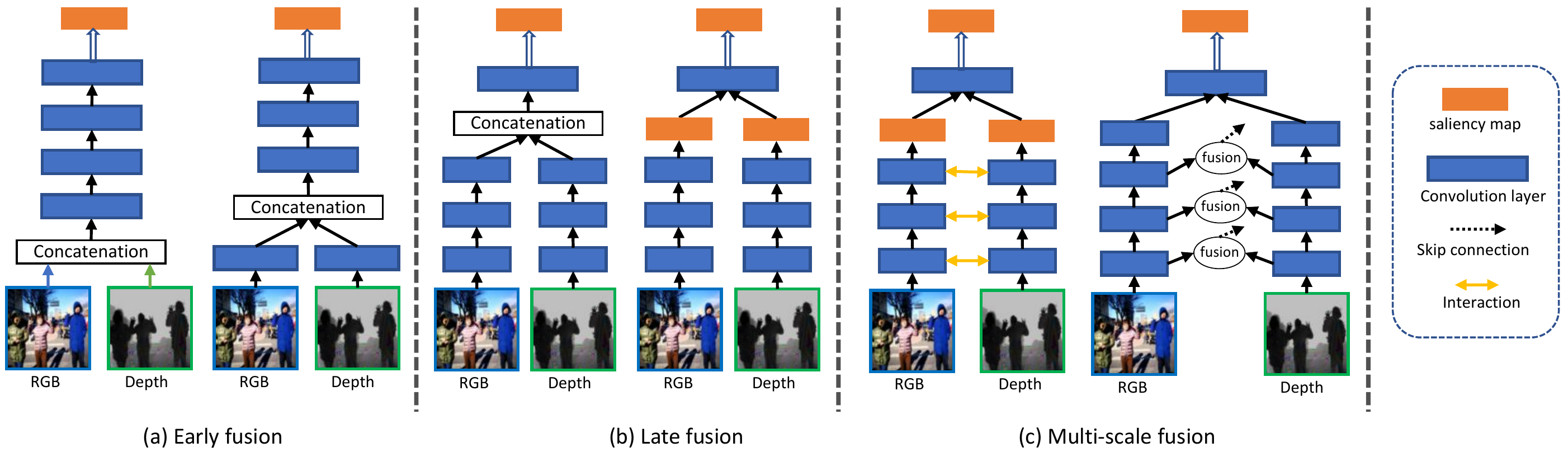} \vspace {-1mm}
    \caption{Comparison of three fusion strategies that explore the correlation between RGB images and depth maps for RGB-D based SOD. These include: 1) Early fusion; 2) Late fusion; 3) Multi-scale fusion. 
    }\label{fig_1}
\end{figure*}

\renewcommand\arraystretch{1.0}
\begin{table*}[t!]
    \centering
    
    \caption{Summary of RGB-D based SOD models published in 2019 and 2020}
    \scriptsize
    \vspace {-2.5mm}
    \label{tab:003}
    \setlength{\tabcolsep}{4.4pt}
    \begin{tabular}{|p{0.4cm}|p{0.6cm}<{\centering}|r|p{0.95cm}<{\centering}|p{2.0cm}|p{1.2cm}|p{8.0cm}|}

        \hline
         No. & Year & Method & Pub.  & Training Set & Backbone & Description \\ \hline\hline
    
    42 & 2019 & SSRC \cite{liu2019salient} & NEURO  &NLPR(0.65K), NJUD(1.4K) & VGG-16& Uses a single-stream recurrent convolutional neural network with a four-channel input and DRCNN subnetwork \\ \hline  
    
    43 & 2019 & MLF \cite{huang2019rgb} & SPL  & NJUD(1.588K) & VGG-16 & Designs a salient object-aware data augmentation method to expand the training set \\ \hline      

    44 & 2019 & TSRN \cite{liu2019two} & ICIP  & NJUD(1.387K) & VGG-16  & Designs a fusion refinement module to integrate output features from different modalities and resolutions \\ \hline     
    
    45 & 2019 & DIL \cite{du2019salient} & MTAP  & NLPR(0.5K), NJUD(0.5K) & Without & Designs a consistency integration strategy to generate an image pre-segmentation result that is consistent with the depth distribution \\ \hline     
    
    46 & 2019 & CAFM \cite{zhou2019global} & TSMC  & NUS \cite{lang2012depth}, NCTU \cite{ma2015learning} & VGG-16 & Utilizes a content-aware fusion module to integrate global and local information \\ \hline     
    
    47 & 2019 & PDNet \cite{zhu2019pdnet} & ICME  & NLPR(0.5K), NJUD(1.5K) & VGG-16 & Adopts a prior-model guided master network to process RGB information, which is pre-trained on the conventional RGB dataset to overcome the limited size  \\ \hline  

    48 & 2019 & MMCI \cite{chen2019multi} & PR  & NLPR(0.65K), NJUD(1.4K) & VGG-16 & Improves the traditional two-stream architecture by diversifying the multi-modal fusion paths and introducing cross-modal interactions in multiple layers  \\ \hline     
   
    49 & 2019 & TANet \cite{chen2019three} & TIP  & NLPR(0.65K), NJUD(1.4K)  & VGG-16 & Uses a three-stream multi-modal fusion framework to explore cross-modal complementarity in both the bottom-up and top-down processes \\ \hline  
    
    50 & 2019 & DCMF \cite{chen2019discriminative} & TCYB  & NLPR(0.65K), NJUD(1.4K)   & VGG-16 & Formulates a CNN-based cross-modal transfer learning problem for depth-induced SOD, and uses a dense cross-level feedback strategy to exploit cross-level interactions \\ \hline  

    51 & 2019 & DGT \cite{cong2019going} & TCYB  & Without  & Without & Exploits depth cues and provides a general transformation model from RGB saliency to RGB-D saliency \\ \hline 

    52 & 2019 & LSF \cite{chen2019cnn} & arXiv  & NLPR(0.65K), NJUD(1.4K)  & VGG & Designs an RGB-D system with three key components, including modality-specific representation learning, complementary information selection, and cross-modal complements fusion\\ \hline 

    53 & 2019 & AFNet \cite{wang2019adaptive} & ACCESS  & NLPR(0.65K), NJUD(1.4K) & VGG-16  & Learns a switch map that is used to adaptively fuse the predicted saliency maps from the RGB and depth modality \\ \hline  

    54 & 2019 & EPM \cite{jin2019co} & ACCESS  & Without & Without & Develops an effective propagation mechanism for RGB-D co-saliency detection \\ \hline  
    
    55 & 2019 & CPFP \cite{zhao2019contrast} & CVPR  & NLPR(0.65K), NJUD(1.4K) & VGG-16 & Uses a contrast-enhanced network to obtain the one-channel enhanced map, and designs a fluid pyramid integration module to fuse cross-modal cross-level features in a pyramid style  \\ \hline   

    56 & 2019 & DMRA \cite{piao2019depth} & ICCV  & NLPR(0.7K), NJUD(1.485K) & VGG-19  & Designs a depth-induced multiscale recurrent attention network for SOD, including a depth refinement block and a recurrent attention module\\ \hline 
    
    57 & 2019 & DSD \cite{ding2019depth} & JVCIR  & NLPR(0.5K), NJUD(1.5K) & VGG-16  & Uses a saliency fusion network to adaptively fuse both the color and depth saliency maps \\ \hline
    
    58 & 2020 & DPANet \cite{chen2020depth} & arXiv  & NLPR(0.65K), NJUD(1.4K), DUT(0.8K) & ResNet-50 & Uses a saliency-orientated depth perception module to evaluate the potentiality of depth maps and reduce effects of contamination \\ \hline
    
    59 & 2020 & SSDP \cite{wang2020} & arXiv  & NLPR(0.7K), NJUD(1.485K), DUT(0.8K)  & VGG-19 & Makes use of existing labeled RGB saliency datasets together with unlabeled RGB-D data to boost SOD performance\\ \hline    
        
    60 & 2020 & AttNet \cite{zhou2020attention} & IVC  & NLPR(0.65K), NJUD(1.4K)  & VGG-16 & Deploys attention maps to boost the salient objects' location and pays more attention to the appearance information \\ \hline        

    61 & 2020 & --- \cite{liu2020cross} & NEURO  & NLPR(0.65K), NJUD(1.4K) & VGG-16 & Uses an adaptive gated fusion module via a GAN to obtain a better fused saliency map from RGB images and depth cues \\ \hline     

    62 & 2020 & CoCNN \cite{liang2020cocnn} & PR  & STERE, NJUD & VGG-16 & Fuses color and disparity features from low to high layers in a unified deep model \\ \hline

    63 & 2020 & cmSalGAN \cite{jiang2020cmsalgan} & TMM  & NLPR(0.65K), NJUD(1.4K) & ResNet-50 & Aims to learn an optimal view-invariant and consistent pixel-level representation for both RGB and depth images using an adversarial learning framework\\ \hline

    64 & 2020 & PGHF \cite{xiao2020multi} & ACCESS  & NLPR(0.65K), NJUD(1.4K) & VGG-16 & Leverages powerful representations learned from large-scale RGB datasets to boost the model ability\\ \hline
        
        \hline
    \end{tabular}
\end{table*}

\renewcommand\arraystretch{1.0}
\begin{table*}[t!]
    \centering
    
    \caption{Summary of RGB-D based SOD models published in 2020.}
    \scriptsize
    \vspace {-2.5mm}
    \label{tab:004}
    \setlength{\tabcolsep}{4.4pt}
    \begin{tabular}{|p{0.4cm}|p{0.6cm}<{\centering}|r|p{0.75cm}<{\centering}|p{2.0cm}|p{1.2cm}|p{8.5cm}|}
        \hline
         No. & Year & Method & Pub.  & Training Set & Backbone & Description \\ \hline\hline

    65 & 2020 & BiANet \cite{zhang2020bilateral} & TIP  & NLPR(0.7K), NJUD(1.485K) & VGG-16 & Uses a bilateral attention module (BAM) to explore rich foreground and background information from depth maps \\ \hline

    66 & 2020 & ASIF-Net \cite{li2020asif} & TCYB  & NLPR(0.65K), NJUD(1.4K) & VGG-16 & Integrates the attention steered complementarity from RGB-D images and introduces a global semantic constraint using adversarial learning \\ \hline

    67 & 2020 & Triple-Net \cite{huang2020triple} & SPL  &  Triple-Net & ResNe-18 & Uses a triple-complementary network for RGB-D based SOD \\ \hline
    
    68 & 2020 & ICNet \cite{li2020icnet} & TIP  &  Triple-Net & VGG-16 & Uses a novel information conversion module to fuse high-level RGB and depth features in an interactive and adaptive way \\ \hline

    69 & 2020 & SDF \cite{chen2020improved} & TIP  & NLPR,NJUD, DEC,LFSD(1.5K) & VGG-16 & Proposes a exemplar-driven method to estimate relatively trustworthy depth maps, and uses a selective deep saliency fusion network to effectively integrate RGB images, original depths, and newly estimated depths \\ \hline

    70 & 2020 & GFNet \cite{zhou2020gfnet} & SPL  & NLPR(0.8K), NJUD(1.588K)& Res2Net & Designs a gate fusion block to regularize feature fusion \\ \hline    
    
    71 & 2020 & RGBS \cite{liu2020salient} & MTAP  &NLPR(0.65K), NJUD(1.4K) & VGG-16 & Utilizes a GAN to generate the saliency map\\ \hline    
    
    72 & 2020 & D$^{3}${Net} \cite{fan2019rethinking} & TNNLS  & NLPR(0.7K), NJUD(1.485K)& VGG-16 & Uses a depth depurator unit (DDU) and a three-stream feature learning module to employ low-quality depth cue filtering and cross-modal feature learning, respectively \\ \hline
        
    73 & 2020 & JL-DCF \cite{fu2020jl} & CVPR  & NLPR(0.7K), NJUD(1.5K) & VGG-16, ResNet-101 & Uses a joint learning strategy and a densely-cooperative fusion module to achieve better SOD performance \\ \hline

    74 & 2020 & A2dele \cite{piao2020} & CVPR  & NLPR(0.7K), NJUD(1.485K)  & VGG-16 & Employs a depth distiller to explore ways of using network prediction and attention as two bridges to transfer depth knowledge to RGB images \\ \hline   

    75 & 2020 & SSF \cite{zhang2020} & CVPR  &NLPR(0.7K), NJUD(1.485K), DUT(0.8K) & AGG-16  & Designs a complimentary interaction module to select useful representations from the RGB and depth images and then integrate cross-modal features \\ \hline

    76 & 2020 & S${^2}$MA \cite{liu2020} & CVPR  &NLPR(0.65K), NJUD(1.4K)  & VGG-16 & Fuses multi-modal information via self-attention and each other’s attention strategies, and reweights the mutual attention term to filter out unreliable information \\ \hline
    
    77 & 2020 & UC-Net \cite{zhang2020uc} & CVPR  & NLPR(0.7K), NJUD(1.5K) & VGG-16 & Uses a probabilistic RGB-D saliency detection network via a conditional VAE to generate multiple saliency maps  \\ \hline

    78 & 2020 & CMWNet \cite{lieccv20} & ECCV  & NLPR(0.65K), NJUD(1.4K)  & VGG-16 & Exploits feature interactions using three cross-modal cross-scale weighting modules to improve SOD performance \\ \hline    

    79 & 2020 & HDFNet \cite{paneccv2020} & ECCV  & NLPR(0.7K), NJUD(1.485K), DUT(0.8K)  & VGG-16 & Designs a hierarchical dynamic filtering network to effectively make use of cross-modal fusion information \\ \hline      

    80 & 2020 & CAS-GNN \cite{luoECCV2020} & ECCV  & NLPR(0.65K), NJUD(1.4K)   & VGG-16 & Designs cascaded graph neural networks to exploit useful knowledge from RGB and depth images for building powerful feature embeddings \\ \hline      
    
    81 & 2020 & CMMS \cite{li2020} & ECCV  & NLPR(0.7K), NJUD(1.485K) & VGG-16   & Proposes a cross-modality feature modulation module to enhance feature representations and an adaptive feature selection module to gradually select saliency-related features \\ \hline    

    82 & 2020 & DANet \cite{zhaoeccv20} & ECCV  & NLPR(0.65K), NJUD(1.4K) & VGG-16, VGG-19   & Develops a single-stream network combined with a depth-enhanced dual attention to achieve real-time SOD \\ \hline    

    83 & 2020 & CoNet \cite{Wei_2020_ECCV} & ECCV  & NLPR(0.7K), NJUD(1.485K), DUT(0.8K) & ResNet & Develops a collaborative learning framework for RGB-D based SOD. Three collaborators (edge detection, coarse salient object detection and depth estimation) are utilized to jointly boost the performance \\ \hline   

    84 & 2020 & BBS-Net \cite{faneccv20} & ECCV  & NLPR(0.65K), NJUD(1.4K)  & VGG-16, VGG-19, ResNet-50  & Uses a bifurcated backbone strategy to learn teacher and student features, and utilizes a depth-enhanced module to excavate informative parts of depth cues \\ \hline    

    85 & 2020 & ATSA \cite{Zhangeccv20} & ECCV  & NLPR(0.7K), NJUD(1.485K), DUT(0.8K)  & VGG-19 & Proposes an asymmetric two-stream architecture taking account of the inherent differences between RGB and depth data for SOD \\ \hline    
    
    86 & 2020 & PGAR \cite{chen2020progressively} & ECCV  & NLPR(0.7K), NJUD(1.485K)   & VGG-16 & Propose a progressively guided alternate refinement network to produce a coarse initial prediction using a multi-scale residual block \\ \hline 

    87 & 2020 & MCINet \cite{huang2020multi} & arXiv  & NLPR(0.65K), NJUD(1.4K)  & ResNet-50  & Develops a novel multi-level cross-modal interaction network for RGB-D SOD  \\ \hline 
    
    88 & 2020 & DRLF \cite{wang2020data} & TIP  & NLPR(0.65K), NJUD(1.4K)   & VGG-16 & Develops a channel-wise fusion network to conduct multi-net and multi-level selective fusion for RGB-D SOD  \\ \hline 
    
    89 & 2020 & DQAM \cite{wang2020knowing} & arXiv  & NLPR(0.65K), NJUD(1.4K) & Without  & Proposes a depth quality assessment solution to conduct ``quality-aware" SOD for RGB-D images  \\ \hline 
        
    90 & 2020 & DQSD \cite{chenchen2020depth} & TIP  & NLPR(0.65K), NJUD(1.4K)  & VGG-19 &  Integrates a depth quality aware subnet into a bi-stream structure to assess the depth quality before conducting RGB-D fusion \\ \hline 
    
    91 & 2020 & DASNet \cite{zhao2020} & ACM MM  & NLPR(0.7K), NJUD(1.5K) & ResNet-50  & Proposes a new perspective of containing the depth constraints in the learning process rather than using depths as inputs  \\ \hline 

    92 & 2020 & DCMF \cite{chen2020rgbd} & TIP  & NLPR(0.65K), NJUD(1.4K) & VGG-16, ResNet-50  & Designs a disentangled
cross-modal fusion network to expose structural and content representations from RGB and depth images \\ \hline 
        \hline
    \end{tabular}
\end{table*}

\subsection{Fusion-wise Models}

For RGB-D based SOD models, it is important to effectively fuse RGB images and depth maps. The existing fusion strategies can be grouped into three categories, including 1) early fusion, 2) multi-scale fusion, and 3) late fusion. We provide details for each fusion strategy as follows.

\textbf{Early Fusion}. Early fusion-based methods can follow one of two veins: 1) RGB images and depth maps are directly integrated to form a four-channel input \cite{peng2014rgbd,ren2015exploiting,song2017depth,liu2019salient,song2017depth}. This is denoted as ``input fusion" (shown in Fig.~\ref{fig_1}); 2) RGB and depth images are first fed into each independent network and their low-level representations are combined as joint representations, which are then fed into a subsequent network for further saliency map prediction \cite{qu2017rgbd}. This is denoted as ``early feature fusion" (shown in Fig.~\ref{fig_1}).

\textbf{Late Fusion}. Late fusion-based methods can also be further divided into two families: 1) Two parallel network streams are adopted to learn high-level features for RGB and depth data, respectively, which are concatenated and then used for generating the final saliency prediction \cite{han2017cnns,desingh2013depth,wang2019adaptive}. This is denoted as ``later feature fusion" (shown in Fig.~\ref{fig_1}). 2) Two parallel network streams are used to obtain the independent saliency maps for RGB images and depth cues, and then the two saliency maps are concatenated to obtain a final prediction map \cite{ding2019depth}. This is denoted as ``late result fusion" (shown in Fig.~\ref{fig_1}).

\textbf{Multi-scale Fusion}. To effectively explore the correlations between RGB images and depth maps, several methods propose a multi-scale fusion strategy \cite{chen2019multi,li2020icnet,zhang2020bilateral,fu2020jl,faneccv20,chen2020depth,lieccv20,paneccv2020}. These models can be divided into two categories. The first category learn the cross-modal interactions and then fuse them into a feature learning network. For example, Chen \etal~\cite{chen2019multi} developed a multi-scale multi-path fusion network to integrate RGB images and depth maps, with a cross-modal interaction (termed MMCI) module. This method introduces cross-modal interactions into multiple layers, which can empower additional gradients for enhancing the learning of the depth stream, as well as enable complementarity across low-level and high-level representations to be explored. The second category fuse the features from RGB images and depth maps in different layers and then integrate them into a decoder network (\eg, skip connection) to produce the final saliency detection map (as shown in Fig.~\ref{fig_1}). Some representative works are briefly discussed as follows.

$\bullet$ \textbf{ICNet} \cite{li2020icnet} proposes an information conversion module to convert high-level features in an interactive manner. In this model, a cross-modal depth-weighted combination (CDC) block is introduced to enhance RGB features with depth features at different levels.

$\bullet$ \textbf{DPANet} \cite{chen2020depth} uses a gated multi-modality attention (GMA) module to exploit long-range dependencies. The GMA module can extract the most discriminative features by utilizing a spatial attention mechanism. Besides, this model controls the fusion rate of the cross-modal information using a gate function, which can reduce some effects brought by the unreliable depth cues.

$\bullet$ \textbf{BiANet} \cite{zhang2020bilateral} employs a multi-scale bilateral attention module (MBAM) to capture better global information in multiple layers. 

$\bullet$ \textbf{JL-DCF} \cite{fu2020jl} treats a depth image as a special case of a color image and employs a shared CNN for both RGB and depth feature extraction. It also proposes a densely-cooperative fusion strategy to effectively combine the learned features from different modalities. 

$\bullet$ \textbf{BBS-Net} \cite{faneccv20} uses a bifurcated backbone strategy (BBS) to split the multi-level feature representations into teacher and student features, and develops a depth-enhanced module (DEM) to explore informative parts in depth maps from the spatial and channel views.

\begin{figure*}[t]
    \vspace {-4mm}
    \centering
    \includegraphics[width=0.98\linewidth]{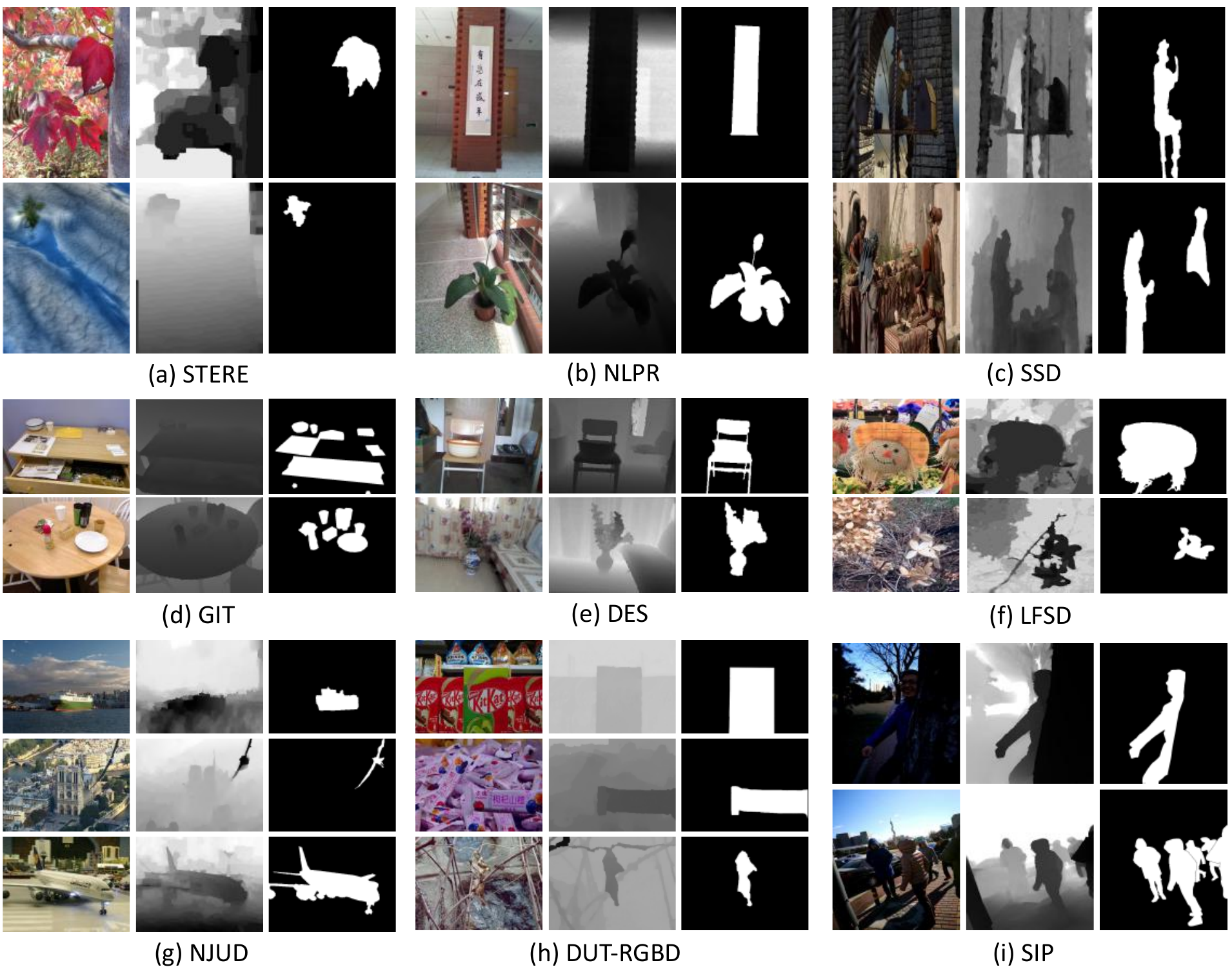} \vspace {-1mm}
    \caption{Examples of images, depth maps and annotations in nine RGB-D dataset, including (a) STERE \cite{niu2012leveraging}, (b) NLPR  \cite{peng2014rgbd}, (c) SSD \cite{zhu2017three}, (d) GIT \cite{ciptadi2013depth}, (e) DES \cite{cheng2014depth} , (f) LFSD \cite{li2014saliency}, (g) NJUD \cite{ju2014depth}, (h) DUT-RGBD \cite{piao2019saliency}, and (i) SIP \cite{fan2019rethinking}. In each dataset, the RGB image, depth map and annotation are shown from left to right.
    }\label{fig_03}
\end{figure*}

\subsection{Single-stream/Multi-stream Models}

\textbf{Single-stream Models}. Several RGB-D based SOD works \cite{shi2017learning,song2017depth,qu2017rgbd,huang2018rgbd,liu2019salient,zhu2019pdnet,zhao2019contrast} focus on a single-stream architecture to achieve saliency prediction. These models often fuse RGB images and depth information in the input channel or feature learning part. For example, MDSF \cite{song2017depth} employs a multi-scale discriminative saliency fusion framework as the SOD model, in which four types of features in three levels are computed and then fused to obtain the final saliency map. BED \cite{shi2017learning} utilizes a CNN architecture to integrate bottom-up and top-down information for SOD, which also incorporates multiple features, including background enclosure distribution (BED) and low level depth maps (\eg, depth histogram distance and depth contrast) to boost the SOD performance. PDNet \cite{zhu2019pdnet} extracts depth-based features using a subsidiary network, which makes full use of depth information to assist the main-stream network.

\textbf{Multi-stream Models}. Two-stream models \cite{wang2019adaptive,piao2019depth,zhou2020attention} consist of two independent branches that process RGB images and depth cues, respectively, and often generate different high-level features or saliency maps and then incorporate them in the middle stage or end of the two streams. It is worth noting that most recent deep learning-based models \cite{chen2018progressively,chen2019multi,chen2019cnn,chen2019discriminative,piao2020,jiang2020cmsalgan,li2020icnet,li2020asif,chen2020depth,liu2020cross} utilize this two-stream architecture with several models capturing the correlations between RGB images and depth cues across multiple layers. Moreover, some models utilize a multi-stream structure \cite{chen2019three,fan2019rethinking} and then design different fusion modules to effectively fuse RGB and depth information in order to exploit their correlations.

\subsection{Attention-aware Models}
Existing RGB-D based SOD methods often treat all regions equally using the extracted features equally, while ignoring the fact that different regions can have different contributions to the final prediction map. These methods are easily affected by cluttered backgrounds. In addition, some methods either regard the RGB images and depth maps as having the same status or overly rely on depth information. This prevents them from considering the importance of different domains (RGB images or depth cues). To overcome this, several methods introduce attention mechanisms to weight the importance of different regions or domains. 

$\bullet$ \textbf{ASIF-Net} \cite{li2020asif} captures complementary information from RGB images and depth cues using an interweaved fusion, and weights the saliency regions through a deeply supervised attention mechanism. 

$\bullet$ \textbf{AttNet} \cite{zhou2020attention} introduces attention maps for differentiating between salient objects and background regions to reduce the negative influence of some low-quality depth cues. 

$\bullet$ \textbf{TANet} \cite{chen2019three} formulates a multi-modal fusion framework using RGB images and depth maps from the bottom-up and top-down views. It then introduces a channel-wise attention module to effectively fuse the complementary information from different modalities and levels. 

\subsection{Open-source Implementations}
We summarize the open-source implementations of RGB-D based SOD models reviewed in this survey. The implementations and hyperlinks of the source codes of these models are provided in Tab~\ref{tab:04}. More source codes will be updated at: \href{https://github.com/taozh2017/RGBD-SODsurvey}{https://github.com/taozh2017/RGBD-SODsurvey}.

\begin{table*}[t!]
    \centering
    \begin{threeparttable}
    \renewcommand\arraystretch{1.0}
    \caption{A summary of RGB-D based SOD models with open-source implementations.}
    \scriptsize
    \vspace {-2.5mm}
    \label{tab:04}
    \setlength{\tabcolsep}{4.4pt}
    \begin{tabular}{p{1.0cm}||p{2.0cm}|p{2.0cm}|p{11.0cm}}
        \hline\hline
         Year & Model & Implementation & Code link \\ \hline\hline
         
\multirow{2}{*}{2014}
 & LHM \cite{peng2014rgbd}      & Matlab  & \href{https://sites.google.com/site/rgbdsaliency/code}{https://sites.google.com/site/rgbdsaliency/code} \bigstrut\\\cline{2-4}
 & DESM \cite{cheng2014depth}     & Matlab  & \href{https://github.com/HzFu/DES\_code}{https://github.com/HzFu/DES\_code} \\ \hline

2015 & GP \cite{ren2015exploiting}    & Matlab  & \href{https://github.com/JianqiangRen/Global\_Priors\_RGBD\_Saliency\_Detection}{https://github.com/JianqiangRen/Global\_Priors\_RGBD\_Saliency\_Detection} \\ \hline

\multirow{2}{*}{2016}
 &DCMC \cite{cong2016saliency}   & Matlab  & \href{https://github.com/rmcong/Code-for-DCMC-method}{https://github.com/rmcong/Code-for-DCMC-method} \bigstrut\\\cline{2-4}
 &LBE \cite{feng2016local}   & Matlab \& C++  & \href{http://users.cecs.anu.edu.au/~u4673113/lbe.html}{http://users.cecs.anu.edu.au/~u4673113/lbe.html} \\ \hline

\multirow{5}{*}{2017}
 &BED \cite{shi2017learning}     & Caffe  & \href{https://github.com/sshige/rgbd-saliency}{https://github.com/sshige/rgbd-saliency} \bigstrut\\\cline{2-4}
 & CDCP \cite{zhu2017innovative}  & Matlab  & \href{https://github.com/ChunbiaoZhu/ACVR2017}{https://github.com/ChunbiaoZhu/ACVR2017} \bigstrut\\\cline{2-4}
& MDSF \cite{song2017depth}      & Matlab  & \href{https://github.com/ivpshu}{https://github.com/ivpshu} \bigstrut\\\cline{2-4}
& DF \cite{qu2017rgbd} & Matlab  & \href{https://pan.baidu.com/s/1Y-PqAjuH9xREBjfl7H45HA}{https://pan.baidu.com/s/1Y-PqAjuH9xREBjfl7H45HA} \\ \hline

\multirow{3}{*}{2018}
& CTMF \cite{han2017cnns} & Caffe  & \href{https://github.com/haochen593/CTMF}{https://github.com/haochen593/CTMF} \bigstrut\\\cline{2-4}

& PCF \cite{chen2018progressively} & Caffe  & \href{https://github.com/haochen593/PCA-Fuse\_RGBD\_CVPR18}{https://github.com/haochen593/PCA-Fuse\_RGBD\_CVPR18} \bigstrut\\\cline{2-4}
& PDNet \cite{zhu2019pdnet}   & TensorFlow &  \href{https://github.com/cai199626/PDNet}{https://github.com/cai199626/PDNet} \\ \hline

\multirow{4}{*}{2019}
& AFNet \cite{wang2019adaptive}   & TensorFlow &  \href{https://github.com/Lucia-Ningning/Adaptive_Fusion\_RGBD\_Saliency\_Detection}{https://github.com/Lucia-Ningning/Adaptive\_Fusion\_RGBD\_Saliency\_Detection} \bigstrut\\\cline{2-4}

& CPFP \cite{zhao2019contrast} & Caffe & \href{https://github.com/JXingZhao/ContrastPrior}{https://github.com/JXingZhao/ContrastPrior} \bigstrut\\\cline{2-4}
 &DMRA \cite{piao2019depth}  & PyTorch & \href{https://github.com/jiwei0921/DMRA}{https://github.com/jiwei0921/DMRA} \bigstrut\\\cline{2-4}

& DGT \cite{cong2019going}  & Matlab & \href{https://github.com/rmcong/Code-for-DTM-Method}{https://github.com/rmcong/Code-for-DTM-Method} \\ \hline

\multirow{24}{*}{2020}
&ICNet \cite{li2020icnet} & Caffe & \href{https://github.com/MathLee/ICNet-for-RGBD-SOD}{https://github.com/MathLee/ICNet-for-RGBD-SOD} \bigstrut\\\cline{2-4}
 &JL-DCF \cite{fu2020jl} & Pytorch, Caffe & \href{https://github.com/kerenfu/JLDCF}{https://github.com/kerenfu/JLDCF} \bigstrut\\\cline{2-4}

 &A2dele \cite{piao2020} & PyTorch & \href{https://github.com/OIPLab-DUT/CVPR2020-A2dele}{https://github.com/OIPLab-DUT/CVPR2020-A2dele} \bigstrut\\\cline{2-4}

& SSF \cite{zhang2020} & PyTorch & \href{https://github.com/OIPLab-DUT/CVPR\_SSF-RGBD}{https://github.com/OIPLab-DUT/CVPR\_SSF-RGBD} \bigstrut\\\cline{2-4}

& ASIF-Net \cite{li2020asif} & TensorFlow & \href{https://github.com/Li-Chongyi/ASIF-Net}{https://github.com/Li-Chongyi/ASIF-Net} \bigstrut\\\cline{2-4}

& S${^2}$MA \cite{liu2020} & PyTorch & \href{https://github.com/nnizhang/S2MA}{https://github.com/nnizhang/S2MA} \bigstrut\\\cline{2-4}

&UC-Net \cite{zhang2020uc}   & PyTorch &  \href{https://github.com/JingZhang617/UCNet}{https://github.com/JingZhang617/UCNet} \bigstrut\\\cline{2-4}

 & D$^{3}${Net} \cite{fan2019rethinking} & PyTorch & \href{https://github.com/DengPingFan/D3NetBenchmark}{https://github.com/DengPingFan/D3NetBenchmark} \bigstrut\\\cline{2-4}

  & CMWNet \cite{lieccv20} & Caffe & \href{https://github.com/MathLee/CMWNet}{https://github.com/MathLee/CMWNet} \bigstrut\\\cline{2-4}

  & HDFNet \cite{paneccv2020} & PyTorch & \href{https://github.com/lartpang/HDFNet}{https://github.com/lartpang/HDFNet} \bigstrut\\\cline{2-4}
 
  & CMMS \cite{li2020} & TensorFlow & \href{https://github.com/Li-Chongyi/cmMS-ECCV20}{https://github.com/Li-Chongyi/cmMS-ECCV20} \bigstrut\\\cline{2-4}

  & CAS-GNN \cite{luoECCV2020} & PyTorch & \href{https://github.com/LA30/Cas-Gnn}{https://github.com/LA30/Cas-Gnn} \bigstrut\\\cline{2-4}
  
  & DANet \cite{zhaoeccv20} & PyTorch & \href{https://github.com/Xiaoqi-Zhao-DLUT/DANet-RGBD-Saliency}{https://github.com/Xiaoqi-Zhao-DLUT/DANet-RGBD-Saliency} \bigstrut\\\cline{2-4}

  & CoNet \cite{Wei_2020_ECCV} & PyTorch & \href{https://github.com/jiwei0921/CoNet}{https://github.com/jiwei0921/CoNet} \bigstrut\\\cline{2-4}   
   
  & DASNet \cite{zhao2020} & PyTorch & \href{http://cvteam.net/projects/2020/DASNet/}{http://cvteam.net/projects/2020/DASNet/} \bigstrut\\\cline{2-4}   

  & BBS-Net \cite{faneccv20} & PyTorch & \href{https://github.com/DengPingFan/BBS-Net}{https://github.com/DengPingFan/BBS-Net}\bigstrut\\\cline{2-4}   

  & ATSA \cite{Zhangeccv20} & PyTorch & \href{https://github.com/sxfduter/ATSA}{https://github.com/sxfduter/ATSA} \bigstrut\\\cline{2-4}   
  
  & PGAR \cite{chen2020progressively} & PyTorch & \href{https://github.com/ShuhanChen/PGAR\_ECCV20}{https://github.com/ShuhanChen/PGAR\_ECCV20} \bigstrut\\\cline{2-4}

  & FRDT \cite{zhangmm2020} & PyTorch & \href{https://github.com/jack-admiral/ACM-MM-FRDT}{https://github.com/jack-admiral/ACM-MM-FRDT} \\ \hline

  \hline\hline
\end{tabular}
    
\end{threeparttable}
\end{table*}

\renewcommand\arraystretch{1.1}
\begin{table*}[t!]
    \centering
    
    \caption{Statistics of nine RGB-D benchmark datasets in terms of year (Year), publication (Pub.), dataset size (Size), number of objects in the images (\#Obj.), type of scene (Types), depth sensor (Sensor), and resolution (Resolution). See \secref{sec:dataset} for more details on each dataset. These datasets can be downloaded from our website: \href{ http://dpfan.net/d3netbenchmark/}{http://dpfan.net/d3netbenchmark/}.
    }

    \scriptsize
    \vspace {-2.5mm}
    \label{tab:03}
    \setlength{\tabcolsep}{4.4pt}
    \begin{tabular}{|p{0.2cm}||p{2.3cm}|p{0.5cm}|p{0.95cm}|p{0.5cm}|p{1.0cm}|p{2.5cm}|p{3.4cm}|p{3.0cm}|}
        \hline
         \# & Dataset & Year & Pub. & Size & \#Obj. & Types& Sensor & Resolution \\ \hline\hline
         1 & \textbf{STERE} \cite{niu2012leveraging} &2012 & CVPR & 1,000 & $\sim${One} &{Internet}         & Stereo camera+sift flow & 
         $[251\sim 1200]\times[222\sim 900]$ \\ \hline
         2 & \textbf{GIT} \cite{ciptadi2013depth}   &2013 & BMVC   & 80 &Multiple &Home environment & Microsoft Kinect        & $640\times 480$ \\ \hline
         3 & \textbf{DES} \cite{cheng2014depth}   &2014 & ICIMCS & 135  &One      &Indoor           & Microsoft Kinect        & $640\times 480$ \\ \hline
         4 & \textbf{NLPR} \cite{peng2014rgbd}    &2014 & ECCV   & 1,000 &Multiple &Indoor/outdoor   & Microsoft Kinect        & $640\times 480$, $480\times 640$ \\ \hline
         5 & \textbf{LFSD} \cite{li2014saliency}  &2014 & CVPR   & 100  &One      &Indoor/outdoor   & Lytro Illum camera      & $360\times 360$ \\ \hline
         6 & \textbf{NJUD} \cite{ju2014depth}  &2014 & ICIP   & 1,985 &$\sim${One} &Movie/internet/photo & FujiW3 camera+optical flow & 
         $[231\sim 1213]\times[274\sim 828]$  \\ \hline
         7 & \textbf{SSD} \cite{zhu2017three}  &2017 & ICCVW  & 80            &Multiple &Movies & Sun’s optical flow        & $960\times 1080$ \\ \hline
         8 & \textbf{DUT-RGBD} \cite{piao2019saliency}  &2019 & ICCV   & 1,200 &Multiple &Indoor/outdoor & --        & $400\times 600$ \\ \hline
         9 & \textbf{SIP} \cite{fan2019rethinking}   &2020 & TNNLS   & 929   &Multiple &Person in the wild & Huawei Mate10        & $992\times 744$ \\ \hline

         \hline
    \end{tabular}
\end{table*}

\section{RGB-D Datasets}\label{sec:Results}
\label{sec:dataset}

With the rapid development of RGB-D based SOD, various datasets have been constructed over the past several years. Tab~\ref{tab:03} summarizes nine popular RGB-D datasets, and Fig.~\ref{fig_03} shows examples of images (including RGB images, depth maps, and annotations) from these datasets. Moreover, we provide the details for each dataset as follows.

$\bullet$ \textbf{STERE} \cite{niu2012leveraging}. The authors first collected 1,250 stereoscopic images from Flickr \footnote{http://www.flickr.com/}, NVIDIA 3D Vision Live \footnote{http://photos.3dvisionlive.com/}, and Stereoscopic Image Gallery \footnote{http://www.stereophotography.com/}. The most salient objects in each image were annotated by three users. All annotated images were then sorted based on the overlaping salient regions and the top 1,000 images were selected to construct the final dataset. This is the first collection of stereoscopic images in this field.

$\bullet$ \textbf{GIT} \cite{ciptadi2013depth} consists of 80 color and depth images, which were collected using a mobile-manipulator robot in a real-world home environment. Moreover, each image is annotated based on the pixel-level segmentation of the objects.

$\bullet$ \textbf{DES} \cite{cheng2014depth} consists of 135 indoor RGB-D images, which were taken by Kinect with a resolution of $640\times{640}$. When collecting this dataset, three users were asked to label the salient object in each image, and then the overlapping areas of the labeled object were regarded as the ground truth.

$\bullet$ \textbf{NLPR} \cite{peng2014rgbd} consists of 1,000 RGB images and their corresponding depth maps, which were obtained by a standard Microsoft Kinect. This dataset includes a series of outdoor and indoor locations, \eg, offices, supermarkets, campuses, streets, and so on.

$\bullet$ \textbf{LFSD} \cite{li2014saliency} includes 100 light fields collected using a Lytro light field camera, and consists of 60 indoor and 40 outdoor scenes. To label this dataset, three individuals were asked to manually segment salient regions, and then the segmented results were deemed ground truth when the overlap of the three results was over $90\%$. 

$\bullet$ \textbf{NJUD} \cite{ju2014depth} consists of 1,985 stereo image pairs, and these images were collected from the internet, 3D movies, and photographs that are taken by a Fuji W3 stereo camera.

$\bullet$ \textbf{SSD} \cite{zhu2017three} was constructed using three stereo movies and includes indoor and outdoor scenes. This dataset includes 80 samples, and each image has the size of $960\times{1080}$.

$\bullet$ \textbf{DUT-RGBD} \cite{piao2019saliency} consists of 800 indoor and 400 outdoor scenes with their corresponding depth images. This dataset includes several challenging factors, \ie, multiple or transparent objects, complex backgrounds, similar foregrounds and backgrounds, and low-intensity environments.

$\bullet$ \textbf{SIP} \cite{fan2019rethinking} consists of 929 annotated high-resolution images, with multiple salient persons in each image. In this dataset, depth maps were captured using a real smartphone (\ie, Huawei Mate10). Besides, it is worth noting that this dataset covers diverse scenes, and various challenging factors, and is annotated with pixel-level ground truths.

Note that a detailed dataset statistics analysis (including center bias, size of objects, background objects, object boundary conditions, and number of salient objects) can be found in \cite{fan2019rethinking}.

\renewcommand\arraystretch{1.0}
\begin{table*}[t!]
    \centering
    
    \caption{Summary of popular light field SOD methods.}
    \scriptsize
    \vspace {-2.5mm}
    \label{tab:031}
    \setlength{\tabcolsep}{4.4pt}
    \begin{tabular}{|p{0.3cm}|p{0.5cm}<{\centering}|r|p{0.8cm}<{\centering}|p{3.1cm}|p{9.0cm}|}
        \hline
         No. & Year & Method & Pub.  & Dataset & Description \\ \hline\hline
    
    1 & 2014 & LFS \cite{li2014saliency} & CVPR  & LFSD  & The first light-field saliency detection algorithm employs objectness and focusness cues based on the refocusing capability of the light field \\ \hline  

    2 & 2015 & WSC \cite{li2015weighted} & CVPR  & LFSD &  Uses a weighted sparse coding framework to learn a saliency/non-saliency dictionary \\ \hline  
    
    3 & 2015 & DILF \cite{zhang2015saliency} & IJCAI  & LFSD & Incorporates depth contrast to complement the disadvantage of color and conducts focusness-based background priors to boost the saliency detection performance \\ \hline  

    4 & 2016 & RL \cite{sheng2016relative} & ICASSP  & LFSD & Utilizes the inherent structure information in light field images to improve saliency detection \\ \hline  
    
    5 & 2017 & MA \cite{zhang2017saliency} & TOMM  & HFUT, LFSD &  Integrates multiple saliency cues extracted from light field images using a random-search-based weighting strategy \\ \hline  
    
    6 & 2017 & BIF \cite{wang2017two} & NPL  & LFSD & Integrates color-based contrast, depth-induced contrast, focusness map of foreground slice, and background weighted depth contrast using a two-stage Bayesian integration framework \\ \hline 
 
    7 & 2017 & LFS \cite{li2017pami} & TPAMI  & LFSD & An extension of \cite{li2014saliency} \\ \hline    

    8 & 2017 & RLM \cite{li2017saliency} & ICIVC  & LFSD & Utilizes the light field relative location measurement for SOD on light field images \\ \hline    
        
    9 & 2018 & SGDC \cite{wang2018salience} & CVPR  & LFSD & Designs a saliency-guided depth optimization framework for multi-layer light field displays \\ \hline    

    10 & 2018 & DCA \cite{piao2018depth} & FiO  & LFSD & Proposes a graph model depth-induced cellular automata to optimize saliency maps using light field data \\ \hline  
    
    11 & 2019 & DLLF \cite{wang2019deep} & ICCV  & DUTLF-FS, LFSD & Utilizes a recurrent attention network to fuse each slice from the focal stack to learn the most informative features  \\ \hline  
    
    12 & 2019 & DLSD \cite{piao2019deep} & IJCAI  & DUTLF-MV & Formulates saliency detection into two subproblems, including 1) light field synthesis from a single view and 2) light-field-driven saliency detection \\ \hline  

    13 & 2019 & Molf \cite{zhang2019memory} & NIPS  & UTLF-FS & Uses a memory-oriented decoder for light field SOD \\ \hline 
    
    14 & 2020 & ERNet \cite{piaoexploit} & AAAI  & DUTLF-FS, HFUT, LFSD & Uses an asymmetrical two-stream architecture to overcome computation-intensive and memory-intensive challenges in a high-dimensional light field data \\ \hline 
 
    15 & 2020 & DCA \cite{piao2019saliency} & TIP  & LFSD & Presents a saliency detection framework on light fields based on the depth-induced cellular automata (DCA) model. It can enforce spatial consistency to optimize the inaccurate saliency map using the DCA model \\ \hline 
 
    16 & 2020 & RDFD \cite{wang2020region} & MTAP  & LFSD & Defines a region-based depth feature descriptor extracted from the light field focal stack to facilitate low- and high-level cues for saliency detection \\ \hline %
    
    17 & 2020 & LFNet \cite{zhang2020lfnet} & TIP  & DUTLF-FS, LFSD, HFUT & Utilizes a light field refinement module and a light field integration module to effectively integrate multiple cues (\ie, focusness, depths and objectness) from light field images \\ \hline %
    
    18 & 2020 & LFDCN \cite{zhang2020light} & TIP  & Lytro Illum, LFSD, HFUT & Uses a deep convolutional network based on the modified DeepLab-v2 model to explore spatial and multi-view properties of light field images for saliency detection \\ \hline %
    
        \hline
    \end{tabular}
\end{table*}

\section{Saliency Detection on Light Field}
\label{sec:lightfield}

\subsection{Light Field SOD Models}
\label{sec:lf_models}

Existing works for SOD can be grouped into three categories according to the input data type, including RGB SOD, RGB-D SOD, and light field SOD~\cite{zhang2020lfnet}. We have already reviewed RGB-D based SOD models, in which depth maps provide layout information to improve SOD performance to some extent. However, inaccurate or low-quality depth maps often decrease the performance. To overcome this issue, light field SOD methods have been proposed to make use of rich information captured by the light field. Specifically, light field data contains an all-focus image, a focal stack, and a rough depth map \cite{piao2019saliency}. A summary of related light field SOD works is provided in Tab~\ref{tab:031}. Further, to provide an in-depth understanding of these models, we also review them in more detail as follows.

\textbf{Traditional/Deep Models}. The classic models for light field SOD often use superpixel-level handcrafted features \cite{li2014saliency,li2017pami,li2015weighted,zhang2015saliency,sheng2016relative,zhang2017saliency,wang2017two,wang2018salience,wang2020region,piao2019saliency}. Early work \cite{li2014saliency,li2017pami} showed that the unique refocusing capability of light fields can provide useful focusness, depth, and objectness cues. Thus, several SOD models using light field data were further proposed. For example, Zhang \etal~\cite{zhang2015saliency} utilized a set of focal slices to compute the background prior, and then combined it with the location prior for SOD. Wang \etal~\cite{wang2017two} proposed a two-stage Bayesian fusion model to integrate multiple contrasts for boosting SOD performance. Recently, several deep learning-based light field SOD models \cite{wang2019deep,piao2019deep,zhang2019memory,piaoexploit,zhang2020lfnet,zhang2020light} have also been developed, obtaining remarkable performance. Besides, in \cite{wang2019deep}, an attentive recurrent CNN was developed to fuse all focal slices, while the data diversity was increased using adversarial examples to enhance model robustness. Zhang \etal~\cite{zhang2019memory} developed a memory-oriented decoder for light field SOD, which fuses multi-level features in a top-down manner using high-level information to guide low-level feature selection. LFNet \cite{zhang2020lfnet} employs a new integration module to fuse features from light field data according to their contributions and captures the spatial structure of a scene to improve SOD performance. 

\textbf{Refinement based Models}. Several refinement strategies have been used to enforce neighboring constraints or reduce the homogeneity of multiple modalities for SOD. For example, in \cite{li2015weighted}, the saliency dictionary was refined using the estimated saliency map. The MA method \cite{zhang2017saliency} employs a two-stage saliency refinement strategy to produce the final prediction map, which enables adjacent superpixels to obtain similar saliency values. Besides, LFNet \cite{zhang2020lfnet} presents an effective refinement module to reduce the homogeneity among different modalities as well refine their dissimilarities

\subsection{Light Field Data for SOD}
\label{sec:lf_daataset}

There are five representative datasets widely used in existing light field SOD models. We describe the details of each dataset as follows.

$\bullet$ \textbf{LFSD} \cite{li2014saliency} \footnote{{https://sites.duke.edu/nianyi/publication/saliency-detection-on-light-field/}} consists of 100 light fields of different scenes with a $360\times{360}$ spatial resolution, captured using a Lytro light field camera. This dataset contains 60 indoor and 40 outdoor scenes, and most scenes consist of only one salient object. Besides, three individuals were asked to manually segment salient regions in each image, and then the ground truth was determined when all three segmentation results had an overlap of over 90\%. 

$\bullet$ \textbf{HFUT} \cite{zhang2017saliency} \footnote{https://github.com/pencilzhang/HFUT-Lytro-dataset} consists of 255 light fields captured using a Lytro camera. In this dataset, most scenes contain multiple objects that appear within different locations and scales under complex background clutter. 

$\bullet$ \textbf{DUTLF-FS} \cite{wang2019deep} \footnote{https://github.com/OIPLab-DUT/ICCV2019\_Deeplightfield\_Saliency} consists of 1,465 samples, 1,000 of which are used as the training set, while the remaining 465 images make up the test set. The resolution of each image is $600\times{400}$. This dataset contains several challenges, including lower contrast between salient objects and cluttered background, multiple disconnected salient objects, and dark or strong light conditions.

$\bullet$ \textbf{DUTLF-MV} \cite{piao2019deep} \footnote{https://github.com/OIPLab-DUT/IJCAI2019-Deep-Light-Field-Driven-Saliency-Detection-from-A-Single-View} consists of 1,580 samples, 1,100 of which are for training and the remaining is for testing. Images were captured by a Lytro Illum camera, and each light field consists of multi-view images and a corresponding ground truth.

$\bullet$ \textbf{Lytro Illum} \cite{zhang2020light} \footnote{https://github.com/pencilzhang/MAC-light-field-saliency-net} consists of 640 light fields and the corresponding per-pixel ground-truth saliency maps. It includes several challenging factors, \eg, inconsistent illumination conditions, and small salient objects existing in a similar or cluttered background.

\begin{figure*}[t]
    \vspace {-4mm}
    \centering
    \includegraphics[width=0.99\linewidth]{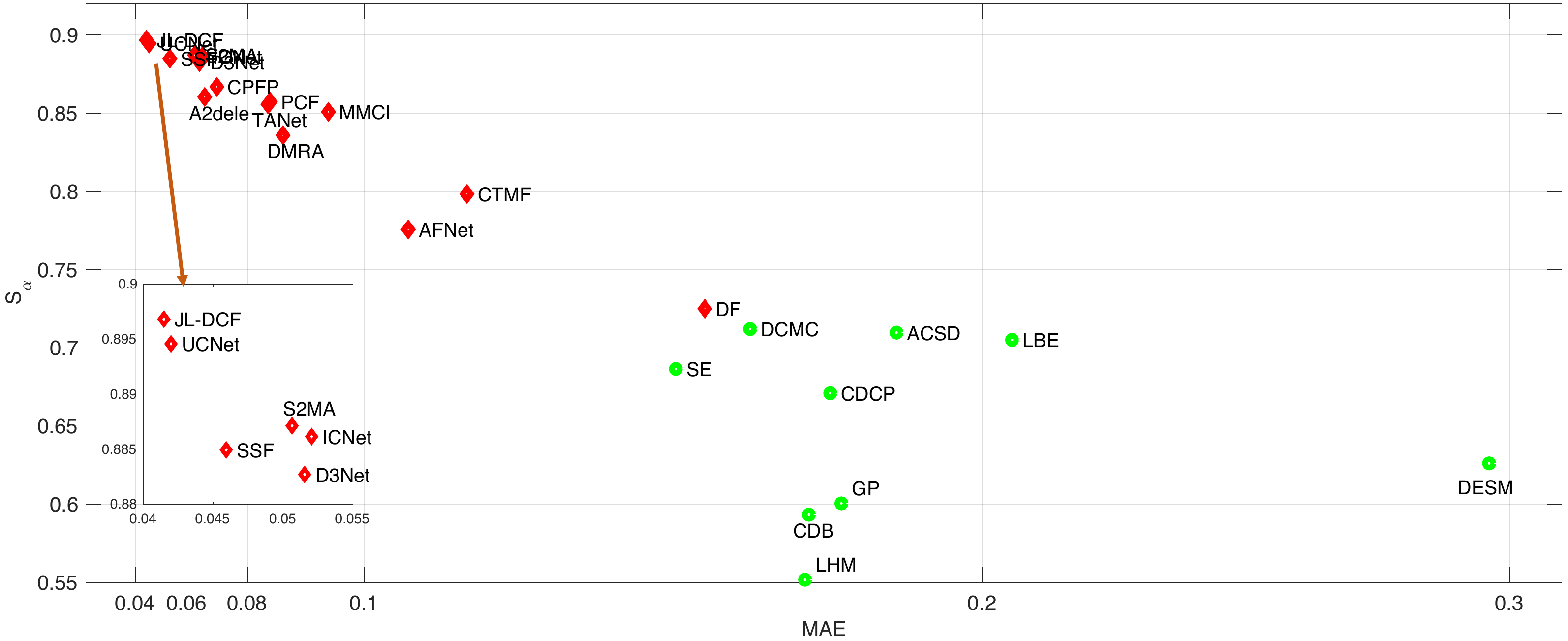} \vspace {-1mm}
    \caption{A comprehensive evaluation for 24 representative RGB-D based SOD models, including LHM \cite{peng2014rgbd}, ACSD \cite{ju2014depth}, DESM \cite{cheng2014depth}, GP \cite{ren2015exploiting}, LBE \cite{feng2016local}, DCMC \cite{cong2016saliency}, SE \cite{guo2016salient}, CDCP \cite{zhu2017innovative}, CDB \cite{liang2018stereoscopic}, DF \cite{qu2017rgbd}, PCF \cite{chen2018progressively}, CTMF \cite{han2017cnns}, CPFP \cite{zhao2019contrast}, TANet \cite{chen2019three}, AFNet \cite{wang2019adaptive}, MMCI \cite{chen2019multi}, DMRA \cite{piao2019depth}, D$^3$Net \cite{fan2019rethinking}, SSF \cite{zhang2020}, A2dele \cite{piao2020}, S$^2$MA \cite{liu2020}, ICNet \cite{li2020icnet}, JL-DCF \cite{fu2020jl}, and UC-Net \cite{zhang2020uc}. We report the mean values of $S_{\alpha}$ and {MAE} across the five datasets (\ie, STERE \cite{niu2012leveraging}, NLPR \cite{peng2014rgbd}, LFSD \cite{li2014saliency}, DES \cite{cheng2014depth}, and SIP \cite{fan2019rethinking}) in each model. Note that better models are shown in the upper left corner (\ie, with a larger $S_{\alpha}$ and smaller MAE). Here, red diamonds denote deep models and green circles denote traditional models.}  \label{fig_04}

\end{figure*}

\section{Model Evaluation and Analysis}
\label{sec:evaluation}

\subsection{Evaluation Metrics}
\label{sec:metrics}

We briefly review several popular metrics for SOD evaluation, \ie, precision-recall (PR), F-measure \cite{achanta2009frequency,borji2015salient}, mean absolute error (MAE) \cite{perazzi2012saliency}, structural measure (S-measure) \cite{fan2017structure}, and enhanced-alignment measure (E-measure) \cite{Fan2018Enhanced}.

$\bullet$ {\textbf{PR}}. Given a saliency map $S$, we can convert it to a binary mask $M$, and then compute the \emph{precision}
and \emph{recall} by comparing $M$ with ground-truth $G$:
\begin{equation}
Precision=\frac{|M\cap{G}|}{|M|},~ Recall=\frac{|M\cap{G}|}{|G|}.
\end{equation}

A popular strategy is to partition the saliency map $S$ using a set of thresholds (\ie, it changes from 0 to 255). For each threshold, we first calculate a pair of recall and precision scores, and then combine them to obtain a PR curve that describes the performance of the model at the different thresholds.


$\bullet$ {\textbf{F-measure} ($F_{\beta}$)}. To comprehensively consider both precision and recall, the F-measure is proposed by calculating the weighted harmonic mean:
\begin{equation}
F_{\beta}=\left(1+\beta ^2\right)\frac{P*R}{\beta ^{2}P+R},
\end{equation}
where $\beta^2$ is set to 0.3 to emphasize the precision \cite{achanta2009frequency}. We use different fixed $[0,255]$ thresholds to  compute the $F$-measure metric. This yields a set of $F$-measure values for which we report the maximal or average $F_{\beta}$.

$\bullet$ {\textbf{MAE}}. This measures the average pixel-wise absolute error between a predicted saliency map $S$ and a ground truth $G$ for all pixels, which can be defined by
\begin{equation}
MAE=\frac{1}{W*H}\sum_{i=1}^{W}\sum_{i=1}^{H}\left|S_{i,j}-G_{i,j} \right|,
\end{equation}
where $W$ and $H$ denote the width and height of the map, respectively. MAE values are normalized to [0,1].

$\bullet$ {\textbf{S-measure} ($S_{\alpha}$)}. To capture the importance of the structural information in an image, $S_{\alpha}$ \cite{fan2017structure} is used to assess the structural similarity between the regional perception ($S_r$) and object perception ($S_o$). Thus, $S_{\alpha}$ can be defined by
\begin{equation}
S_{\alpha}=\alpha * S_{o}+\left(1-\alpha\right)*S_{r},  
\end{equation}
where $ \alpha \in \left[ 0,1\right]$ is a trade-off parameter. Here, we set $\alpha$ = 0.5 as the default setting, as suggested by Fan \etal~\cite{fan2017structure}. 
	
$\bullet$ {\textbf{E-measure} ($E_{\phi}$)}. $E_{\phi}$ \cite{Fan2018Enhanced} was proposed based on cognitive vision studies to capture image-level statistics and their local pixel matching information. Thus, $E_{\phi}$ can be defined by
\begin{equation}
E_{\phi}=\frac{1}{W*H}\sum_{i=1}^{W}\sum_{i=1}^{H}\phi_{FM}\left(i,j\right), 
\end{equation}
where $ \phi_{FM} $ denotes the enhanced-alignment matrix \cite{Fan2018Enhanced}.

\begin{figure*}[t]
\centering
\begin{overpic}[width=0.99\linewidth]{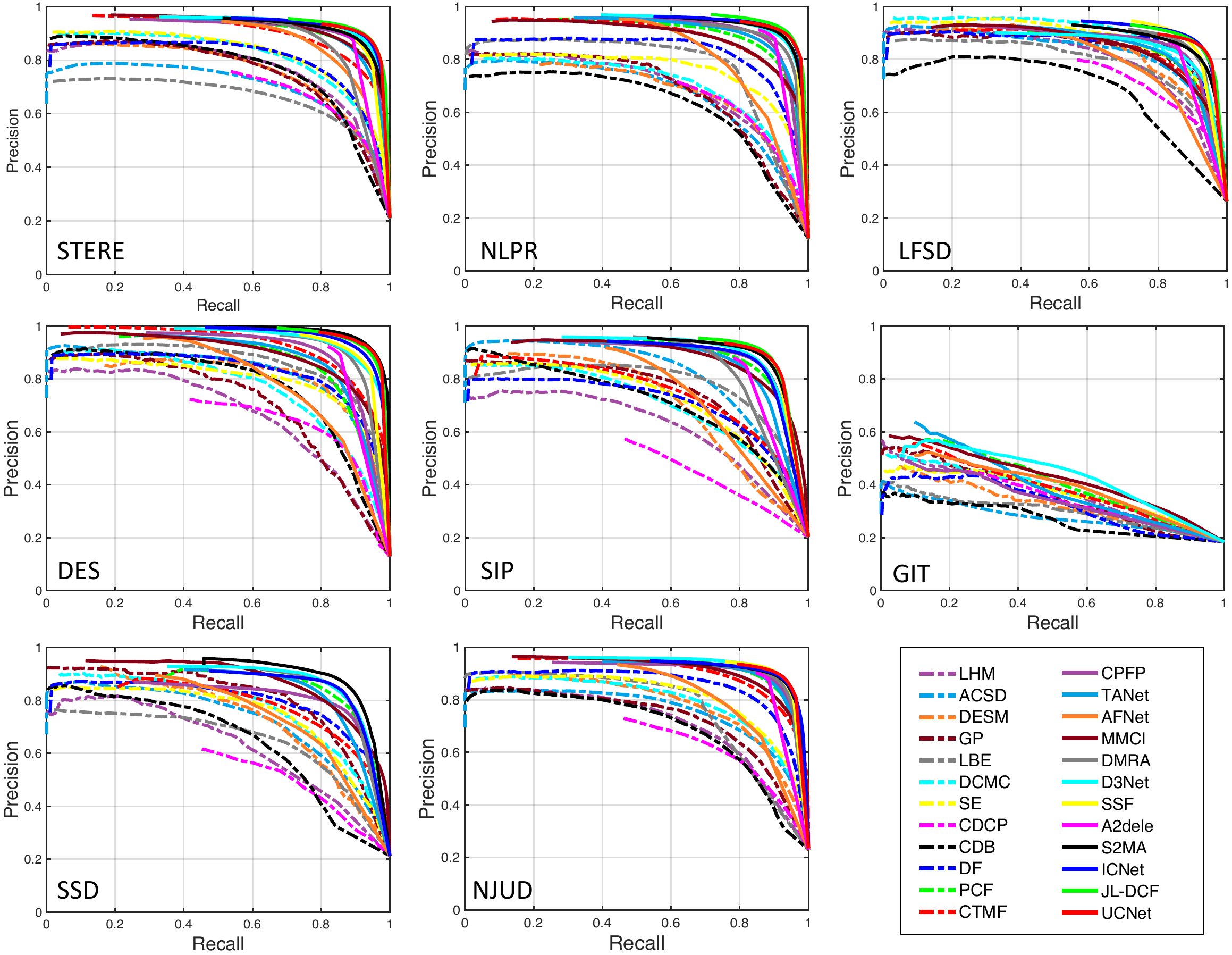}
\end{overpic}\vspace{-0.25cm}

\caption{PR curves for 24 RGB-D based models on the STERE \cite{niu2012leveraging}, NLPR \cite{peng2014rgbd}, LFSD \cite{li2014saliency}, DES \cite{cheng2014depth}, SIP \cite{fan2019rethinking}, GIT \cite{ciptadi2013depth}, SSD \cite{zhu2017three}, and NJUD \cite{ju2014depth} datasets.} 
    \label{fig_05}

\end{figure*}

\begin{figure*}[t]
    \vspace {-4mm}
    \centering
    \includegraphics[width=0.99\linewidth]{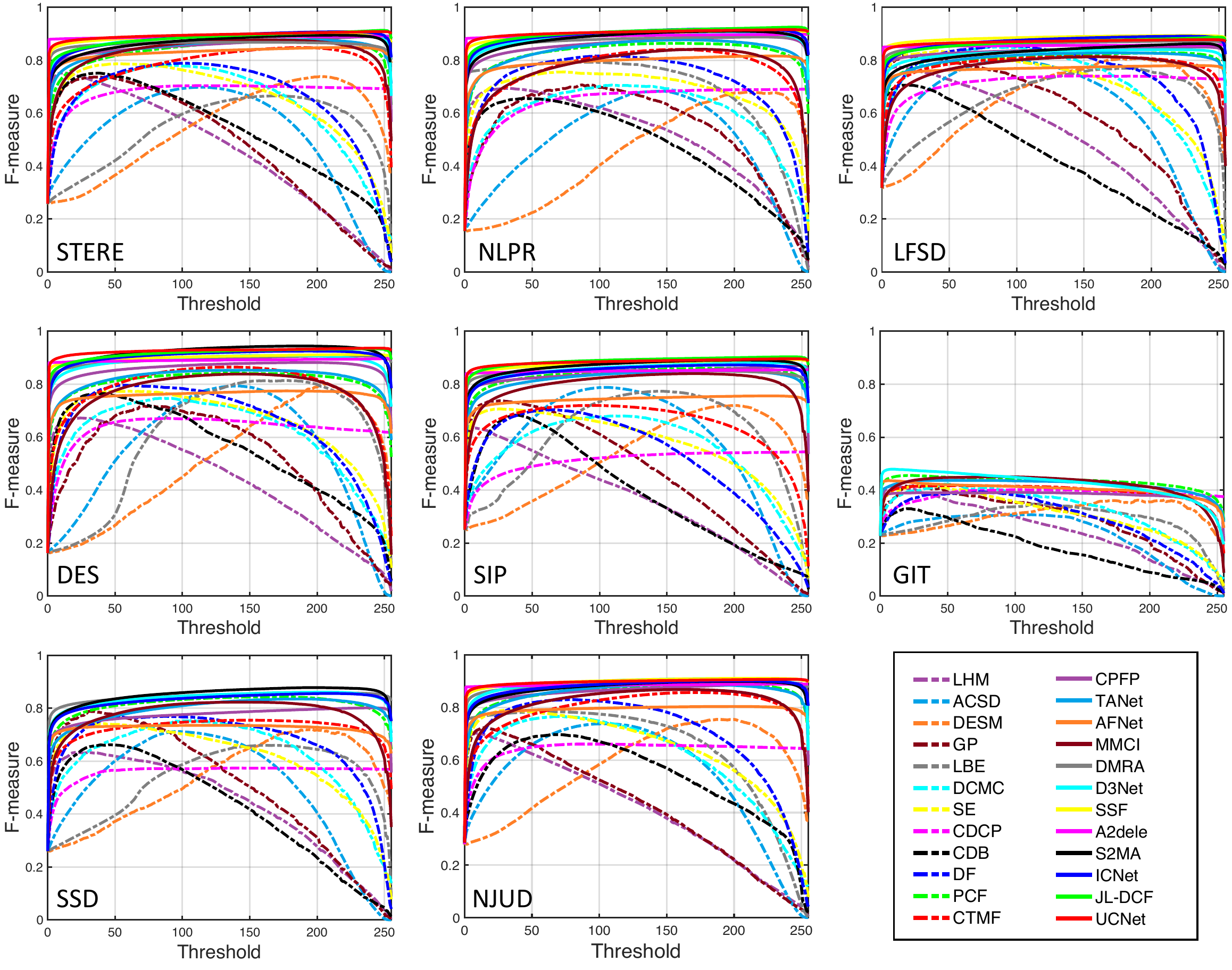} \vspace {-1mm}
    \caption{F-measures under different thresholds for 24 RGB-D based models on the STERE \cite{niu2012leveraging}, NLPR \cite{peng2014rgbd}, LFSD \cite{li2014saliency}, DES \cite{cheng2014depth}, SIP \cite{fan2019rethinking}, GIT \cite{ciptadi2013depth}, SSD \cite{zhu2017three}, and NJUD \cite{ju2014depth} datasets.}  \label{fig_06}

\end{figure*}

\subsection{Performance Comparison and Analysis}
\label{sec:comparison}

\renewcommand\arraystretch{1.1}
\begin{table*}[t!]
    \centering
    
    \caption{Attribute-based study \emph{w.r.t.} salient object scales. Comparison results for 24 representative RGB-D based SOD models (9 traditional models and 15 deep learning-based models) are provided in terms of MAE and $S_{\alpha}$. The three best results are shown in \rev{red}, \blu{blue} and \gre{green} fonts.}
    
    \scriptsize
    \vspace {-2.5mm}
    \label{tab:006}
    \setlength{\tabcolsep}{4.4pt}
    \begin{tabular}{p{0.12cm}<{\centering}|p{0.75cm}<{\centering}|p{0.33cm}<{\centering}|p{0.33cm}<{\centering}|p{0.33cm}<{\centering}|p{0.33cm}<{\centering}|p{0.33cm}<{\centering}|p{0.33cm}<{\centering}|p{0.33cm}<{\centering}|p{0.33cm}<{\centering}|p{0.33cm}<{\centering}|p{0.33cm}<{\centering}|p{0.33cm}<{\centering}|p{0.33cm}<{\centering}|p{0.33cm}<{\centering}|p{0.33cm}<{\centering}|p{0.33cm}<{\centering}|p{0.33cm}<{\centering}|p{0.33cm}<{\centering}|p{0.33cm}<{\centering}|p{0.33cm}<{\centering}|p{0.33cm}<{\centering}|p{0.33cm}<{\centering}|p{0.33cm}<{\centering}|p{0.33cm}<{\centering}|p{0.33cm}<{\centering}}
        \hline
    
     \multirow{2}{*}{}    
     & & \multicolumn{9}{c|}{\textbf{Traditional models} } & \multicolumn{15}{c}{\textbf{Deep learning-based models}}\\ \hline
     &\rotatebox{90}{Scale} 
     &\rotate{LHM \cite{peng2014rgbd}}   &\rotate{ACSD \cite{ju2014depth}}   &\rotate{DESM \cite{cheng2014depth}}       &\rotate{GP \cite{ren2015exploiting}}  
     &\rotate{LBE \cite{feng2016local}}  &\rotate{DCMC \cite{cong2016saliency}} &\rotate{SE \cite{guo2016salient}}      &\rotate{CDCP \cite{zhu2017innovative}} 
     &\rotate{CDB \cite{liang2018stereoscopic}} &\rotate{DF \cite{qu2017rgbd}}      &\rotate{PCF \cite{chen2018progressively}}  
     &\rotate{CTMF \cite{han2017cnns}}   &\rotate{CPFP \cite{zhao2019contrast}}&\rotate{TANet \cite{chen2019three}}     &\rotate{AFNet \cite{wang2019adaptive}} 
     &\rotate{MMCI \cite{chen2019multi}} &\rotate{DMRA \cite{piao2019depth}}   &\rotate{D$^3$Net \cite{fan2019rethinking}}  &\rotate{SSF \cite{zhang2020}}
     &\rotate{A2dele \cite{piao2020}} &\rotate{S$^2$MA \cite{liu2020}}       &\rotate{ICNet \cite{li2020icnet}} &\rotate{JL-DCF \cite{fu2020jl}} &\rotate{UC-Net \cite{zhang2020uc}}  
    \\ \hline\hline

    \multirow{4}{*}{\rotate{MAE}}
    & Small 
    &.065 &.149 &.319 &.098 &.177 &.108 &.056 &.128 &.073 &.087 &.042 &.065 &.044 &.041 &.046 &.051 &\rev{.030} &.033 &\blu{.031} &\gre{.032} &.035 &.036 &\gre{.032} &.034 \\
    & Medium 
    &.178 &.183 &.287 &.180 &.210 &.158 &.150 &.173 &.179 &.152 &.068 &.107 &.055 &.067 &.095 &.079 &.069 &.053 &\gre{.045} &.054 &.052 &.052 &\rev{.041} &\blu{.042} \\
    & Large
    &.403 &.311 &.310 &.377 &.261 &.305 &.364 &.308 &.385 &.310 &.112 &.183 &.093 &.118 &.213 &.130 &.181 &.102 &.105 &.114 &\gre{.088} &.104 &\blu{.085} &\rev{.072} \bigstrut\\\cline{2-26}
    & Overall
    &.166 &.184 &.296 &.173 &.206 &.156 &.142 &.171 &.167 &.147 &.065 &.102 &.055 &.065 &.091 &.076 &.067 &.052 &\gre{.046} &.053 &.051 &.052 &\rev{.041} &\blu{.042} \\ \hline \hline
       
    \multirow{4}{*}{\rotate{$S_{\alpha}$}}
    & Small 
    &.624 &.668 &.517 &.650 &.645 &.700 &.775 &.661 &.666 &.745 &.847 &.789 &.840 &.846 &.792 &.832 &.860 &.879 &.876 &.859 &.877 &\blu{.882} &\gre{.881} &\rev{.883} \\
    & Medium 
    &.543 &.732 &.658 &.598 &.723 &.727 &.676 &.683 &.585 &.730 &.863 &.805 &.877 &.862 &.779 &.859 &.838 &.888 &\gre{.893} &.865 &\gre{.893} &.892 &\rev{.906} &\blu{.901} \\
    & Large
    &.386 &.630 &.686 &.450 &.731 &.604 &.479 &.586 &.424 &.597 &.838 &.761 &.855 &.827 &.682 &.830 &.734 &.846 &.837 &.815 &\blu{.863} &.845 &\gre{.859} &\rev{.876} \bigstrut\\\cline{2-26}
    & Overall
    &.552 &.710 &.626 &.601 &.705 &.712 &.686 &.671 &.593 &.725 &.857 &.798 &.867 &.856 &.776 &.851 &.836 &.883 &.885 &.860 &\gre{.887} &.886 &\rev{.897} &\blu{.895} \\ \hline \hline

    \end{tabular}
\end{table*}

\renewcommand\arraystretch{1.0}
\begin{table*}[t!]
    \centering
    \caption{Attribute-based study \emph{w.r.t.} background clutter. Comparison results for 24 representative RGB-D based SOD models (9 traditional models and 15 deep learning-based models) are provided in terms of MAE and $S_{\alpha}$. The three best results are shown in \rev{red}, \blu{blue} and \gre{green} fonts.}
    
    \scriptsize
    \vspace {-2.5mm}
    \label{tab:007}
    \setlength{\tabcolsep}{4.4pt}
    \begin{tabular}{p{0.12cm}<{\centering}|p{0.98cm}<{\centering}|p{0.33cm}<{\centering}|p{0.33cm}<{\centering}|p{0.33cm}<{\centering}|p{0.33cm}<{\centering}|p{0.33cm}<{\centering}|p{0.33cm}<{\centering}|p{0.33cm}<{\centering}|p{0.33cm}<{\centering}|p{0.33cm}<{\centering}|p{0.33cm}<{\centering}|p{0.33cm}<{\centering}|p{0.33cm}<{\centering}|p{0.33cm}<{\centering}|p{0.33cm}<{\centering}|p{0.33cm}<{\centering}|p{0.33cm}<{\centering}|p{0.33cm}<{\centering}|p{0.33cm}<{\centering}|p{0.33cm}<{\centering}|p{0.33cm}<{\centering}|p{0.33cm}<{\centering}|p{0.33cm}<{\centering}|p{0.33cm}<{\centering}|p{0.33cm}<{\centering}}
        \hline
    
     \multirow{2}{*}{}    
     & & \multicolumn{9}{c|}{\textbf{Traditional models} } & \multicolumn{15}{c}{\textbf{Deep learning-based models}}\\ \hline
     &\rotatebox{90}{background} 
     &\rotate{LHM \cite{peng2014rgbd}}   &\rotate{ACSD \cite{ju2014depth}}   &\rotate{DESM \cite{cheng2014depth}}       &\rotate{GP \cite{ren2015exploiting}}  
     &\rotate{LBE \cite{feng2016local}}  &\rotate{DCMC \cite{cong2016saliency}} &\rotate{SE \cite{guo2016salient}}      &\rotate{CDCP \cite{zhu2017innovative}} 
     &\rotate{CDB \cite{liang2018stereoscopic}} &\rotate{DF \cite{qu2017rgbd}}      &\rotate{PCF \cite{chen2018progressively}}  
     &\rotate{CTMF \cite{han2017cnns}}   &\rotate{CPFP \cite{zhao2019contrast}}&\rotate{TANet \cite{chen2019three}}     &\rotate{AFNet \cite{wang2019adaptive}} 
     &\rotate{MMCI \cite{chen2019multi}} &\rotate{DMRA \cite{piao2019depth}}   &\rotate{D$^3$Net \cite{fan2019rethinking}}  &\rotate{SSF \cite{zhang2020}}
     &\rotate{A2dele \cite{piao2020}} &\rotate{S$^2$MA \cite{liu2020}}       &\rotate{ICNet \cite{li2020icnet}} &\rotate{JL-DCF \cite{fu2020jl}} &\rotate{UC-Net \cite{zhang2020uc}}  
    \\ \hline\hline

    \multirow{4}{*}{\rotate{MAE}}
    & Simple 
    &.100 &.163 &.219 &.150 &.202 &.056 &.084 &.028 &.136 &.045 &.031 &.053 &.018 &.033 &.031 &.041 &.028 &.017 &\blu{.012} &\rev{.010} &.016 &\gre{.013} &.014 &\gre{.013}\\
    & Uncertain
    &.164 &.195 &.294 &.175 &.210 &.140 &.133 &.139 &.159 &.129 &.062 &.081 &.050 &.059 &.075 &.070 &.058 &.045 &\gre{.043} &\gre{.043} &.049 &\blu{.041} &\rev{.037} &\rev{.037}\\
    & Complex
    &.159 &.190 &.349 &.180 &.205 &.190 &.147 &.236 &.143 &.163 &.085 &.110 &.079 &.077 &.108 &.094 &.087 &.071 &\blu{.065} &\gre{.070} &.072 &.079 &\rev{.063} &\blu{.065} 
    \bigstrut\\\cline{2-26}
    & Overall
    &.160 &.193 &.295 &.174 &.209 &.140 &.132 &.141 &.157 &.127 &.063 &.082 &.051 &.059 &.076 &.070 &.059 &\gre{.046} &\blu{.043} &\blu{.043} &.049 &\blu{.043} &\rev{.038} &\rev{.038} \\
    \hline \hline
       
    \multirow{4}{*}{\rotate{$S_{\alpha}$}}
    & Simple 
    &.781 &.787 &.761 &.694 &.748 &.930 &.856 &.941 &.704 &.944 &.944 &.913 &.958 &.937 &.922 &.933 &.935 &.960 &\blu{.966} &\gre{.965} &\gre{.965} &\rev{.969} &.961 &.962 \\
    & Uncertain
    &.572 &.694 &.638 &.606 &.695 &.736 &.723 &.727 &.610 &.774 &.873 &.853 &.882 &.873 &.818 &.868 &.854 &.900 &.894 &.884 &.895 &\rev{.910} &\blu{.909} &\gre{.907}\\
    & Complex
    &.496 &.627 &.509 &.545 &.616 &.577 &.605 &.487 &.575 &.627 &.782 &.742 &.787 &.790 &.694 &.768 &.751 &\gre{.822} &.815 &.786 &.813 &.808 &\blu{.829} &\rev{.833}  \bigstrut\\\cline{2-26}
    & Overall
    &.576 &.693 &.633 &.606 &.691 &.732 &.720 &.718 &.612 &.770 &.869 &.847 &.878 &.869 &.813 &.863 &.850 &\blu{.896} &.891 &.879 &\gre{.892} &\rev{.904} &\rev{.904} &\rev{.904} \\ \hline \hline

    \end{tabular}
\end{table*}

\subsubsection{Overall Evaluation}

To quantify the performance of different models, we conduct a comprehensive evaluation of 24 representative RGB-D based SOD models, including 1) nine traditional methods: LHM \cite{peng2014rgbd}, ACSD \cite{ju2014depth}, DESM \cite{cheng2014depth}, GP \cite{ren2015exploiting}, LBE \cite{feng2016local}, DCMC \cite{cong2016saliency}, SE \cite{guo2016salient}, CDCP \cite{zhu2017innovative}, CDB \cite{liang2018stereoscopic}; and 2) fifteen deep learning-based methods: DF \cite{qu2017rgbd}, PCF \cite{chen2018progressively}, CTMF \cite{han2017cnns}, CPFP \cite{zhao2019contrast}, TANet \cite{chen2019three}, AFNet \cite{wang2019adaptive}, MMCI \cite{chen2019multi}, DMRA \cite{piao2019depth}, D$^3$Net \cite{fan2019rethinking}, SSF \cite{zhang2020}, A2dele \cite{piao2020}, S$^2$MA \cite{liu2020}, ICNet \cite{li2020icnet}, JL-DCF \cite{fu2020jl}, and UC-Net \cite{zhang2020uc}. We report the mean values of $S_{\alpha}$ and {MAE} across the five datasets (STERE \cite{niu2012leveraging}, NLPR \cite{peng2014rgbd} , LFSD \cite{li2014saliency}, DES \cite{cheng2014depth}, and SIP \cite{fan2019rethinking}) for each model in Fig.~\ref{fig_04}. It is worth noting that better models are shown in the upper left corner (\ie, with a larger $S_{\alpha}$ and smaller MAE). From Fig.~\ref{fig_04}, we have following observations:
\begin{itemize}
\item \textbf{Traditional vs. Deep Models}. Compared with traditional RGB-D based SOD models, deep learning methods obtain significantly better performance. This confirms the powerful feature learning ability of deep networks. 

\item \textbf{Comparison of Deep Models}. Among the deep learning-based models, D{$^3$}Net \cite{fan2019rethinking}, JL-DCF \cite{fu2020jl}, UC-Net \cite{zhang2020uc}, SSF \cite{zhang2020}, {ICNet} \cite{li2020icnet}, and S${^2}$MA \cite{liu2020} obtain the best performance. 
\end{itemize}

Moreover, Fig.~\ref{fig_05} and Fig.~\ref{fig_06} show the PR and F-measure curves for the 24 representative RGB-D based SOD models on eight datasets (\ie, STERE \cite{niu2012leveraging}, NLPR \cite{peng2014rgbd}, LFSD \cite{li2014saliency}, DES \cite{cheng2014depth}, SIP \cite{fan2019rethinking}, GIT \cite{ciptadi2013depth}, SSD \cite{zhu2017three} , and NJUD \cite{ju2014depth}). Note that there are 1000, 300, 100, 135, 929, 80, and 80 test samples for NLPR, LFSD, DES, SIP, GIT, and SSD, respectively. For the NJUD \cite{ju2014depth} dataset, there are 485 test images for CPFP \cite{zhao2019contrast}, S$^2$MA \cite{liu2020}, ICNet \cite{li2020icnet}, JL-DCF \cite{fu2020jl}, and UC-Net \cite{zhang2020uc}, while 498 testing images for all other models. 

To understand the top six models in depth, we discuss their main advantages for the six models below.

$\bullet$ D{$^3$}Net \cite{fan2019rethinking} consists of two key components, \ie, a three-stream feature learning module and a depth depurator unit. In the three-stream feature learning module, there are three subnetworks, \ie, RgbNet, RgbdNet, and DepthNet. The RgbNet and DepthNet are used to learn high-level feature representations for RGB and depth images, respectively, while the RgbdNet is used to learn their fused representations. It is worth noting that this three-stream feature learning module can capture modality-specific information as well as the correlation between modalities. Thus, balancing the two aspects is very important for multi-modal learning and it has helped to improve the SOD performance. Besides, the depth depurator unit acts as a gate to explicitly filter out low-quality depth maps, which several existing methods do not consider the effects. Because low-quality depth maps can inhibit the fusion between RGB images and depth maps, thus the depth depurator unit can ensure effective multi-modal fusion to achieve robust SOD performance.

$\bullet$ In JL-DCF \cite{fu2020jl}, there are two key components, \ie, a joint learning (JL) and a densely-cooperative fusion (DCF). Specifically, the JL module is used to learn robust saliency features, while the DCF module is used for complementary feature discovery. It is worth noting that this method uses a middle-fusion strategy to extract deep hierarchical features from RGB images and depth maps, in which the cross-modal complementarity can be effectively exploited to achieve accurate prediction.

$\bullet$ In UC-Net \cite{zhang2020uc}, instead of producing a single saliency prediction, this model produces multiple predictions by modeling the distribution of the feature output space as a generative model conditioned on RGB-D images. Because each person has some specific preferences in labeling a saliency map, it could fail to capture the stochastic characteristic of saliency while only a single saliency map is produced for an image pair using a deterministic learning pipeline. Thus, the strategy in this model can take into account human uncertainty in saliency annotations. Moreover, considering the fact that depth maps could suffer from noise, directly fusing RGB images and depth maps could cause the network to fit to this noise. Therefore, a depth correction network, designed as an auxiliary component, is proposed to refine depth information with a semantic guided loss. Thus, the above key components are all helpful for improving SOD performance. 

$\bullet$ In SSF \cite{zhang2020}, a complementary interaction module (CIM) is developed to explore discriminative cross-modal complementarities and fuse cross-modal features, where a region-wise attention is introduced to supplement rich boundary information for each modality. Besides, a compensation-aware loss is proposed to improve the network’s confidence for hard samples in unreliable depth maps. Thus, these key components enable the proposed model to effectively explore and establish the complementarity of cross-modal feature representations, while at the same time reducing the negative effects introduced by low-quality depth maps, boosting SOD performance. 

$\bullet$ In {ICNet} \cite{li2020icnet}, an information conversion module is proposed to interactively and adaptively explore the correlations between high-level RGB and depth features. Besides, a cross-modal depth-weighted combination block is introduced to enhance the difference between the RGB and depth features in each level, which ensures that the features are treated differently. It is also worth noting that ICNet exploits the complementarity of cross-modal features, as well as explores the continuity of cross-level features, both of which are helpful for achieving accurate predictions. 

$\bullet$ In S${^2}$MA \cite{liu2020}, a self-mutual attention module (SAM) is proposed to fuse RGB and depth images, integrating self-attention and each other’s attention to propagate context more accurately. The SAM can provide additional complementary information from multi-modal data to improve SOD performance, overcoming the limitations of the original self-attention, which only uses a single modality. Besides, to reduce the low-quality (\eg, noise) effects of depth cues, a selection mechanism is proposed to reweight the mutual attention. This mechanism can filter out unreliable information, resulting in more accurate saliency prediction.

\renewcommand\arraystretch{1.0}
\begin{table*}[t!]
    \centering
    
    \caption{Attribute-based study \emph{w.r.t.} background objects (\ie, car, barrier, flower, grass, road, sign, tree, and other). The comparison methods including 24 representative RGB-D based SOD models (9 traditional models and 15 deep learning-based models) evaluated on the SIP dataset \cite{fan2019rethinking} in terms of MAE and $S_{\alpha}$. The three best results are shown in \rev{red}, \blu{blue} and \gre{green} fonts.}
    
    \scriptsize
    \vspace {-2.5mm}
    \label{tab:010}
    \setlength{\tabcolsep}{4.4pt}
    \begin{tabular}{p{0.12cm}<{\centering}|p{0.75cm}<{\centering}|p{0.33cm}<{\centering}|p{0.33cm}<{\centering}|p{0.33cm}<{\centering}|p{0.33cm}<{\centering}|p{0.33cm}<{\centering}|p{0.33cm}<{\centering}|p{0.33cm}<{\centering}|p{0.33cm}<{\centering}|p{0.33cm}<{\centering}|p{0.33cm}<{\centering}|p{0.33cm}<{\centering}|p{0.33cm}<{\centering}|p{0.33cm}<{\centering}|p{0.33cm}<{\centering}|p{0.33cm}<{\centering}|p{0.33cm}<{\centering}|p{0.33cm}<{\centering}|p{0.33cm}<{\centering}|p{0.33cm}<{\centering}|p{0.33cm}<{\centering}|p{0.33cm}<{\centering}|p{0.33cm}<{\centering}|p{0.33cm}<{\centering}|p{0.33cm}<{\centering}}
        \hline
    
     \multirow{2}{*}{}    
     & & \multicolumn{9}{c|}{\textbf{Traditional models}} & \multicolumn{15}{c}{\textbf{Deep learning-based models}}\\ \hline
     &\rotatebox{90}{Categories} 
     &\rotate{LHM \cite{peng2014rgbd}}   &\rotate{ACSD \cite{ju2014depth}}   &\rotate{DESM \cite{cheng2014depth}}       &\rotate{GP \cite{ren2015exploiting}}  
     &\rotate{LBE \cite{feng2016local}}  &\rotate{DCMC \cite{cong2016saliency}} &\rotate{SE \cite{guo2016salient}}      &\rotate{CDCP \cite{zhu2017innovative}} 
     &\rotate{CDB \cite{liang2018stereoscopic}} &\rotate{DF \cite{qu2017rgbd}}      &\rotate{PCF \cite{chen2018progressively}}  
     &\rotate{CTMF \cite{han2017cnns}}   &\rotate{CPFP \cite{zhao2019contrast}}&\rotate{TANet \cite{chen2019three}}     &\rotate{AFNet \cite{wang2019adaptive}} 
     &\rotate{MMCI \cite{chen2019multi}} &\rotate{DMRA \cite{piao2019depth}}   &\rotate{D$^3$Net \cite{fan2019rethinking}}  &\rotate{SSF \cite{zhang2020}}
     &\rotate{A2dele \cite{piao2020}} &\rotate{S$^2$MA \cite{liu2020}}       &\rotate{ICNet \cite{li2020icnet}} &\rotate{JL-DCF \cite{fu2020jl}} &\rotate{UC-Net \cite{zhang2020uc}}  
    \\ \hline\hline

    \multirow{9}{*}{\rotate{{MAE}}}
    & Car
    &.158 &.163 &.301 &.159 &.201 &.185 &.154 &.202 &.171 &.171 &.085 &.134 &.094 &.084 &.101 &.093 &.069 &.061 &.063 &.078 &\rev{.055} &.067 &\gre{.058} &\blu{.057} \\
    & Barrier
    &.197 &.177 &.308 &.180 &.201 &.196 &.176 &.251 &.203 &.202 &.073 &.149 &.060 &.078 &.128 &.089 &.093 &.068 &\gre{.054} &.074 &.057 &.075 &\rev{.052} &\blu{.053} \\
    & Flower
    &.105 &.122 &.306 &.099 &.186 &.158 &.063 &.141 &.101 &.132 &.091 &.075 &.133 &.100 &.090 &.081 &\blu{.046} &.095 &.107 &.051 &.104 &\rev{.025} &\gre{.054} &.075 \\
    & Grass
    &.164 &.161 &.279 &.155 &.184 &.167 &.138 &.182 &.176 &.167 &.041 &.110 &.035 &.048 &.088 &.059 &.056 &.037 &\gre{.030} &.046 &.033 &.043 &\rev{.023} &\blu{.029} \\
    & Road
    &.189 &.167 &.281 &.176 &.187 &.181 &.164 &.225 &.189 &.169 &.070 &.140 &.054 &.072 &.125 &.078 &.093 &.059 &\gre{.049} &.072 &.050 &.065 &\blu{.045} &\rev{.044}\\
    & Sign
    &.107 &.126 &.268 &.110 &.184 &.126 &.079 &.134 &.118 &.096 &.058 &.101 &.063 &.060 &.077 &.083 &.051 &.055 &\gre{.051} &.054 &\rev{.048} &.054 &\blu{.050} &.057 \\
    & Tree
    &.192 &.193 &.310 &.190 &.241 &.194 &.183 &.230 &.219 &.205 &.083 &.157 &.083 &.091 &.132 &.109 &.106 &.083 &\blu{.067} &.074 &.092 &.097 &\rev{.063} &\gre{.071}\\
    & Other
    &.246 &.217 &.329 &.224 &.229 &.216 &.229 &.274 &.233 &.233 &.106 &.177 &.111 &.111 &.170 &.124 &.140 &.095 &\rev{.083} &.099 &.100 &.100 &\blu{.084} &\gre{.086} \bigstrut\\\cline{2-26}
    & Overall
    &.184 &.172 &.298 &.173 &.200 &.186 &.164 &.224 &.192 &.185 &.071 &.139 &.064 &.075 &.118 &.086 &.085 &.063 &\gre{.053} &.070 &.057 &.069 &\rev{.049} &\blu{.051} \\ \hline\hline

\multirow{9}{*}{\rotate{{$S_{\alpha}$}}}
    & Car
    &.516 &.731 &.590 &.603 &.714 &.671 &.591 &.613 &.546 &.631 &.811 &.726 &.786 &.807 &.736 &.813 &.817 &\gre{.856} &.845 &.804 &\rev{.870} &.846 &.855 &\blu{.859} \\
    & Barrier
    &.497 &.727 &.609 &.575 &.728 &.672 &.612 &.553 &.552 &.643 &.837 &.698 &.860 &.831 &.708 &.830 &.792 &.855 &.874 &.821 &\gre{.871} &.848 &\rev{.876} &\blu{.875} \\
    & Flower
    &.477 &.775 &.573 &.673 &.703 &.707 &.772 &.667 &.639 &.750 &.771 &.738 &.714 &.760 &.688 &.785 &.824 &.789 &.768 &\gre{.845} &.804 &\rev{.901} &\blu{.856} &.811 \\
    & Grass
    &.537 &.756 &.643 &.605 &.760 &.728 &.683 &.672 &.559 &.672 &.908 &.770 &.908 &.899 &.780 &.888 &.876 &.917 &\gre{.924} &.878 &\blu{.928} &.910 &\rev{.939} &\gre{.924} \\
    & Road
    &.521 &.739 &.634 &.598 &.751 &.685 &.641 &.595 &.576 &.680 &.851 &.722 &.871 &.848 &.705 &.847 &.807 &.873 &\gre{.885} &.832 &\gre{.885} &.868 &\blu{.889} &\rev{.892} \\
    & Sign
    &.578 &.786 &.634 &.628 &.719 &.745 &.761 &.714 &.615 &.757 &.855 &.756 &.833 &.857 &.771 &.818 &.848 &.849 &.849 &.842 &\rev{.871} &\blu{.861} &\gre{.859} &.840 \\
    & Tree
    &.505 &.699 &.606 &.577 &.661 &.648 &.600 &.588 &.543 &.625 &.802 &.679 &.804 &.778 &.691 &.779 &.748 &.806 &\blu{.837} &.807 &.800 &.788 &\rev{.848} &\gre{.825}\\
    & Other
    &.460 &.687 &.594 &.532 &.706 &.669 &.563 &.554 &.542 &.600 &.786 &.677 &.774 &.782 &.647 &.790 &.722 &.800 &\rev{.828} &.785 &.809 &.799 &\gre{.821} &\blu{.823} \bigstrut\\\cline{2-26}
    & Overall
    &.511 &.732 &.616 &.588 &.727 &.683 &.628 &.595 &.557 &.653 &.842 &.716 &.850 &.835 &.720 &.833 &.806 &.860 &\gre{.874} &.828 &.872 &.854 &\rev{.880} &\blu{.875} \\\hline\hline

    \end{tabular}
\end{table*}

\subsubsection{Attribute-based Evaluation}

To investigate the influence of different factors, such as object scale, background clutter, number of salient objects, indoor or outdoor scene, background objects, and lighting conditions, we carry out diverse attribute-based evaluations on several representative RGB-D based SOD models.

\begin{figure}[t]
    \centering
    \includegraphics[width=.9\linewidth]{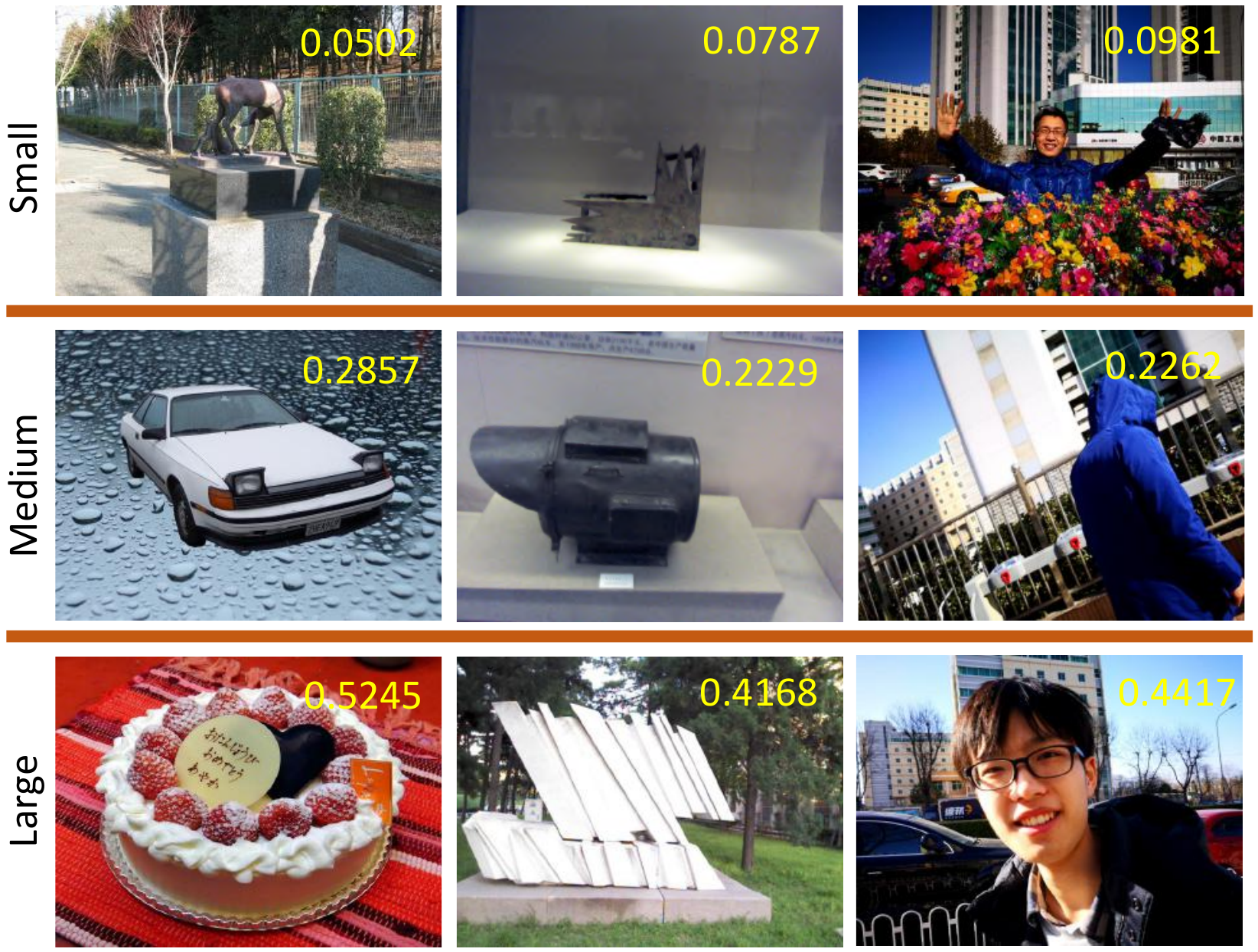} 
    \caption{Sample images with different objects scales. The scale ratios are denoted in yellow.}  \label{fig_0072}
\end{figure}

$\bullet$ \textbf{Object Scale}. To characterize the scale of a salient object area, we compute the ratio between the size of the salient area and the whole image. We define three types of object scales: 1) when the ratio is less than 0.1, it is denoted as ``small"; 2) when the ratio is larger than 0.4, it is denoted as ``large"; and 3) when the ratio is in the range of $[0.1, 0.4]$, it is denoted as ``medium". In this evaluation, we build a hybrid dataset with 2,464 images collected from STERE \cite{niu2012leveraging}, NLPR \cite{peng2014rgbd} , LFSD \cite{li2014saliency}, DES \cite{cheng2014depth}, and SIP \cite{fan2019rethinking}, where 24\%, 69.2\% and 6.8\% of images have small, medium, and large salient object areas, respectively. The constructed hybrid dataset can be found at \href{https://github.com/taozh2017/RGBD-SODsurvey}{https://github.com/taozh2017/RGBD-SODsurvey}. Some sample images with different object scales are shown in Fig.~\ref{fig_0072}. The comparison results of the attribute-based study \emph{w.r.t.} object scale are shown in Tab.~\ref{tab:006}. From the results, it can be observed that all comparison methods obtain better performance in detecting small salient objects while they obtain worse performance in detecting large salient objects. Besides, the three most recent models, \ie, JL-DCF \cite{fu2020jl}, UC-Net \cite{zhang2020uc}, and S$^2$MA \cite{liu2020}, obtain the best performance. D$^3$Net \cite{fan2019rethinking}, SSF \cite{zhang2020}, A2dele \cite{piao2020}, and ICNet \cite{li2020icnet} also obtain promising performance.

\begin{figure}[t]
    \centering
    \includegraphics[width=.9\linewidth]{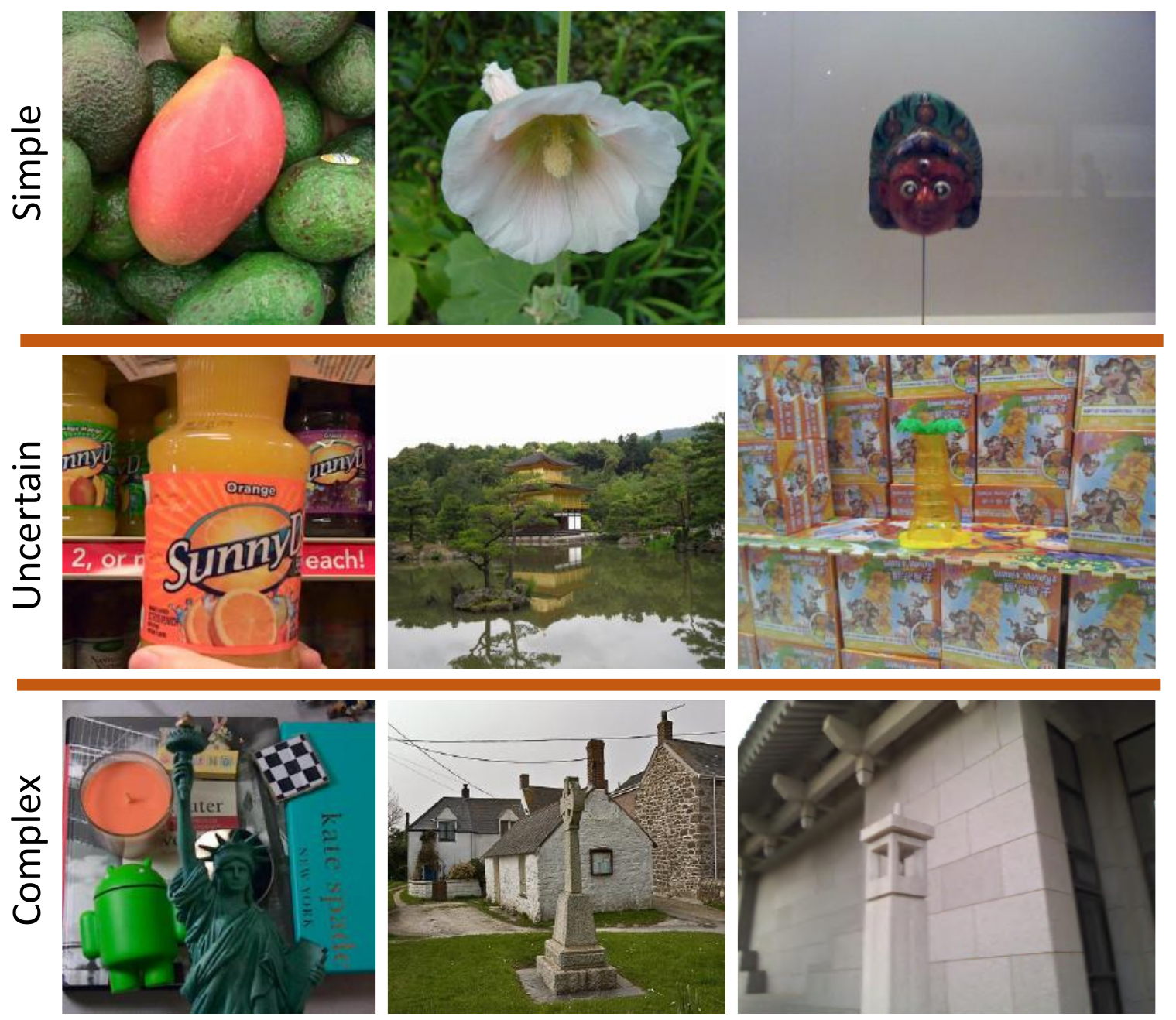} 
    \caption{Sample images with three types of background clutter.
    }  \label{fig_007}
\end{figure}

\begin{figure}[t]
    \centering
    \includegraphics[width=.9\linewidth]{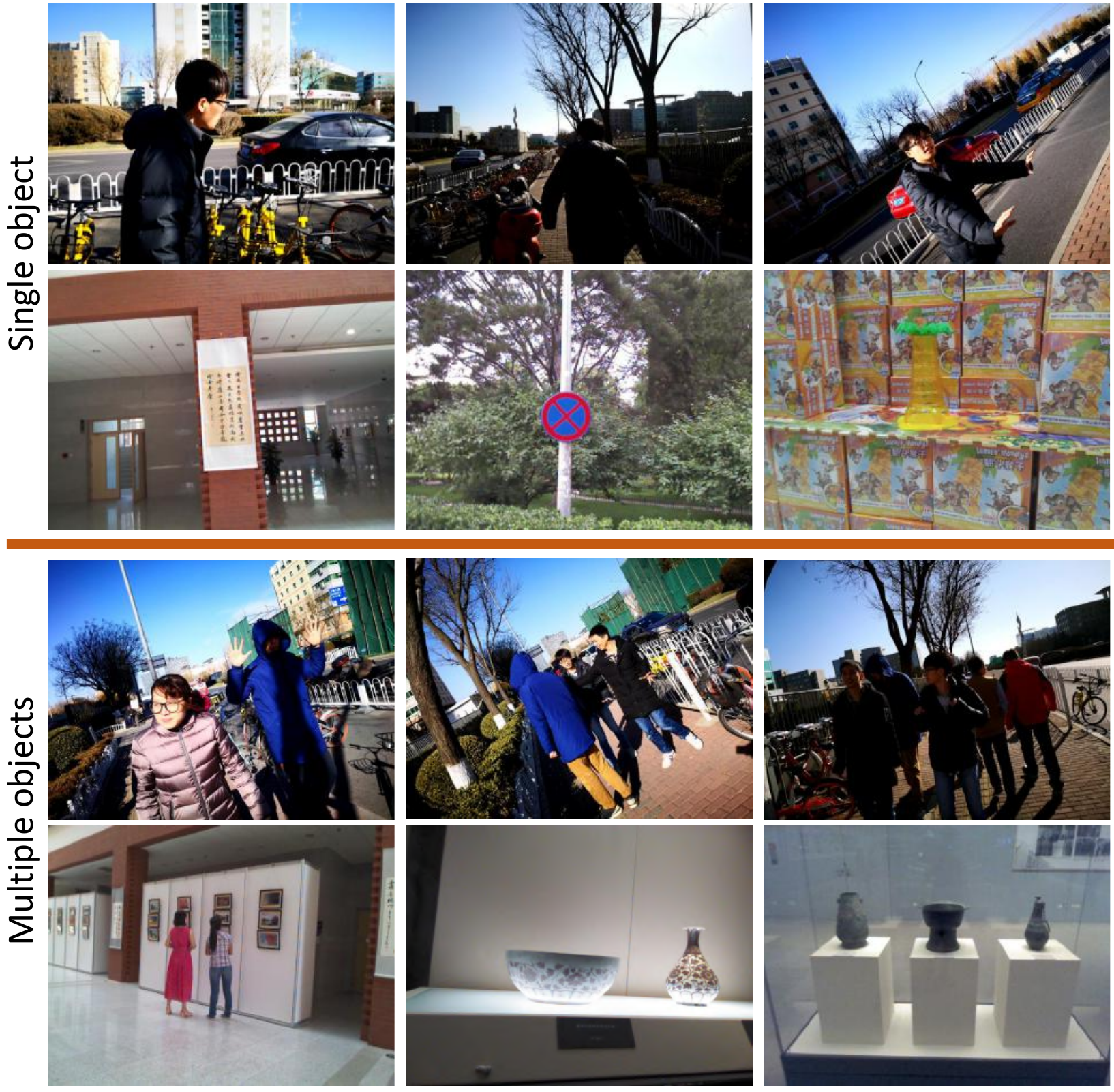} 
    \caption{Sample images with single or multiple salient objects.}  \label{fig_0071}
\end{figure}

\begin{figure*}[t]
    \vspace {-4mm}
    \centering
    \includegraphics[width=1.0\linewidth]{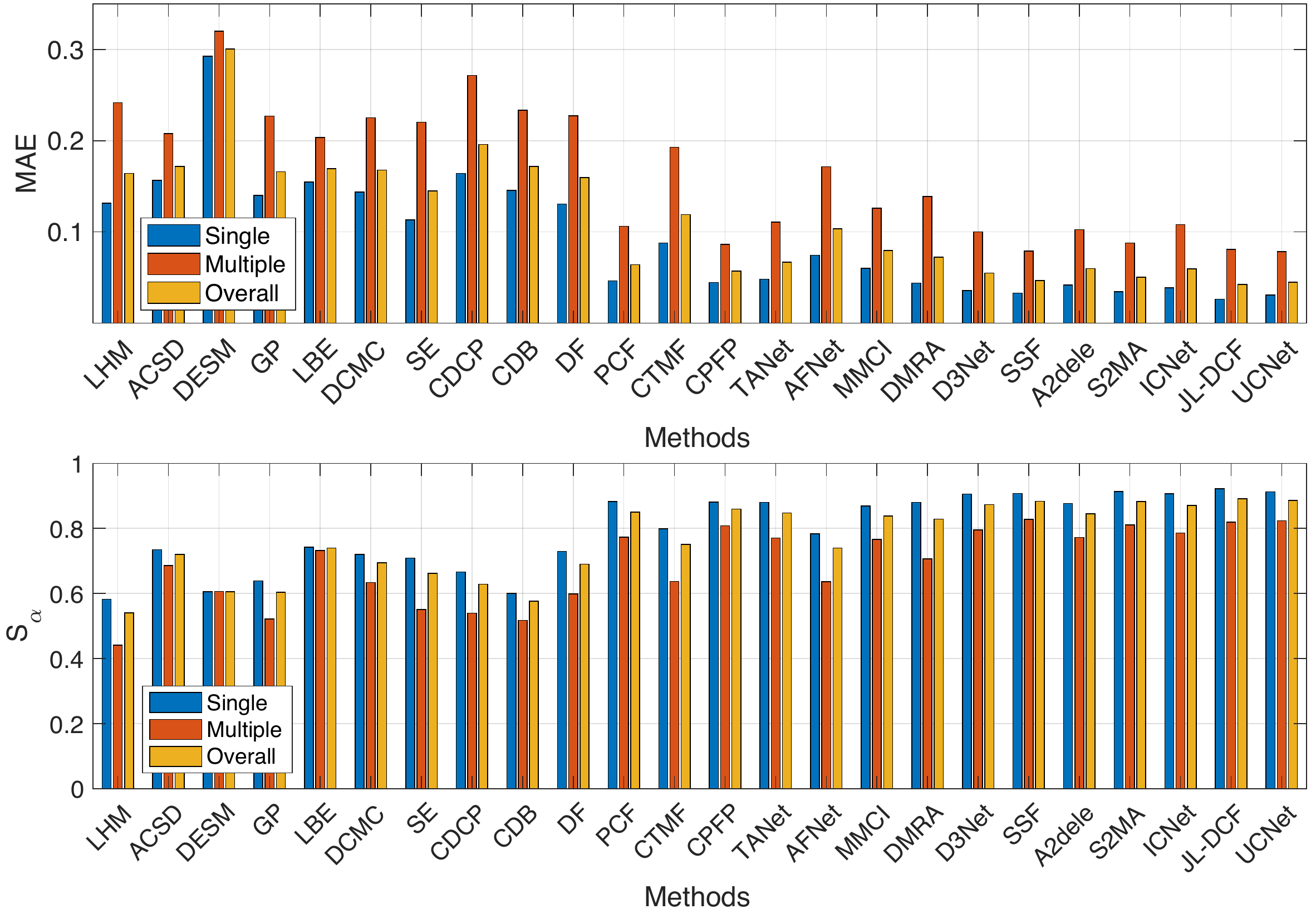} \vspace {-0.45cm}
    \caption{Attribute-based study \emph{w.r.t.} number of salient objects (\ie, single vs. multiple (multi)). The comparison results on 24 representative RGB-D based SOD models (\ie, LHM \cite{peng2014rgbd}, ACSD \cite{ju2014depth}, DESM \cite{cheng2014depth}, GP \cite{ren2015exploiting}, LBE \cite{feng2016local}, DCMC \cite{cong2016saliency}, SE \cite{guo2016salient}, CDCP \cite{zhu2017innovative}, CDB \cite{liang2018stereoscopic}, DF \cite{qu2017rgbd}, PCF \cite{chen2018progressively}, CTMF \cite{han2017cnns}, CPFP \cite{zhao2019contrast}, TANet \cite{chen2019three}, AFNet \cite{wang2019adaptive}, MMCI \cite{chen2019multi}, DMRA \cite{piao2019depth}, D$^3$Net \cite{fan2019rethinking}, SSF \cite{zhang2020}, A2dele \cite{piao2020}, S$^2$MA \cite{liu2020}, ICNet \cite{li2020icnet}, JL-DCF \cite{fu2020jl}, and UC-Net \cite{zhang2020uc}) are given in terms of MAE (top) and $S_{\alpha}$ (bottom).}  \label{fig_08}
\end{figure*}

\begin{figure*}[t]
    \vspace {-4mm}
    \centering
    \includegraphics[width=1.0\linewidth]{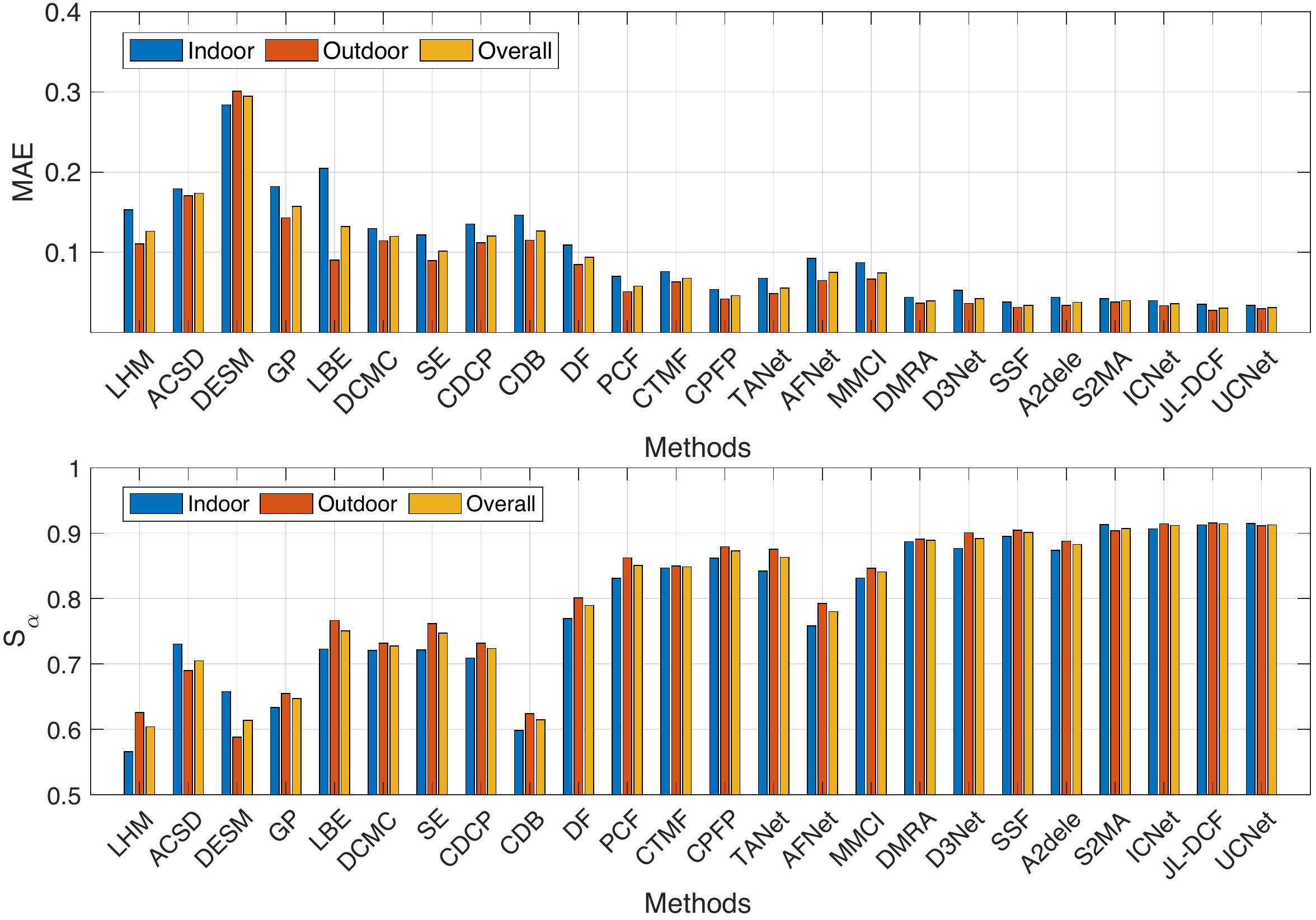} \vspace {-0.45cm}
    \caption{Attribute-based study \emph{w.r.t.} indoor vs. outdoor environments. The comparison results for 24 representative RGB-D based SOD models (\ie, LHM \cite{peng2014rgbd}, ACSD \cite{ju2014depth}, DESM \cite{cheng2014depth}, GP \cite{ren2015exploiting}, LBE \cite{feng2016local}, DCMC \cite{cong2016saliency}, SE \cite{guo2016salient}, CDCP \cite{zhu2017innovative}, CDB \cite{liang2018stereoscopic}, DF \cite{qu2017rgbd}, PCF \cite{chen2018progressively}, CTMF \cite{han2017cnns}, CPFP \cite{zhao2019contrast}, TANet \cite{chen2019three}, AFNet \cite{wang2019adaptive}, MMCI \cite{chen2019multi}, DMRA \cite{piao2019depth}, D$^3$Net \cite{fan2019rethinking}, SSF \cite{zhang2020}, A2dele \cite{piao2020}, S$^2$MA \cite{liu2020}, ICNet \cite{li2020icnet}, JL-DCF \cite{fu2020jl}, and UC-Net \cite{zhang2020uc}) are provided in terms of MAE (top) and $S_{\alpha}$ (bottom).}  \label{fig_09}
\end{figure*}

$\bullet$ \textbf{Background Clutter}. It is difficult to directly characterize background clutter. Since classic SOD methods tend to use prior information or color contrast to locate salient objects, they often fail under complex backgrounds. Thus, in this evaluation, we utilize five traditional SOD methods, \ie, BSCA \cite{qin2015saliency}, CLC \cite{zhou2015salient}, MDC \cite{huang2017}, MIL \cite{huang2017salient}, and WFD \cite{huang2018water}, to first detect salient objects in various images and then group these images into different categories (\eg, simple or complex background) according to the results. Specifically, we first construct a hybrid dataset with 1,400 images collected from three datasets (STERE \cite{niu2012leveraging}, NLPR \cite{peng2014rgbd}, and LFSD \cite{li2014saliency}). Then, we apply the five models to this dataset and obtain the $S_{\alpha}$ values for each, which we use to characterize images as follows: 1) If all $S_{\alpha}$ values are higher than $0.9$, the image is denoted as having a ``simple" background; 2) If all $S_{\alpha}$ values are lower than $0.6$, the image is said to have a ``complex" background; 3) The remaining images are denoted as ``uncertain". Some example images with the three types of background clutter are shown in Fig.~\ref{fig_007}. The constructed hybrid dataset can be found at \href{https://github.com/taozh2017/RGBD-SODsurvey}{https://github.com/taozh2017/RGBD-SODsurvey}. The comparison results of the attribute-based study \emph{w.r.t.} background clutter are shown in Tab.~\ref{tab:007}. As can be seen, all models obtain worse SOD performance on images containing complex backgrounds than simple ones. Among the representative models, JL-DCF \cite{fu2020jl}, UC-Net \cite{zhang2020uc} and SSF \cite{zhang2020} achieve the top-three best results. Besides, the four most recent models, \ie, D$^3$Net \cite{fan2019rethinking}, S$^2$MA \cite{liu2020}, A2dele \cite{piao2020}, and ICNet \cite{li2020icnet}, also obtain better performance than the other models.

\renewcommand\arraystretch{1.0}
\begin{table*}[t!]
    \centering
    
    \caption{Attribute-based study \emph{w.r.t.} light conditions (sunny vs. low-light). The comparison methods include 24 representative RGB-D based SOD models (9 traditional models and 15 deep learning-based models) evaluated on the SIP dataset \cite{fan2019rethinking} in terms of MAE and $S_{\alpha}$. The three best results are shown in \rev{red}, \blu{blue} and \gre{green} fonts.}
    
    \scriptsize
    \vspace {-2.5mm}
    \label{tab:011}
    \setlength{\tabcolsep}{4.4pt}
    \begin{tabular}{p{0.12cm}<{\centering}|p{1.1cm}<{\centering}|p{0.33cm}<{\centering}|p{0.33cm}<{\centering}|p{0.33cm}<{\centering}|p{0.33cm}<{\centering}|p{0.33cm}<{\centering}|p{0.33cm}<{\centering}|p{0.33cm}<{\centering}|p{0.33cm}<{\centering}|p{0.33cm}<{\centering}|p{0.33cm}<{\centering}|p{0.33cm}<{\centering}|p{0.33cm}<{\centering}|p{0.33cm}<{\centering}|p{0.33cm}<{\centering}|p{0.33cm}<{\centering}|p{0.33cm}<{\centering}|p{0.33cm}<{\centering}|p{0.33cm}<{\centering}|p{0.33cm}<{\centering}|p{0.31cm}<{\centering}|p{0.31cm}<{\centering}|p{0.31cm}<{\centering}|p{0.31cm}<{\centering}|p{0.31cm}<{\centering}}
        \hline
    
     \multirow{2}{*}{}    
     & & \multicolumn{9}{c|}{\textbf{Traditional models}} & \multicolumn{15}{c}{\textbf{Deep learning-based models}}\\ \hline
     &\rotatebox{90}{Conditions} 
     &\rotate{LHM \cite{peng2014rgbd}}   &\rotate{ACSD \cite{ju2014depth}}   &\rotate{DESM \cite{cheng2014depth}}       &\rotate{GP \cite{ren2015exploiting}}  
     &\rotate{LBE \cite{feng2016local}}  &\rotate{DCMC \cite{cong2016saliency}} &\rotate{SE \cite{guo2016salient}}      &\rotate{CDCP \cite{zhu2017innovative}} 
     &\rotate{CDB \cite{liang2018stereoscopic}} &\rotate{DF \cite{qu2017rgbd}}      &\rotate{PCF \cite{chen2018progressively}}  
     &\rotate{CTMF \cite{han2017cnns}}   &\rotate{CPFP \cite{zhao2019contrast}}&\rotate{TANet \cite{chen2019three}}     &\rotate{AFNet \cite{wang2019adaptive}} 
     &\rotate{MMCI \cite{chen2019multi}} &\rotate{DMRA \cite{piao2019depth}}   &\rotate{D$^3$Net \cite{fan2019rethinking}}  &\rotate{SSF \cite{zhang2020}}
     &\rotate{A2dele \cite{piao2020}} &\rotate{S$^2$MA \cite{liu2020}}       &\rotate{ICNet \cite{li2020icnet}} &\rotate{JL-DCF \cite{fu2020jl}} &\rotate{UC-Net \cite{zhang2020uc}}  
    \\ \hline\hline

    \multirow{4}{*}{\rotate{{MAE}}}
    & Sunny 
    &.182 &.171 &.294 &.171 &.200 &.183 &.160 &.218 &.190 &.181 &.069 &.137 &.062 &.075 &.116 &.085 &.083 &.062 &\gre{.052} &.068 &.057 &.068 &\rev{.048} &\blu{.051} \\
    & Low-light
    &.198 &.178 &.323 &.187 &.201 &.207 &.193 &.268 &.208 &.211 &.078 &.154 &.073 &.076 &.130 &.091 &.103 &.067 &\gre{.059} &.080 &\blu{.058} &.081 &\gre{.059} &\rev{.055} \\
    & Overall
    &.184 &.172 &.298 &.173 &.200 &.186 &.164 &.224 &.192 &.185 &.071 &.139 &.064 &.075 &.118 &.086 &.085 &.063 &\gre{.053} &.070 &.057 &.069 &\rev{.049} &\blu{.051}\\ \hline\hline

    \multirow{4}{*}{\rotate{$S_{\alpha}$}}
    & Sunny 
    &.516 &.733 &.622 &.593 &.728 &.690 &.639 &.607 &.560 &.660 &.843 &.718 &.852 &.834 &.723 &.833 &.811 &.861 &\gre{.875} &.831 &.872 &.856 &\rev{.882} &\rev{.876} \\
    & low-light
    &.481 &.721 &.573 &.554 &.722 &.635 &.556 &.515 &.543 &.610 &.838 &.701 &.838 &.837 &.700 &.832 &.775 &\gre{.855} &\blu{.867} &.810 &\rev{.871} &.839 &\blu{.867} &\rev{.871} \\
    & Overall
    &.511 &.732 &.616 &.588 &.727 &.683 &.628 &.595 &.557 &.653 &.842 &.716 &.850 &.835 &.720 &.833 &.806 &.860 &\gre{.874} &.828 &.872 &.854 &\rev{.880} &\blu{.875} \\ \hline\hline

    \end{tabular}
\end{table*} 

$\bullet$ \textbf{Single vs. Multiple Objects}. In this evaluation, we construct a hybrid dataset with 1,229 images collected from the NLPR  \cite{peng2014rgbd} and SIP \cite{fan2019rethinking} datasets. Some example images  with single or multiple salient objects are shown  in Fig.~\ref{fig_0071}. The comparison results are shown in Fig.~\ref{fig_08}. From the results, we can see that it is easier to detect single salient object than multiple ones. 

\begin{figure*}[t]
    \vspace {-4mm}
    \centering
    \includegraphics[width=.98\linewidth]{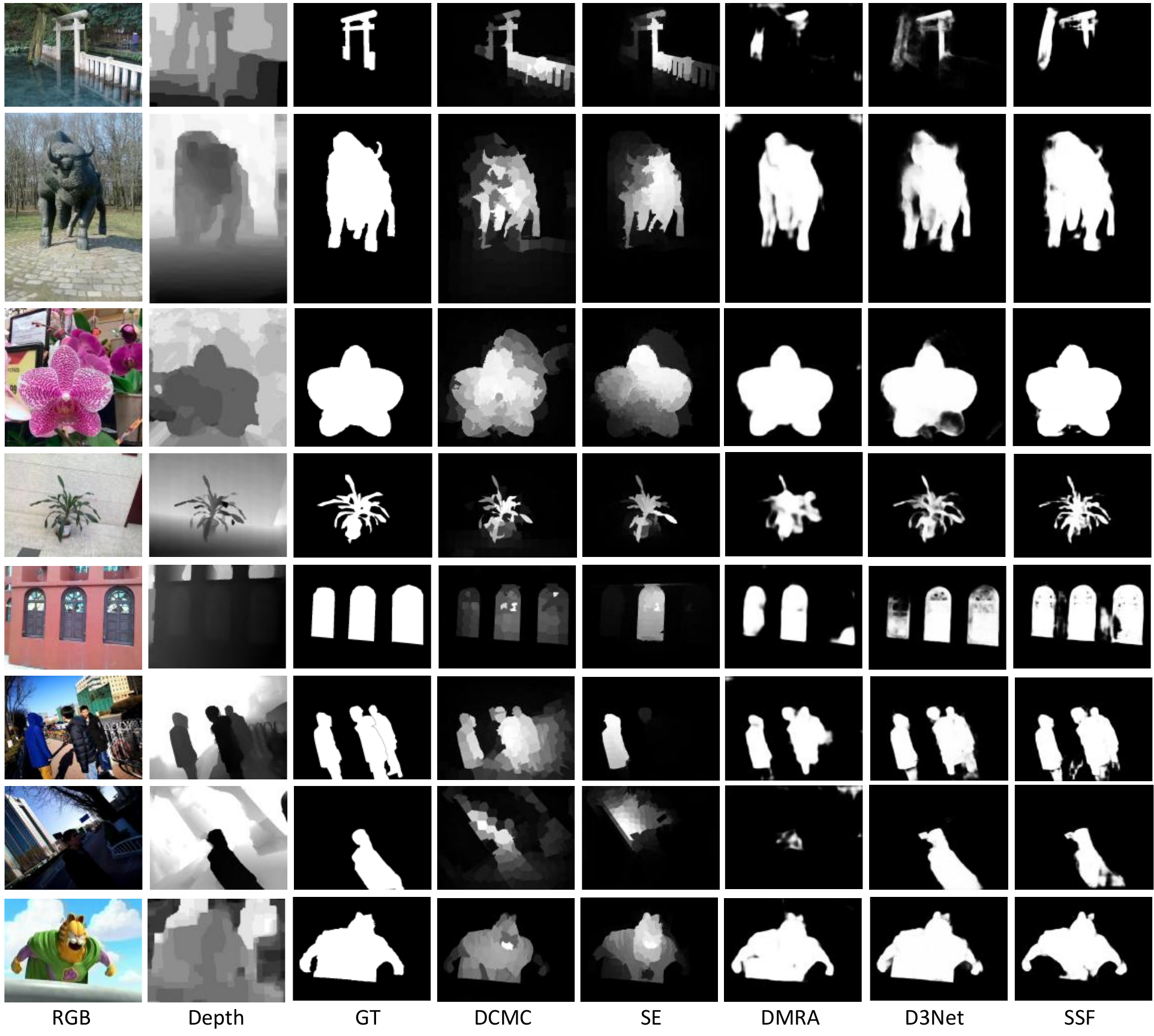} \vspace {-0.15cm}
    \caption{Visual comparisons for two classical non-deep methods (DCMC \cite{cong2016saliency} and SE \cite{guo2016salient}) and three state-of-the-art CNN-based models (DMRA \cite{piao2019depth}, D$^3$Net \cite{fan2019rethinking}, SSF \cite{zhang2020}).}  \label{fig_010}
\end{figure*}

\begin{figure*}[t]
    \vspace {-4mm}
    \centering
    \includegraphics[width=\linewidth]{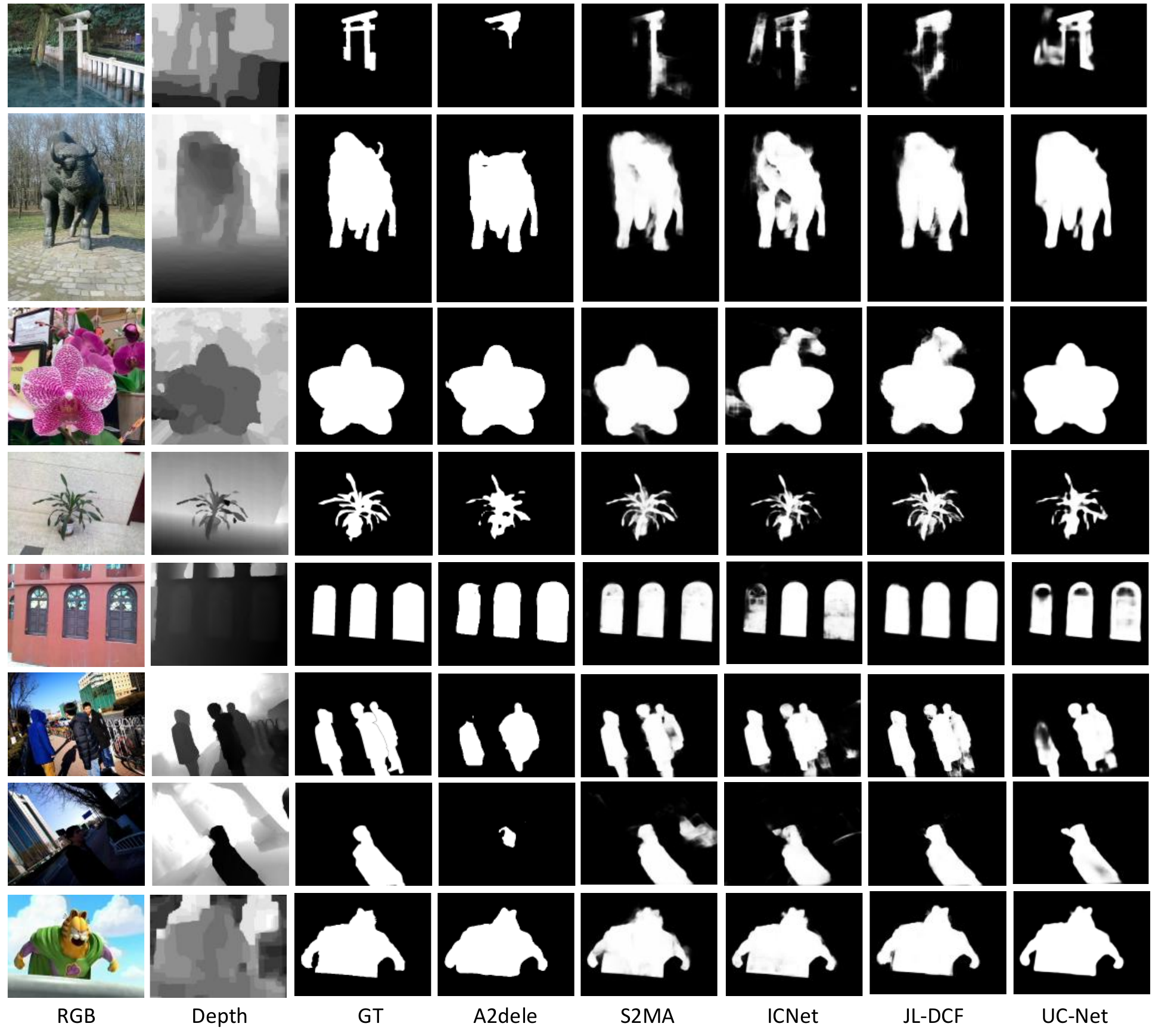} \vspace {-0.45cm}
    \caption{Visual comparisons for five state-of-the-art CNN-based models (A2dele \cite{piao2020}, S$^2$MA \cite{liu2020}, ICNet \cite{li2020icnet}, JL-DCF \cite{fu2020jl}, and UC-Net \cite{zhang2020uc}).}  \label{fig_011}
\end{figure*}

$\bullet$ \textbf{Indoor vs. Outdoor}. We evaluate the performance of different RGB-D based SOD models on indoor and outdoor scenes. In this evaluation, we construct a hybrid dataset collected from the DES \cite{cheng2014depth}, NLPR \cite{peng2014rgbd}, and LFSD \cite{li2014saliency} datasets. The comparison results are shown in Fig.~\ref{fig_09}. From the results, it can be seen that most models struggle more to detect salient objects in indoor scene than outdoor scenes. This is possibly because indoor environments often have varying light conditions. 

$\bullet$ \textbf{Background Objects}. We evaluate the performance of the RGB-D based SOD models when different background objects are present.  We use the SIP dataset \cite{fan2019rethinking}, and split it into nine categories, \ie, car, barrier, flower, grass, road, sign, tree, and other. The comparison results are shown in Tab.~\ref{tab:010}. As can be seen, all methods obtain diverse performances under different background objects. Among the 24 representative RGB-D based models, JL-DCF \cite{fu2020jl}, UC-Net \cite{zhang2020uc} and SSF \cite{zhang2020} achieve the top-three best results. In addition, the four most recent models, \ie, D$^3$Net \cite{fan2019rethinking}, S$^2$MA \cite{liu2020}, A2dele \cite{piao2020}, and ICNet \cite{li2020icnet} obtain better performance than the others.

$\bullet$ \textbf{Lighting Conditions}. The performance of SOD can be affected by different lighting conditions. To determine the performance of different RGB-D based SOD models under different lighting conditions, we conduct an evaluation on the SIP dataset \cite{fan2019rethinking}, which we split it into two categories, \ie, sunny and low-light. The comparison results are shown in Tab.~\ref{tab:011}. 
As can be seen, low-light negatively impacts SOD performance. Among comparison models, UC-Net \cite{zhang2020uc} obtains the best performance under sunny conditions while JL-DCF \cite{fu2020jl} achieves the best result under low-light condition.

In addition, we report the saliency maps generated for various challenging scenes to visualize the performance of different RGB-D based SOD models. Fig.~\ref{fig_010} and Fig.~\ref{fig_011} show some representative examples using two classic non-deep methods (DCMC \cite{cong2016saliency} and SE \cite{guo2016salient}) and eight state-of-the-art CNN-based models (DMRA \cite{piao2019depth}, D$^3$Net \cite{fan2019rethinking}, SSF \cite{zhang2020}, A2dele \cite{piao2020}, S$^2$MA \cite{liu2020}, ICNet \cite{li2020icnet}, JL-DCF \cite{fu2020jl}, and UC-Net \cite{zhang2020uc}). The $1^{st}$ row shows a small object, while the $2^{nd}$ row is an example of a large one. The $3^{rd}$ and  $4^{th}$ rows contain complex backgrounds and boundaries, respectively. The $5^{th}$ and  $6^{th}$ rows contain multiple salient objects. In the $7^{th}$ row, there are low-light condition. In the $8^{th}$ row, the depth map is coarse with very inaccurate object boundaries, which could inhibit the SOD performance. From the results in Fig.~\ref{fig_010} and Fig.~\ref{fig_011}, it can be observed that deep models perform better than non-deep models on these challenging scenes, confirming the powerful expression ability of deep features over handcrafted ones. In addition, D$^3$Net \cite{fan2019rethinking}, S$^2$MA \cite{liu2020}, JL-DCF \cite{fu2020jl}, and UC-Net \cite{zhang2020uc} perform better than other deep models.

\section{Challenges and Open Directions}
\label{sec:challenge}

\subsection{Effects of Imperfect Depth}

\textbf{Effects of Low-quality Depth Maps}. Depth maps with affluent spatial information have been proven beneficial in detecting salient objects from cluttered backgrounds, while the depth quality also directly affects the subsequent SOD performance. The quality of depth maps varies tremendously across different scenarios due to the limitations of depth sensors, posing a challenge when trying to reduce the effects of low-quality depth maps. However, most existing methods directly fuse RGB images and original raw data from depth maps, without considering the effects of low-quality depth maps. There are a few notable exceptions. For example, in \cite{zhao2019contrast}, a contrast-enhanced network was proposed to learn enhanced depth maps, which have much higher contrasts compared with the original depths. In \cite{zhang2020}, a compensation-aware loss was designed to pay more attention to hard samples containing unreliable depth information. Moreover, D{$^3$}Net \cite{fan2019rethinking} uses a depth depurator unit (DDU) to classify depth maps into two classes (\ie, reasonable and low-quality). The DDU also acts as a gate that can filter out the low-quality depth maps. However, the above methods often employ a two-step strategy to achieve depth enhancement and multi-modal fusion \cite{zhao2019contrast,zhang2020} or an independent gate operation for filtering out poor depths, which could bring a suboptimal problem. There is thus a need to develop an end-to-end framework that can achieve depth enhancement or adaptively weight the depth maps (\eg, assign low weights to poor depth maps) during multi-modal fusion, which would be more helpful for reducing the effects of low-quality depth maps and boosting SOD performance. 

\textbf{Incomplete Depth Maps}. In RGB-D datasets, it is inevitable for there to be some low-quality depth maps due to the limitations of the acquisition devices. As previously discussed, several depth enhancement algorithms have been used to improve the quality of depth maps. However, depth maps that suffer from severe noise or blurred edges, are often discarded. In this case, we have complete RGB images but some samples do not have depth maps, which is similar to the incomplete multi-view/modal learning problem \cite{xu2015multi,zhou2019effective,zhou2019latent,zhou2020multi,zhou2020hi}. Thus, we call it ``incomplete RGB-D based SOD". As current models only focus on the SOD task using complete RGB images and depth maps, we believe this could be a new direction for RGB-D SOD. 

\textbf{Depth Estimation}. Depth estimation provides an effective solution to recover high-quality depths and overcome the effects of low-quality depth maps. Various depth estimation approaches \cite{godard2017unsupervised,liu2015deep,wang2020sdc,jin2020geometric} have been developed, which could be introduced into the RGB-D based SOD task to improve performance.

\subsection{Effective Fusion Strategies} 

\textbf{Adversarial Learning-based Fusion}. It is important to effectively fuse RGB images and depth maps for RGB-D based SOD. Existing models often employ different fusion strategies (\eg, early fusion, middle fusion, or late fusion) to exploit the correlations between RGB images and depth maps. Recently, generative adversarial networks (GANs) \cite{mirza2014conditional} have gained widespread attention for the saliency detection task \cite{zhu2018multi,pan2017salgan}. In common GAN-based SOD models, a generator takes RGB images as inputs and generates the corresponding saliency maps, while a discriminator is adopted to determine whether a given image is synthetic or ground-truth. GAN-based models could easily be extended to RGB-D SOD, which could be helpful for boosting performance due to their superior feature learning ability. Moreover, GANs could also be used to learn common feature representations for RGB images and depth maps \cite{jiang2020cmsalgan}, which could help with feature or saliency map fusion and further boost the SOD performance.

\textbf{Attention-induced Fusion}. Attention mechanisms have been widely applied to various deep learning-based tasks \cite{vaswani2017attention,wang2017residual,fang2018pairwise,wang2017deep}, allowing networks to selectively pay attention to a subset of regions for extracting discriminative and powerful features. Besides, co-attention mechanisms have been developed to explore the underlying correlations across multiple modalities, and are widely studied in visual question answering \cite{lu2016hierarchical,yu2019deep} and video object segmentation \cite{lu2019see}. Thus, for the RGB-D based SOD task, we could also develop attention-based fusion algorithms to exploit correlations between RGB images and depth cues to improve the performance.

\subsection{Different Supervision Strategies}

Existing RGB-D models often use a fully supervised strategy to learn saliency prediction models. However, annotating pixel-level saliency maps is a tedious and time-consuming procedure. To alleviate this issue, there has been increased interest in weakly and semi-supervised learning, which have been applied to salient object detection \cite{zeng2019multi,zhang2017bridging,qian2019language,yan2019semi,zhou2018semi}. Semi-/weak supervision could also be introduced into RGB-D SOD, by leveraging image-level tags \cite{zeng2019multi} and pseudo pixel-wise annotations \cite{zhang2017supervision,yan2019semi}, for improving the detection performance. Besides, several studies \cite{chen2020adversarial,dai2020sg} have suggested that models pretrained using self-supervision can effectively be used to achieve better performance. Therefore, we could train saliency prediction models on large amounts of annotated RGB images in a self-supervised manner and then transfer the pretrained models to the RGB-D SOD task.  

\subsection{Dataset Collection}

\textbf{Dataset size}. Although there are nine public RGB-D datasets for SOD, their size is quite limited, \eg, the maximum size is about 2,000 samples for NJUD \cite{ju2014depth}. When compared with other RGB-D datasets for generic object detection or action recognition \cite{lai2011large,zhang2016large}, the size of RGB-D datasets for SOD is also very small. Thus, it is essential to develop new large-scale RGB-D datasets that can serve as baselines for future research.

\textbf{Complex Background \& Task-driven Datasets}. Most existing RGB-D datasets collect images that contain one salient object or multiple objects but with a relatively clean background. However, real-world applications often suffer from much more complicated situations (\eg, occlusion, appearance change, low illumination, etc), which could decrease the SOD performance. Thus, collecting images with complex background is critical to improve the generalization ability of RGB-D SOD models. Moreover, for some tasks, images with specific salient object(s) must be collected. For example, one important technology is road sign recognition in driver assistance systems, which requires images with road signs to be collected. Thus, it is essential to construct task-driven RGB-D datasets like SIP ~\cite{fan2019rethinking}.

\subsection{Model Design for Real-world Scenarios}

Some smartphones can capture depth maps (\eg, images in the SIP dataset were captured using Huawei Mate 10). Thus it would be feasible to conduct the SOD task in real-world applications, \eg, on smart devices. However, most existing methods include complicated and deep DNNs to increase the model capacity and achieve better performance, preventing them from being directly applied on real-work platforms. To overcome this, model compression \cite{he2018amc,cheng2017survey} techniques could be used to learn compact RGB-D based SOD models with promising detection accuracy. Moreover, JL-DCF \cite{fu2020jl} utilizes a shared network to locate salient objects using RGB and depth views, which largely reduces the model parameters and makes real-world applications feasible.

\subsection{Extension to RGB-T SOD}

In addition to RGB-D SOD, there are several other methods that fuse different modalities for better detection, such as RGB-T SOD, which integrates RGB and thermal infrared data. Thermal infrared cameras can capture the radiation emitted from any object with a temperature above absolute zero, making thermal infrared images insensitive to illumination conditions \cite{ma2017learning}. Therefore, thermal images can provide supplementary information to improve SOD performance when salient objects suffer from varying light, reflective light, or shadows. Some RGB-T models \cite{li2017unified,ma2017learning,wang2018rgb,sun2019rgb,tu2019m3s,tu2020multi,tu2019rgb,zhang2019rgb,tu2020rgbt} and datasets (VT821 \cite{wang2018rgb}, VT1000 \cite{tu2019rgb} and VT5000 \cite{tu2020rgbt}) have already been proposed over the past few years. Similar to RGB-D SOD, the key aim of RGB-T SOD is to fuse RGB and thermal infrared images and exploit the correlations between the two modalities. Thus, several advanced multi-modal fusion technologies in RGB-D SOD could be extended to the RGB-T SOD task.

\section{Conclusion}
\label{sec:conclusion}

In this paper we present, to the best of our knowledge, the first comprehensive review of RGB-D based SOD models. We first review the models from different perspectives, and then summarize popular RGB-D SOD datasets as well as provide details for each. Considering the fact that light fields also provide depth information, we also review popular light field SOD models and the related benchmark datasets. Next, we provide a comprehensive evaluation of 24 representative RGB-D based SOD models as well as an attribute-based evaluation. Specifically, we perform attribute-based performance analysis by constructing new datasets for the 24 representative RGB-D based SOD models. Moreover, we discuss several challenges and highlight open directions for future research. In addition, we briefly discuss the extension work to RGB-T SOD to improve performance when salient objects suffer from varying light, reflective light, or shadows. Although RGB-D based SOD has made notable progress over the past several decades, there is still significant room for improvement. We hope this survey will generate more interest in this field.

\footnotesize
\bibliographystyle{IEEEbib}
\bibliography{ref}

\begin{thebibliography}{100}

\bibitem{cong2016saliency}
Runmin Cong, Jianjun Lei, Changqing Zhang, Qingming Huang, Xiaochun Cao, and
  Chunping Hou,
\newblock ``Saliency detection for stereoscopic images based on depth
  confidence analysis and multiple cues fusion,''
\newblock {\em IEEE Signal Processing Letters}, vol. 23, no. 6, pp. 819--823,
  2016.

\bibitem{guo2016salient}
Jingfan Guo, Tongwei Ren, and Jia Bei,
\newblock ``Salient object detection for {RGB-D} image via saliency
  evolution,''
\newblock in {\em Proceedings of the IEEE International Conference on
  Multimedia and Expo}. IEEE, 2016, pp. 1--6.

\bibitem{fan2019rethinking}
Deng-Ping Fan, Zheng Lin, Zhao Zhang, Menglong Zhu, and Ming-Ming Cheng,
\newblock ``Rethinking {RGB-D} salient object detection: Models, data sets, and
  large-scale benchmarks,''
\newblock {\em IEEE Transactions on Neural Networks and Learning Systems},
  2020.

\bibitem{zhang2020}
Miao Zhang, Weisong Ren, Yongri Piao, Zhengkun Rong, and Huchuan Lu,
\newblock ``Select, supplement and focus for {RGB-D} saliency detection,''
\newblock in {\em Proceedings of the IEEE Conference on Computer Vision and
  Pattern Recognition}, 2020.

\bibitem{piao2020}
Yongri Piao, Zhengkun Rong, Miao Zhang, Weisong Ren, and Huchuan Lu,
\newblock ``A2dele: Adaptive and attentive depth distiller for efficient
  {RGB-D} salient object detection,''
\newblock {\em Proceedings of the IEEE Conference on Computer Vision and
  Pattern Recognition}, 2020.

\bibitem{liu2020}
Nian Liu, Ni~Zhang, and Junwei Han,
\newblock ``Learning selective self-mutual attention for {RGB-D} saliency
  detection,''
\newblock in {\em Proceedings of the IEEE Conference on Computer Vision and
  Pattern Recognition}, 2020.

\bibitem{li2020icnet}
Gongyang Li, Zhi Liu, and Haibin Ling,
\newblock ``Icnet: Information conversion network for {RGB-D} based salient
  object detection,''
\newblock {\em IEEE Transactions on Image Processing}, vol. 29, pp. 4873--4884,
  2020.

\bibitem{fu2020jl}
Keren Fu, Deng-Ping Fan, Ge-Peng Ji, and Qijun Zhao,
\newblock ``Jl-dcf: Joint learning and densely-cooperative fusion framework for
  {RGB-D} salient object detection,''
\newblock {\em Proceedings of the IEEE Conference on Computer Vision and
  Pattern Recognition}, 2020.

\bibitem{zhang2020uc}
Jing Zhang, Deng-Ping Fan, Yuchao Dai, Saeed Anwar, Fatemeh~Sadat Saleh, Tong
  Zhang, and Nick Barnes,
\newblock ``Uc-net: uncertainty inspired rgb-d saliency detection via
  conditional variational autoencoders,''
\newblock in {\em Proceedings of the IEEE Conference on Computer Vision and
  Pattern Recognition}, 2020.

\bibitem{fan2018salient}
Deng-Ping Fan, Ming-Ming Cheng, Jiang-Jiang Liu, Shang-Hua Gao, Qibin Hou, and
  Ali Borji,
\newblock ``Salient objects in clutter: Bringing salient object detection to
  the foreground,''
\newblock in {\em Proceedings of the European Conference on Computer Vision}.
  Springer, 2018, pp. 186--202.

\bibitem{nie2019multi}
Guang-Yu Nie, Ming-Ming Cheng, Yun Liu, Zhengfa Liang, Deng-Ping Fan, Yue Liu,
  and Yongtian Wang,
\newblock ``Multi-level context ultra-aggregation for stereo matching,''
\newblock in {\em Proceedings of the IEEE Conference on Computer Vision and
  Pattern Recognition}, 2019, pp. 3283--3291.

\bibitem{zhu2014unsupervised}
Jun-Yan Zhu, Jiajun Wu, Yan Xu, Eric Chang, and Zhuowen Tu,
\newblock ``Unsupervised object class discovery via saliency-guided multiple
  class learning,''
\newblock {\em IEEE Transactions on Pattern Analysis and Machine Intelligence},
  vol. 37, no. 4, pp. 862--875, 2014.

\bibitem{deng2020re}
Deng-Ping Fan, Tengpeng Li, Zheng Lin, Ge-Peng Ji, Dingwen Zhang, Ming-Ming
  Cheng, Huazhu Fu, and Jianbing Shen,
\newblock ``Re-thinking co-salient object detection,''
\newblock {\em arXiv preprint arXiv:2007.03380}, 2020.

\bibitem{rapantzikos2009dense}
Konstantinos Rapantzikos, Yannis Avrithis, and Stefanos Kollias,
\newblock ``Dense saliency-based spatiotemporal feature points for action
  recognition,''
\newblock in {\em Proceedings of the IEEE Conference on Computer Vision and
  Pattern Recognition}. IEEE, 2009, pp. 1454--1461.

\bibitem{fan2019shifting}
Deng-Ping Fan, Wenguan Wang, Ming-Ming Cheng, and Jianbing Shen,
\newblock ``Shifting more attention to video salient object detection,''
\newblock in {\em Proceedings of the IEEE Conference on Computer Vision and
  Pattern Recognition}, 2019, pp. 8554--8564.

\bibitem{wang2017saliency}
Wenguan Wang, Jianbing Shen, Ruigang Yang, and Fatih Porikli,
\newblock ``Saliency-aware video object segmentation,''
\newblock {\em IEEE Transactions on Pattern Analysis and Machine Intelligence},
  vol. 40, no. 1, pp. 20--33, 2017.

\bibitem{song2018pyramid}
Hongmei Song, Wenguan Wang, Sanyuan Zhao, Jianbing Shen, and Kin-Man Lam,
\newblock ``Pyramid dilated deeper convlstm for video salient object
  detection,''
\newblock in {\em Proceedings of the European Conference on Computer Vision}.
  Springer, 2018, pp. 715--731.

\bibitem{wang2017video}
Wenguan Wang, Jianbing Shen, and Ling Shao,
\newblock ``Video salient object detection via fully convolutional networks,''
\newblock {\em IEEE Transactions on Image Processing}, vol. 27, no. 1, pp.
  38--49, 2017.

\bibitem{shimoda2016distinct}
Wataru Shimoda and Keiji Yanai,
\newblock ``Distinct class-specific saliency maps for weakly supervised
  semantic segmentation,''
\newblock in {\em Proceedings of the European Conference on Computer Vision}.
  Springer, 2016, pp. 218--234.

\bibitem{zeng2019joint}
Yu~Zeng, Yunzhi Zhuge, Huchuan Lu, and Lihe Zhang,
\newblock ``Joint learning of saliency detection and weakly supervised semantic
  segmentation,''
\newblock in {\em Proceedings of the IEEE International Conference on Computer
  Vision}. Springer, 2019, pp. 7223--7233.

\bibitem{fan2020pra}
Deng-Ping Fan, Ge-Peng Ji, Tao Zhou, Geng Chen, Huazhu Fu, Jianbing Shen, and
  Ling Shao,
\newblock ``Pranet: Parallel reverse attention network for polyp
  segmentation,''
\newblock in {\em Medical Image Computing and Computer-Assisted Intervention},
  2020.

\bibitem{fan2020inf}
Deng-Ping Fan, Tao Zhou, Ge-Peng Ji, Yi~Zhou, Geng Chen, Huazhu Fu, Jianbing
  Shen, and Ling Shao,
\newblock ``Inf-net: Automatic covid-19 lung infection segmentation from ct
  images,''
\newblock {\em IEEE Transactions on Medical Imaging}, 2020.

\bibitem{wu2020jcs}
Yu-Huan Wu, Shang-Hua Gao, Jie Mei, Jun Xu, Deng-Ping Fan, Chao-Wei Zhao, and
  Ming-Ming Cheng,
\newblock ``Jcs: An explainable covid-19 diagnosis system by joint
  classification and segmentation,''
\newblock {\em arXiv preprint arXiv:2004.07054}, 2020.

\bibitem{mahadevan2009saliency}
Vijay Mahadevan and Nuno Vasconcelos,
\newblock ``Saliency-based discriminant tracking,''
\newblock in {\em Proceedings of the IEEE Conference on Computer Vision and
  Pattern Recognition}. IEEE, 2009, pp. 1007--1013.

\bibitem{hong2015online}
Seunghoon Hong, Tackgeun You, Suha Kwak, and Bohyung Han,
\newblock ``Online tracking by learning discriminative saliency map with
  convolutional neural network,''
\newblock in {\em Proceedings of the International Conference on Machine
  Learning}, 2015, pp. 597--606.

\bibitem{zhao2016person}
Rui Zhao, Wanli Oyang, and Xiaogang Wang,
\newblock ``Person re-identification by saliency learning,''
\newblock {\em IEEE Transactions on Pattern Analysis and Machine Intelligence},
  vol. 39, no. 2, pp. 356--370, 2016.

\bibitem{martinel2015kernelized}
Niki Martinel, Christian Micheloni, and Gian~Luca Foresti,
\newblock ``Kernelized saliency-based person re-identification through multiple
  metric learning,''
\newblock {\em IEEE Transactions on Image Processing}, vol. 24, no. 12, pp.
  5645--5658, 2015.

\bibitem{fan2020camouflaged}
Deng-Ping Fan, Ge-Peng Ji, Guolei Sun, Ming-Ming Cheng, Jianbing Shen, and Ling
  Shao,
\newblock ``Camouflaged object detection,''
\newblock in {\em Proceedings of the IEEE Conference on Computer Vision and
  Pattern Recognition}, 2020, pp. 2777--2787.

\bibitem{liu2013model}
Guanghai Liu and Dengping Fan,
\newblock ``A model of visual attention for natural image retrieval,''
\newblock in {\em Proceedings of the IEEE Conference on Information Science and
  Cloud Computing Companion}. IEEE, 2013, pp. 728--733.

\bibitem{zhao2019egnet}
Jia-Xing Zhao, Jiang-Jiang Liu, Deng-Ping Fan, Yang Cao, Jufeng Yang, and
  Ming-Ming Cheng,
\newblock ``Egnet: Edge guidance network for salient object detection,''
\newblock in {\em Proceedings of the IEEE International Conference on Computer
  Vision}, 2019, pp. 8779--8788.

\bibitem{tu2016real}
Wei-Chih Tu, Shengfeng He, Qingxiong Yang, and Shao-Yi Chien,
\newblock ``Real-time salient object detection with a minimum spanning tree,''
\newblock in {\em Proceedings of the IEEE conference on Computer Vision and
  Pattern Recognition}, 2016, pp. 2334--2342.

\bibitem{xia2017and}
Changqun Xia, Jia Li, Xiaowu Chen, Anlin Zheng, and Yu~Zhang,
\newblock ``What is and what is not a salient object? learning salient object
  detector by ensembling linear exemplar regressors,''
\newblock in {\em Proceedings of the IEEE Conference on Computer Vision and
  Pattern Recognition}, 2017, pp. 4142--4150.

\bibitem{hou2007saliency}
Xiaodi Hou and Liqing Zhang,
\newblock ``Saliency detection: A spectral residual approach,''
\newblock in {\em Proceedings of the IEEE Conference on Computer Vision and
  Pattern Recognition}. IEEE, 2007, pp. 1--8.

\bibitem{yan2013hierarchical}
Qiong Yan, Li~Xu, Jianping Shi, and Jiaya Jia,
\newblock ``Hierarchical saliency detection,''
\newblock in {\em Proceedings of the IEEE Conference on Computer Vision and
  Pattern Recognition}, 2013, pp. 1155--1162.

\bibitem{yang2013saliency}
Chuan Yang, Lihe Zhang, Huchuan Lu, Xiang Ruan, and Ming-Hsuan Yang,
\newblock ``Saliency detection via graph-based manifold ranking,''
\newblock in {\em Proceedings of the IEEE Conference on Computer Vision and
  Pattern Recognition}, 2013, pp. 3166--3173.

\bibitem{li2016deep}
Guanbin Li and Yizhou Yu,
\newblock ``Deep contrast learning for salient object detection,''
\newblock in {\em Proceedings of the IEEE Conference on Computer Vision and
  Pattern Recognition}, 2016, pp. 478--487.

\bibitem{zhang2016co}
Dingwen Zhang, Deyu Meng, and Junwei Han,
\newblock ``Co-saliency detection via a self-paced multiple-instance learning
  framework,''
\newblock {\em IEEE Transactions on Pattern Analysis and Machine Intelligence},
  vol. 39, no. 5, pp. 865--878, 2016.

\bibitem{zhang2017amulet}
Pingping Zhang, Dong Wang, Huchuan Lu, Hongyu Wang, and Xiang Ruan,
\newblock ``Amulet: Aggregating multi-level convolutional features for salient
  object detection,''
\newblock in {\em Proceedings of the IEEE International Conference on Computer
  Vision}, 2017, pp. 202--211.

\bibitem{zhang2017learning}
Pingping Zhang, Dong Wang, Huchuan Lu, Hongyu Wang, and Baocai Yin,
\newblock ``Learning uncertain convolutional features for accurate saliency
  detection,''
\newblock in {\em Proceedings of the IEEE International Conference on Computer
  Vision}, 2017, pp. 212--221.

\bibitem{wang2017stagewise}
Tiantian Wang, Ali Borji, Lihe Zhang, Pingping Zhang, and Huchuan Lu,
\newblock ``A stagewise refinement model for detecting salient objects in
  images,''
\newblock in {\em Proceedings of the IEEE International Conference on Computer
  Vision}, 2017, pp. 4019--4028.

\bibitem{Li_2018_ECCV}
Xin Li, Fan Yang, Hong Cheng, Wei Liu, and Dinggang Shen,
\newblock ``Contour knowledge transfer for salient object detection,''
\newblock in {\em Proceedings of the Proceedings of the European Conference on
  Computer Vision}. Springer, September 2018.

\bibitem{wangwen2019salient}
Wenguan Wang, Shuyang Zhao, Jianbing Shen, Steven~CH Hoi, and Ali Borji,
\newblock ``Salient object detection with pyramid attention and salient
  edges,''
\newblock in {\em Proceedings of the IEEE Conference on Computer Vision and
  Pattern Recognition}, 2019, pp. 1448--1457.

\bibitem{su2019selectivity}
Jinming Su, Jia Li, Yu~Zhang, Changqun Xia, and Yonghong Tian,
\newblock ``Selectivity or invariance: Boundary-aware salient object
  detection,''
\newblock in {\em Proceedings of the IEEE International Conference on Computer
  Vision}, 2019, pp. 3799--3808.

\bibitem{zhao2019pyramid}
Ting Zhao and Xiangqian Wu,
\newblock ``Pyramid feature attention network for saliency detection,''
\newblock in {\em Proceedings of the IEEE Conference on Computer Vision and
  Pattern Recognition}, 2019, pp. 3085--3094.

\bibitem{chen2019cnn}
Hao Chen and Youfu Li,
\newblock ``Cnn-based rgb-d salient object detection: Learn, select and fuse,''
\newblock {\em arXiv preprint arXiv:1909.09309}, 2019.

\bibitem{lang2012depth}
Congyan Lang, Tam~V Nguyen, Harish Katti, Karthik Yadati, Mohan Kankanhalli,
  and Shuicheng Yan,
\newblock ``Depth matters: Influence of depth cues on visual saliency,''
\newblock in {\em Proceedings of the European Conference on Computer Vision}.
  Springer, 2012, pp. 101--115.

\bibitem{ciptadi2013depth}
Arridhana Ciptadi, Tucker Hermans, and James~M Rehg,
\newblock ``An in depth view of saliency,''
\newblock Georgia Institute of Technology, 2013.

\bibitem{desingh2013depth}
Karthik Desingh, K~Madhava Krishna, Deepu Rajan, and CV~Jawahar,
\newblock ``Depth really matters: Improving visual salient region detection
  with depth,''
\newblock in {\em Proceedings of the British Machine Vision Conference}, 2013.

\bibitem{cheng2014depth}
Yupeng Cheng, Huazhu Fu, Xingxing Wei, Jiangjian Xiao, and Xiaochun Cao,
\newblock ``Depth enhanced saliency detection method,''
\newblock in {\em Proceedings of the International Conference on Internet
  Multimedia Computing and Service}, 2014, pp. 23--27.

\bibitem{ren2015exploiting}
Jianqiang Ren, Xiaojin Gong, Lu~Yu, Wenhui Zhou, and Michael Ying~Yang,
\newblock ``Exploiting global priors for {RGB-D} saliency detection,''
\newblock in {\em Proceedings of the IEEE Conference on Computer Vision and
  Pattern Recognition Workshops}, 2015, pp. 25--32.

\bibitem{peng2014rgbd}
Houwen Peng, Bing Li, Weihua Xiong, Weiming Hu, and Rongrong Ji,
\newblock ``Rgbd salient object detection: a benchmark and algorithms,''
\newblock in {\em Proceedings of the European Conference on Computer Vision}.
  Springer, 2014, pp. 92--109.

\bibitem{qu2017rgbd}
Liangqiong Qu, Shengfeng He, Jiawei Zhang, Jiandong Tian, Yandong Tang, and
  Qingxiong Yang,
\newblock ``{RGBD} salient object detection via deep fusion,''
\newblock {\em IEEE Transactions on Image Processing}, vol. 26, no. 5, pp.
  2274--2285, 2017.

\bibitem{zhao2019contrast}
Jia-Xing Zhao, Yang Cao, Deng-Ping Fan, Ming-Ming Cheng, Xuan-Yi Li, and
  Le~Zhang,
\newblock ``Contrast prior and fluid pyramid integration for {RGBD} salient
  object detection,''
\newblock in {\em Proceedings of the IEEE Conference on Computer Vision and
  Pattern Recognition}, 2019, pp. 3927--3936.

\bibitem{piao2019depth}
Yongri Piao, Wei Ji, Jingjing Li, Miao Zhang, and Huchuan Lu,
\newblock ``Depth-induced multi-scale recurrent attention network for saliency
  detection,''
\newblock in {\em Proceedings of the IEEE International Conference on Computer
  Vision}, 2019, pp. 7254--7263.

\bibitem{chen2019multi}
Hao Chen, Youfu Li, and Dan Su,
\newblock ``Multi-modal fusion network with multi-scale multi-path and
  cross-modal interactions for {RGB-D} salient object detection,''
\newblock {\em Pattern Recognition}, vol. 86, pp. 376--385, 2019.

\bibitem{ju2014depth}
Ran Ju, Ling Ge, Wenjing Geng, Tongwei Ren, and Gangshan Wu,
\newblock ``Depth saliency based on anisotropic center-surround difference,''
\newblock in {\em Proceedings of the IEEE International Conference on Image
  Processing}. IEEE, 2014, pp. 1115--1119.

\bibitem{feng2016local}
David Feng, Nick Barnes, Shaodi You, and Chris McCarthy,
\newblock ``Local background enclosure for {RGB-D} salient object detection,''
\newblock in {\em Proceedings of the IEEE Conference on Computer Vision and
  Pattern Recognition}, 2016, pp. 2343--2350.

\bibitem{han2017cnns}
Junwei Han, Hao Chen, Nian Liu, Chenggang Yan, and Xuelong Li,
\newblock ``Cnns-based {RGB-D} saliency detection via cross-view transfer and
  multiview fusion,''
\newblock {\em IEEE Transactions on Cybernetics}, vol. 48, no. 11, pp.
  3171--3183, 2017.

\bibitem{borji2015salient}
Ali Borji, Ming-Ming Cheng, Huaizu Jiang, and Jia Li,
\newblock ``Salient object detection: A benchmark,''
\newblock {\em IEEE Transactions on Image Processing}, vol. 24, no. 12, pp.
  5706--5722, 2015.

\bibitem{cong2018review}
Runmin Cong, Jianjun Lei, Huazhu Fu, Ming-Ming Cheng, Weisi Lin, and Qingming
  Huang,
\newblock ``Review of visual saliency detection with comprehensive
  information,''
\newblock {\em IEEE Transactions on Circuits and Systems for Video Technology},
  vol. 29, no. 10, pp. 2941--2959, 2018.

\bibitem{zhang2018review}
Dingwen Zhang, Huazhu Fu, Junwei Han, Ali Borji, and Xuelong Li,
\newblock ``A review of co-saliency detection algorithms: Fundamentals,
  applications, and challenges,''
\newblock {\em ACM Transactions on Intelligent Systems and Technology}, vol. 9,
  no. 4, pp. 1--31, 2018.

\bibitem{han2018advanced}
Junwei Han, Dingwen Zhang, Gong Cheng, Nian Liu, and Dong Xu,
\newblock ``Advanced deep-learning techniques for salient and category-specific
  object detection: a survey,''
\newblock {\em IEEE Signal Processing Magazine}, vol. 35, no. 1, pp. 84--100,
  2018.

\bibitem{nguyen2018attentive}
Tam~V Nguyen, Qi~Zhao, and Shuicheng Yan,
\newblock ``Attentive systems: A survey,''
\newblock {\em International Journal of Computer Vision}, vol. 126, no. 1, pp.
  86--110, 2018.

\bibitem{borji2014salient}
Ali Borji, Ming-Ming Cheng, Qibin Hou, Huaizu Jiang, and Jia Li,
\newblock ``Salient object detection: A survey,''
\newblock {\em Computational Visual Media}, pp. 1--34, 2014.

\bibitem{zhao2019object}
Zhong-Qiu Zhao, Peng Zheng, Shou-tao Xu, and Xindong Wu,
\newblock ``Object detection with deep learning: A review,''
\newblock {\em IEEE Transactions on Neural Networks and Learning Systems}, vol.
  30, no. 11, pp. 3212--3232, 2019.

\bibitem{wang2019salient}
Wenguan Wang, Qiuxia Lai, Huazhu Fu, Jianbing Shen, Haibin Ling, and Ruigang
  Yang,
\newblock ``Salient object detection in the deep learning era: An in-depth
  survey,''
\newblock {\em arXiv preprint arXiv:1904.09146}, 2019.

\bibitem{zhang2012depth}
Hailong Zhang, Jianjun Lei, Xiaohong Fan, Meimin Wu, Peng Zhang, and Shupo Bu,
\newblock ``Depth combined saliency detection based on region contrast model,''
\newblock in {\em Proceedings of International Conference on Computer Science
  \& Education}. IEEE, 2012, pp. 763--766.

\bibitem{lei2013evaluation}
Jianjun Lei, Hailong Zhang, Lei You, Chunping Hou, and Laihua Wang,
\newblock ``Evaluation and modeling of depth feature incorporated visual
  attention for salient object segmentation,''
\newblock {\em Neurocomputing}, vol. 120, pp. 24--33, 2013.

\bibitem{fan2014salient}
Xingxing Fan, Zhi Liu, and Guangling Sun,
\newblock ``Salient region detection for stereoscopic images,''
\newblock in {\em Proceedings of the International Conference on Digital Signal
  Processing}. IEEE, 2014, pp. 454--458.

\bibitem{guo2015salient}
Jingfan Guo, Tongwei Ren, Jia Bei, and Yujin Zhu,
\newblock ``Salient object detection in rgb-d image based on saliency fusion
  and propagation,''
\newblock in {\em Proceedings of the International Conference on Internet
  Multimedia Computing and Service}, 2015, pp. 1--5.

\bibitem{tang2016depth}
Yanlong Tang, Ruofeng Tong, Min Tang, and Yun Zhang,
\newblock ``Depth incorporating with color improves salient object detection,''
\newblock {\em The Visual Computer}, vol. 32, no. 1, pp. 111--121, 2016.

\bibitem{jiang2015salient}
Lixing Jiang, Artur Koch, and Andreas Zell,
\newblock ``Salient regions detection for indoor robots using rgb-d data,''
\newblock in {\em Proceedings of the IEEE International Conference on Robotics
  and Automation}. IEEE, 2015, pp. 1323--1328.

\bibitem{xue2015rgb}
Haoyang Xue, Yun Gu, Yijun Li, and Jie Yang,
\newblock ``Rgb-d saliency detection via mutual guided manifold ranking,''
\newblock in {\em Proceedings of IEEE International Conference on Image
  Processing}. IEEE, 2015, pp. 666--670.

\bibitem{zhu2015selective}
Lei Zhu, Zhiguo Cao, Zhiwen Fang, Yang Xiao, Jin Wu, Huiping Deng, and Jing
  Liu,
\newblock ``Selective features for rgb-d saliency,''
\newblock in {\em Proceedings of Chinese Automation Congress}. IEEE, 2015, pp.
  512--517.

\bibitem{du2016improving}
Huan Du, Zhi Liu, Hangke Song, Lin Mei, and Zheng Xu,
\newblock ``Improving {RGBD} saliency detection using progressive region
  classification and saliency fusion,''
\newblock {\em IEEE Access}, vol. 4, pp. 8987--8994, 2016.

\bibitem{wang2016rgb}
Song-Tao Wang, Zhen Zhou, Han-Bing Qu, and Bin Li,
\newblock ``Rgb-d saliency detection under bayesian framework,''
\newblock in {\em Proceedings of International Conference on Pattern
  Recognition}. IEEE, 2016, pp. 1881--1886.

\bibitem{sheng2016saliency}
Hao Sheng, Xiaoyu Liu, and Shuo Zhang,
\newblock ``Saliency analysis based on depth contrast increased,''
\newblock in {\em Proceedings of IEEE International Conference on Acoustics,
  Speech and Signal Processing}. IEEE, 2016, pp. 1347--1351.

\bibitem{song2016depth}
Hangke Song, Zhi Liu, Huan Du, and Guangling Sun,
\newblock ``Depth-aware saliency detection using discriminative saliency
  fusion,''
\newblock in {\em Proceedings of IEEE International Conference on Acoustics,
  Speech and Signal Processing}. IEEE, 2016, pp. 1626--1630.

\bibitem{wang2016visual}
Song-Tao Wang, Zhen Zhou, Han-Bing Qu, and Bin Li,
\newblock ``Visual saliency detection for {RGB-D} images with generative
  model,''
\newblock in {\em Proceedings of the Asian Conference on Computer Vision}.
  Springer, 2016, pp. 20--35.

\bibitem{feng2017hoso}
David Feng, Nick Barnes, and Shaodi You,
\newblock ``Hoso: Histogram of surface orientation for rgb-d salient object
  detection,''
\newblock in {\em Proceedings of the International Conference on Digital Image
  Computing: Techniques and Applications}. IEEE, 2017, pp. 1--8.

\bibitem{chen2017m}
Hao Chen, You-Fu Li, and Dan Su,
\newblock ``M3net: {M}ulti-scale multi-path multi-modal fusion network and
  example application to rgb-d salient object detection,''
\newblock in {\em Proceedings of IEEE/RSJ International Conference on
  Intelligent Robots and Systems}. IEEE, 2017, pp. 4911--4916.

\bibitem{chen2017rgb}
Hao Chen, Youfu Li, and Dan Su,
\newblock ``{RGB-D} saliency detection by multi-stream late fusion network,''
\newblock in {\em Proceedings of the International Conference on Computer
  Vision Systems}. Springer, 2017, pp. 459--468.

\bibitem{shi2017learning}
Riku Shigematsu, David Feng, Shaodi You, and Nick Barnes,
\newblock ``Learning {RGB-D} salient object detection using background
  enclosure, depth contrast, and top-down features,''
\newblock in {\em Proceedings of the IEEE International Conference on Computer
  Vision Workshops}, 2017, pp. 2749--2757.

\bibitem{zhu2017innovative}
Chunbiao Zhu, Ge~Li, Wenmin Wang, and Ronggang Wang,
\newblock ``An innovative salient object detection using center-dark channel
  prior,''
\newblock in {\em Proceedings of the IEEE International Conference on Computer
  Vision Workshops}, 2017, pp. 1509--1515.

\bibitem{zhu2017three}
Chunbiao Zhu and Ge~Li,
\newblock ``A three-pathway psychobiological framework of salient object
  detection using stereoscopic technology,''
\newblock in {\em Proceedings of the IEEE International Conference on Computer
  Vision Workshops}, 2017, pp. 3008--3014.

\bibitem{wang2017rgb}
Anzhi Wang and Minghui Wang,
\newblock ``{RGB-D} salient object detection via minimum barrier distance
  transform and saliency fusion,''
\newblock {\em IEEE Signal Processing Letters}, vol. 24, no. 5, pp. 663--667,
  2017.

\bibitem{song2017depth}
Hangke Song, Zhi Liu, Huan Du, Guangling Sun, Olivier Le~Meur, and Tongwei Ren,
\newblock ``Depth-aware salient object detection and segmentation via
  multiscale discriminative saliency fusion and bootstrap learning,''
\newblock {\em IEEE Transactions on Image Processing}, vol. 26, no. 9, pp.
  4204--4216, 2017.

\bibitem{cong2017iterative}
Runmin Cong, Jianjun Lei, Huazhu Fu, Weisi Lin, Qingming Huang, Xiaochun Cao,
  and Chunping Hou,
\newblock ``An iterative co-saliency framework for {RGBD} images,''
\newblock {\em IEEE Transactions on Cybernetics}, vol. 49, no. 1, pp. 233--246,
  2017.

\bibitem{imamoglu2018integration}
Nevrez Imamoglu, Wataru Shimoda, Chi Zhang, Yuming Fang, Asako Kanezaki, Keiji
  Yanai, and Yoshifumi Nishida,
\newblock ``An integration of bottom-up and top-down salient cues on rgb-d
  data: saliency from objectness versus non-objectness,''
\newblock {\em Signal, Image and Video Processing}, vol. 12, no. 2, pp.
  307--314, 2018.

\bibitem{cong2018hscs}
Runmin Cong, Jianjun Lei, Huazhu Fu, Qingming Huang, Xiaochun Cao, and Nam
  Ling,
\newblock ``{HSCS}: Hierarchical sparsity based co-saliency detection for
  {RGBD} images,''
\newblock {\em IEEE Transactions on Multimedia}, vol. 21, no. 7, pp.
  1660--1671, 2018.

\bibitem{cong2017co}
Runmin Cong, Jianjun Lei, Huazhu Fu, Qingming Huang, Xiaochun Cao, and Chunping
  Hou,
\newblock ``Co-saliency detection for {RGBD} images based on multi-constraint
  feature matching and cross label propagation,''
\newblock {\em IEEE Transactions on Image Processing}, vol. 27, no. 2, pp.
  568--579, 2017.

\bibitem{chen2018progressively}
Hao Chen and Youfu Li,
\newblock ``Progressively complementarity-aware fusion network for {RGB-D}
  salient object detection,''
\newblock in {\em Proceedings of the IEEE Conference on Computer Vision and
  Pattern Recognition}, 2018, pp. 3051--3060.

\bibitem{huang2018rgbd}
Posheng Huang, Chin-Han Shen, and Hsu-Feng Hsiao,
\newblock ``Rgbd salient object detection using spatially coherent deep
  learning framework,''
\newblock in {\em Proceedings of the IEEE International Conference on Digital
  Signal Processing}. IEEE, 2018, pp. 1--5.

\bibitem{chen2018attention}
Hao Chen, You-Fu Li, and Dan Su,
\newblock ``Attention-aware cross-modal cross-level fusion network for {RGB-D}
  salient object detection,''
\newblock in {\em Proceedings of the IEEE/RSJ International Conference on
  Intelligent Robots and Systems}. IEEE, 2018, pp. 6821--6826.

\bibitem{liang2018stereoscopic}
Fangfang Liang, Lijuan Duan, Wei Ma, Yuanhua Qiao, Zhi Cai, and Laiyun Qing,
\newblock ``Stereoscopic saliency model using contrast and
  depth-guided-background prior,''
\newblock {\em Neurocomputing}, vol. 275, pp. 2227--2238, 2018.

\bibitem{liu2019salient}
Zhengyi Liu, Song Shi, Quntao Duan, Wei Zhang, and Peng Zhao,
\newblock ``Salient object detection for {RGB-D} image by single stream
  recurrent convolution neural network,''
\newblock {\em Neurocomputing}, vol. 363, pp. 46--57, 2019.

\bibitem{zhu2019pdnet}
Chunbiao Zhu, Xing Cai, Kan Huang, Thomas~H Li, and Ge~Li,
\newblock ``{PDNet}: Prior-model guided depth-enhanced network for salient
  object detection,''
\newblock in {\em Proceedings of the IEEE International Conference on
  Multimedia and Expo}. IEEE, 2019, pp. 199--204.

\bibitem{piao2019saliency}
Yongri Piao, Xiao Li, Miao Zhang, Jingyi Yu, and Huchuan Lu,
\newblock ``Saliency detection via depth-induced cellular automata on light
  field,''
\newblock {\em IEEE Transactions on Image Processing}, vol. 29, pp. 1879--1889,
  2020.

\bibitem{chen2019three}
Hao Chen and Youfu Li,
\newblock ``Three-stream attention-aware network for {RGB-D} salient object
  detection,''
\newblock {\em IEEE Transactions on Image Processing}, vol. 28, no. 6, pp.
  2825--2835, 2019.

\bibitem{chen2019discriminative}
Hao Chen, Youfu Li, and Dan Su,
\newblock ``Discriminative cross-modal transfer learning and densely
  cross-level feedback fusion for {RGB-D} salient object detection,''
\newblock {\em IEEE Transactions on Cybernetics}, 2019.

\bibitem{cong2019going}
Runmin Cong, Jianjun Lei, Huazhu Fu, Junhui Hou, Qingming Huang, and Sam Kwong,
\newblock ``Going from {RGB} to {RGBD} saliency: A depth-guided transformation
  model,''
\newblock {\em IEEE Transactions on Cybernetics}, 2019.

\bibitem{wang2019adaptive}
Ningning Wang and Xiaojin Gong,
\newblock ``Adaptive fusion for {RGB-D} salient object detection,''
\newblock {\em IEEE Access}, vol. 7, pp. 55277--55284, 2019.

\bibitem{zhou2020attention}
Xiaofei Zhou, Gongyang Li, Chen Gong, Zhi Liu, and Jiyong Zhang,
\newblock ``Attention-guided {RGBD} saliency detection using appearance
  information,''
\newblock {\em Image and Vision Computing}, vol. 95, pp. 103888, 2020.

\bibitem{liu2020cross}
Zhengyi Liu, Wei Zhang, and Peng Zhao,
\newblock ``A cross-modal adaptive gated fusion generative adversarial network
  for {RGB-D} salient object detection,''
\newblock {\em Neurocomputing}, 2020.

\bibitem{liang2020cocnn}
Fangfang Liang, Lijuan Duan, Wei Ma, Yuanhua Qiao, Zhi Cai, Jun Miao, and
  Qixiang Ye,
\newblock ``Cocnn: {RGB-D} deep fusion for stereoscopic salient object
  detection,''
\newblock {\em Pattern Recognition}, p. 107329, 2020.

\bibitem{li2020asif}
Chongyi Li, Runmin Cong, Sam Kwong, Junhui Hou, Huazhu Fu, Guopu Zhu, Dingwen
  Zhang, and Qingming Huang,
\newblock ``{ASIF-N}et: Attention steered interweave fusion network for {RGB-D}
  salient object detection,''
\newblock {\em IEEE Transactions on Cybernetics}, 2020.

\bibitem{huang2020triple}
Rui Huang, Yan Xing, and Yaobin Zou,
\newblock ``Triple-complementary network for {RGB-D} salient object
  detection,''
\newblock {\em IEEE Signal Processing Letters}, 2020.

\bibitem{chen2020improved}
Chenglizhao Chen, Jipeng Wei, Chong Peng, Weizhong Zhang, and Hong Qin,
\newblock ``Improved saliency detection in {RGB-D} images using two-phase depth
  estimation and selective deep fusion,''
\newblock {\em IEEE Transactions on Image Processing}, vol. 29, pp. 4296--4307,
  2020.

\bibitem{huang2019rgb}
Rui Huang, Yan Xing, and ZeZheng Wang,
\newblock ``{RGB}-{D} salient object detection by a {CNN} with multiple layers
  fusion,''
\newblock {\em IEEE Signal Processing Letters}, vol. 26, no. 4, pp. 552--556,
  2019.

\bibitem{liu2019two}
Di~Liu, Yaosi Hu, Kao Zhang, and Zhenzhong Chen,
\newblock ``Two-stream refinement network for rgb-d saliency detection,''
\newblock in {\em Proceedings of IEEE International Conference on Image
  Processing}. IEEE, 2019, pp. 3925--3929.

\bibitem{du2019salient}
Huan Du, Zhi Liu, and Ran Shi,
\newblock ``Salient object segmentation based on depth-aware image layering,''
\newblock {\em Multimedia Tools and Applications}, vol. 78, no. 9, pp.
  12125--12138, 2019.

\bibitem{zhou2019global}
Wujie Zhou, Ying Lv, Jingsheng Lei, and Lu~Yu,
\newblock ``Global and local-contrast guides content-aware fusion for rgb-d
  saliency prediction,''
\newblock {\em IEEE Transactions on Systems, Man, and Cybernetics: Systems},
  2019.

\bibitem{ma2015learning}
Chih-Yao Ma and Hsueh-Ming Hang,
\newblock ``Learning-based saliency model with depth information,''
\newblock {\em Journal of vision}, vol. 15, no. 6, pp. 19--19, 2015.

\bibitem{jin2019co}
Zhigang Jin, Jingkun Li, and Dong Li,
\newblock ``Co-saliency detection for rgbd images based on effective
  propagation mechanism,''
\newblock {\em IEEE Access}, vol. 7, pp. 141311--141318, 2019.

\bibitem{ding2019depth}
Yu~Ding, Zhi Liu, Mengke Huang, Ran Shi, and Xiangyang Wang,
\newblock ``Depth-aware saliency detection using convolutional neural
  networks,''
\newblock {\em Journal of Visual Communication and Image Representation}, vol.
  61, pp. 1--9, 2019.

\bibitem{chen2020depth}
Zuyao Chen and Qingming Huang,
\newblock ``Depth potentiality-aware gated attention network for {RGB-D}
  salient object detection,''
\newblock {\em arXiv preprint arXiv:2003.08608}, 2020.

\bibitem{wang2020}
Yue Wang, Yuke Li, James~H Elder, Huchuan Lu, and Runmin Wu,
\newblock ``Synergistic saliency and depth prediction for {RGB-D} saliency
  detection,''
\newblock {\em arXiv preprint arXiv:2007.01711}, 2020.

\bibitem{jiang2020cmsalgan}
Bo~Jiang, Zitai Zhou, Xiao Wang, Jin Tang, and Bin Luo,
\newblock ``cmsalgan: {RGB-D} salient object detection with cross-view
  generative adversarial networks,''
\newblock {\em IEEE Transactions on Multimedia}, 2020.

\bibitem{xiao2020multi}
Fen Xiao, Bin Li, Yimu Peng, Chunhong Cao, Kai Hu, and Xieping Gao,
\newblock ``Multi-modal weights sharing and hierarchical feature fusion for
  rgbd salient object detection,''
\newblock {\em IEEE Access}, vol. 8, pp. 26602--26611, 2020.

\bibitem{zhang2020bilateral}
Zhao Zhang, Zheng Lin, Jun Xu, Wenda Jin, Shao-Ping Lu, and Deng-Ping Fan,
\newblock ``Bilateral attention network for rgb-d salient object detection,''
\newblock {\em arXiv preprint arXiv:2004.14582}, 2020.

\bibitem{zhou2020gfnet}
Wujie Zhou, Yuzhen Chen, Chang Liu, and Lu~Yu,
\newblock ``{GFN}et: Gate fusion network with res2net for detecting salient
  objects in rgb-d images,''
\newblock {\em IEEE Signal Processing Letters}, 2020.

\bibitem{liu2020salient}
Zhengyi Liu, Jiting Tang, Qian Xiang, and Peng Zhao,
\newblock ``Salient object detection for rgb-d images by generative adversarial
  network,''
\newblock {\em Multimedia Tools and Applications}, pp. 1--23, 2020.

\bibitem{lieccv20}
Gongyang Li, Zhi Liu, Linwei Ye, Yang Wang, and Haibin Ling,
\newblock ``Cross-modal weighting network for rgb-d salient object detection,''
\newblock in {\em Proceedings of the European Conference on Computer Vision}.
  Springer, 2020.

\bibitem{paneccv2020}
Youwei Pang, Lihe Zhang, Xiaoqi Zhao, and Huchuan Lu,
\newblock ``Hierarchical dynamic filtering network for {RGB-D} salient object
  detection,''
\newblock in {\em Proceedings of the European Conference on Computer Vision}.
  Springer, 2020.

\bibitem{luoECCV2020}
Ao~Luo, Xin Li, Fan Yang, Zhicheng Jiao, Hong Cheng, and Siwei Lyu,
\newblock ``Cascade graph neural networks for rgb-d salient object detection,''
\newblock in {\em Proceedings of the Proceedings of the European Conference on
  Computer Vision}. Springer, 2020.

\bibitem{li2020}
Chongyi Li, Runmin Cong, Yongri Piao, Qianqian Xu, and Chen~Change Loy,
\newblock ``Rgb-d salient object detection with cross-modality modulation and
  selection,''
\newblock in {\em Proceedings of the European Conference on Computer Vision}.
  Springer, 2020.

\bibitem{zhaoeccv20}
Xiaoqi Zhao, Lihe Zhang, Youwei Pang, Huchuan Lu, and Lei Zhang,
\newblock ``A single stream network for robust and real-time rgb-d salient
  object detection,''
\newblock in {\em Proceedings of the European Conference on Computer Vision}.
  Springer, 2020.

\bibitem{Wei_2020_ECCV}
Wei {Ji}, Jingjing {Li}, Miao {Zhang}, Yongri {Piao}, and Huchuan {Lu},
\newblock ``Accurate rgb-d salient object detection via collaborative
  learning,''
\newblock in {\em ECCV}, 2020.

\bibitem{faneccv20}
Deng-Ping Fan, Yingjie Zhai, Ali Borji, Jufeng Yang, and Ling Shao,
\newblock ``Bbs-net: Rgb-d salient object detection with a bifurcated backbone
  strategy network,''
\newblock in {\em Proceedings of the European Conference on Computer Vision}.
  Springer, 2020.

\bibitem{Zhangeccv20}
Miao Zhang, Sun~Xiao Fei, Jie Liu, Shuang Xu, Yongri Piao, and Huchuan Lu,
\newblock ``Asymmetric two-stream architecture for accurate rgb-d saliency
  detection,''
\newblock in {\em Proceedings of the European Conference on Computer Vision}.
  Springer, 2020.

\bibitem{chen2020progressively}
Shuhan Chen and Yun Fu,
\newblock ``Progressively guided alternate refinement network for rgb-d salient
  object detection,''
\newblock in {\em Proceedings of the European Conference on Computer Vision}.
  Springer, 2020.

\bibitem{huang2020multi}
Zhou Huang, Huai-Xin Chen, Tao Zhou, Yun-Zhi Yang, and Chang-Yin Wang,
\newblock ``Multi-level cross-modal interaction network for rgb-d salient
  object detection,''
\newblock {\em arXiv preprint arXiv:2007.14352}, 2020.

\bibitem{wang2020data}
Xuehao Wang, Shuai Li, Chenglizhao Chen, Yuming Fang, Aimin Hao, and Hong Qin,
\newblock ``Data-level recombination and lightweight fusion scheme for rgb-d
  salient object detection,''
\newblock {\em IEEE Transactions on Image Processing}, 2020.

\bibitem{wang2020knowing}
Xuehao Wang, Shuai Li, Chenglizhao Chen, Aimin Hao, and Hong Qin,
\newblock ``Knowing depth quality in advance: A depth quality assessment method
  for rgb-d salient object detection,''
\newblock {\em arXiv preprint arXiv:2008.04157}, 2020.

\bibitem{chenchen2020depth}
Chenglizhao Chen, Jipeng Wei, Chong Peng, and Hong Qin,
\newblock ``Depth quality aware salient object detection,''
\newblock {\em IEEE Transactions on Image Processing}, 2020.

\bibitem{zhao2020}
Jiawei Zhao, Yifan Zhao, Jia Li, and Xiaowu Chen,
\newblock ``Is depth really necessary for salient object detection,''
\newblock in {\em ACM Multimedia}, 2020.

\bibitem{chen2020rgbd}
Hao Chen, Yongjian Deng, Youfu Li, Tzu-Yi Hung, and Guosheng Lin,
\newblock ``Rgbd salient object detection via disentangled cross-modal
  fusion,''
\newblock {\em IEEE Transactions on Image Processing}, vol. 29, pp. 8407--8416,
  2020.

\bibitem{niu2012leveraging}
Yuzhen Niu, Yujie Geng, Xueqing Li, and Feng Liu,
\newblock ``Leveraging stereopsis for saliency analysis,''
\newblock in {\em Proceedings of the IEEE Conference on Computer Vision and
  Pattern Recognition}. IEEE, 2012, pp. 454--461.

\bibitem{li2014saliency}
Nianyi Li, Jinwei Ye, Yu~Ji, Haibin Ling, and Jingyi Yu,
\newblock ``Saliency detection on light field,''
\newblock in {\em Proceedings of the IEEE Conference on Computer Vision and
  Pattern Recognition}, 2014, pp. 2806--2813.

\bibitem{zhangmm2020}
yu~Zhang et~al.,
\newblock ``Feature reintegration over differential treatment: A top-down and
  adaptive fusion network for rgb-d salient object detection,''
\newblock in {\em ACM Multimedia}, 2020.

\bibitem{li2015weighted}
Nianyi Li, Bilin Sun, and Jingyi Yu,
\newblock ``A weighted sparse coding framework for saliency detection,''
\newblock in {\em Proceedings of the IEEE Conference on Computer Vision and
  Pattern Recognition}, 2015, pp. 5216--5223.

\bibitem{zhang2015saliency}
Jun Zhang, Meng Wang, Jun Gao, Yi~Wang, Xudong Zhang, and Xindong Wu,
\newblock ``Saliency detection with a deeper investigation of light field.,''
\newblock in {\em Proceedings of the International Joint Conference on
  Artificial Intelligence}, 2015, pp. 2212--2218.

\bibitem{sheng2016relative}
Hao Sheng, Shuo Zhang, Xiaoyu Liu, and Zhang Xiong,
\newblock ``Relative location for light field saliency detection,''
\newblock in {\em Proceedings of the IEEE Conference on Acoustics, Speech and
  Signal Processing}. IEEE, 2016, pp. 1631--1635.

\bibitem{zhang2017saliency}
Jun Zhang, Meng Wang, Liang Lin, Xun Yang, Jun Gao, and Yong Rui,
\newblock ``Saliency detection on light field: A multi-cue approach,''
\newblock {\em ACM Transactions on Multimedia Computing, Communications, and
  Applications}, vol. 13, no. 3, pp. 1--22, 2017.

\bibitem{wang2017two}
Anzhi Wang, Minghui Wang, Xiaoyan Li, Zetian Mi, and Huan Zhou,
\newblock ``A two-stage bayesian integration framework for salient object
  detection on light field,''
\newblock {\em Neural Processing Letters}, vol. 46, no. 3, pp. 1083--1094,
  2017.

\bibitem{li2017pami}
Nianyi Li, Jinwei Ye, Yu~Ji, Haibing Ling, and Jingyi Yu,
\newblock ``Saliency detection on light field,''
\newblock {\em IEEE Transactions on Pattern Analysis and Machine Intelligence},
  vol. 39, no. 8, pp. 1605--1616, 2017.

\bibitem{li2017saliency}
Chao Li, Bin Zhan, Shuo Zhang, and Hao Sheng,
\newblock ``Saliency detection with relative location measure in light field
  image,''
\newblock in {\em Proceedings of the International Conference on Image, Vision
  and Computing}. IEEE, 2017, pp. 8--12.

\bibitem{wang2018salience}
Shizheng Wang, Wenjuan Liao, Phil Surman, Zhigang Tu, Yuanjin Zheng, and
  Junsong Yuan,
\newblock ``Salience guided depth calibration for perceptually optimized
  compressive light field 3d display,''
\newblock in {\em Proceedings of the IEEE Conference on Computer Vision and
  Pattern Recognition}, 2018, pp. 2031--2040.

\bibitem{piao2018depth}
Yongri Piao, Xiao Li, and Miao Zhang,
\newblock ``Depth-induced cellular automata for light field saliency,''
\newblock in {\em Frontiers in Optics}. Optical Society of America, 2018, pp.
  FTh3E--3.

\bibitem{wang2019deep}
Tiantian Wang, Yongri Piao, Xiao Li, Lihe Zhang, and Huchuan Lu,
\newblock ``Deep learning for light field saliency detection,''
\newblock in {\em Proceedings of the IEEE International Conference on Computer
  Vision}, 2019, pp. 8838--8848.

\bibitem{piao2019deep}
Yongri Piao, Zhengkun Rong, Miao Zhang, Xiao Li, and Huchuan Lu,
\newblock ``Deep light-field-driven saliency detection from a single view,''
\newblock in {\em Proceedings of the International Joint Conference on
  Artificial Intelligence}, 2019.

\bibitem{zhang2019memory}
Miao Zhang, Jingjing Li, JI~WEI, Yongri Piao, and Huchuan Lu,
\newblock ``Memory-oriented decoder for light field salient object detection,''
\newblock in {\em Proceedings of the International Conference on Neural
  Information Processing Systems}, 2019, pp. 896--906.

\bibitem{piaoexploit}
Yongri Piao, Zhengkun Rong, Miao Zhang, and Huchuan Lu,
\newblock ``Exploit and replace: An asymmetrical two-stream architecture for
  versatile light field saliency detection,''
\newblock in {\em Proceedings of the Association for the Advancement of
  Artificial Intelligence}, 2020.

\bibitem{wang2020region}
Xue Wang, Yingying Dong, Qi~Zhang, and Qing Wang,
\newblock ``Region-based depth feature descriptor for saliency detection on
  light field,''
\newblock {\em Multimedia Tools and Applications}, 2020.

\bibitem{zhang2020lfnet}
Miao Zhang, Wei Ji, Yongri Piao, Jingjing Li, Yu~Zhang, Shuang Xu, and Huchuan
  Lu,
\newblock ``Lfnet: Light field fusion network for salient object detection,''
\newblock {\em IEEE Transactions on Image Processing}, vol. 29, pp. 6276--6287,
  2020.

\bibitem{zhang2020light}
Jun Zhang, Yamei Liu, Shengping Zhang, Ronald Poppe, and Meng Wang,
\newblock ``Light field saliency detection with deep convolutional networks,''
\newblock {\em IEEE Transactions on Image Processing}, vol. 29, pp. 4421--4434,
  2020.

\bibitem{achanta2009frequency}
Radhakrishna Achanta, Sheila Hemami, Francisco Estrada, and Sabine Susstrunk,
\newblock ``Frequency-tuned salient region detection,''
\newblock in {\em Proceedings of the IEEE conference on Computer Vision and
  Pattern Recognition}. IEEE, 2009, pp. 1597--1604.

\bibitem{perazzi2012saliency}
Federico Perazzi, Philipp Kr{\"a}henb{\"u}hl, Yael Pritch, and Alexander
  Hornung,
\newblock ``Saliency filters: Contrast based filtering for salient region
  detection,''
\newblock in {\em Proceedings of the IEEE Conference on Computer Vision and
  Pattern Recognition}. IEEE, 2012, pp. 733--740.

\bibitem{fan2017structure}
Deng-Ping Fan, Ming-Ming Cheng, Yun Liu, Tao Li, and Ali Borji,
\newblock ``Structure-measure: A new way to evaluate foreground maps,''
\newblock in {\em Proceedings of the IEEE International Conference on Computer
  Vision}, 2017, pp. 4548--4557.

\bibitem{Fan2018Enhanced}
Deng-Ping Fan, Cheng Gong, Yang Cao, Bo~Ren, Ming-Ming Cheng, and Ali Borji,
\newblock ``Enhanced-alignment measure for binary foreground map evaluation,''
\newblock in {\em Proceedings of the International Joint Conferences on
  Artificial Intelligence}, 2018, pp. 698--704.

\bibitem{qin2015saliency}
Yao Qin, Huchuan Lu, Yiqun Xu, and He~Wang,
\newblock ``Saliency detection via cellular automata,''
\newblock in {\em Proceedings of the IEEE Conference on Computer Vision and
  Pattern Recognition}, 2015, pp. 110--119.

\bibitem{zhou2015salient}
Li~Zhou, Zhaohui Yang, Qing Yuan, Zongtan Zhou, and Dewen Hu,
\newblock ``Salient region detection via integrating diffusion-based
  compactness and local contrast,''
\newblock {\em IEEE Transactions on Image Processing}, vol. 24, no. 11, pp.
  3308--3320, 2015.

\bibitem{huang2017}
Xiaoming Huang and Yu-Jin Zhang,
\newblock ``300-fps salient object detection via minimum directional
  contrast,''
\newblock {\em IEEE Transactions on Image Processing}, vol. 26, no. 9, pp.
  4243--4254, 2017.

\bibitem{huang2017salient}
Fang Huang, Jinqing Qi, Huchuan Lu, Lihe Zhang, and Xiang Ruan,
\newblock ``Salient object detection via multiple instance learning,''
\newblock {\em IEEE Transactions on Image Processing}, vol. 26, no. 4, pp.
  1911--1922, 2017.

\bibitem{huang2018water}
Xiaoming Huang and Yujin Zhang,
\newblock ``Water flow driven salient object detection at 180 fps,''
\newblock {\em Pattern Recognition}, vol. 76, pp. 95--107, 2018.

\bibitem{xu2015multi}
Chang Xu, Dacheng Tao, and Chao Xu,
\newblock ``Multi-view learning with incomplete views,''
\newblock {\em IEEE Transactions on Image Processing}, vol. 24, no. 12, pp.
  5812--5825, 2015.

\bibitem{zhou2019effective}
Tao Zhou, Kim-Han Thung, Xiaofeng Zhu, and Dinggang Shen,
\newblock ``Effective feature learning and fusion of multimodality data using
  stage-wise deep neural network for dementia diagnosis,''
\newblock {\em Human Brain Mapping}, vol. 40, no. 3, pp. 1001--1016, 2019.

\bibitem{zhou2019latent}
Tao Zhou, Mingxia Liu, Kim-Han Thung, and Dinggang Shen,
\newblock ``Latent representation learning for alzheimer’s disease diagnosis
  with incomplete multi-modality neuroimaging and genetic data,''
\newblock {\em IEEE Transactions on Medical Imaging}, vol. 38, no. 10, pp.
  2411--2422, 2019.

\bibitem{zhou2020multi}
Tao Zhou, Kim-Han Thung, Mingxia Liu, Feng Shi, Changqing Zhang, and Dinggang
  Shen,
\newblock ``Multi-modal latent space inducing ensemble svm classifier for early
  dementia diagnosis with neuroimaging data,''
\newblock {\em Medical Image Analysis}, vol. 60, pp. 101630, 2020.

\bibitem{zhou2020hi}
Tao Zhou, Huazhu Fu, Geng Chen, Jianbing Shen, and Ling Shao,
\newblock ``Hi-net: hybrid-fusion network for multi-modal {MR} image
  synthesis,''
\newblock {\em IEEE Transactions on Medical Imaging}, vol. 39, no. 9, pp.
  2772--2781, 2020.

\bibitem{godard2017unsupervised}
Cl{\'e}ment Godard, Oisin Mac~Aodha, and Gabriel~J Brostow,
\newblock ``Unsupervised monocular depth estimation with left-right
  consistency,''
\newblock in {\em Proceedings of the IEEE Conference on Computer Vision and
  Pattern Recognition}, 2017, pp. 270--279.

\bibitem{liu2015deep}
Fayao Liu, Chunhua Shen, and Guosheng Lin,
\newblock ``Deep convolutional neural fields for depth estimation from a single
  image,''
\newblock in {\em Proceedings of the IEEE Conference on Computer Vision and
  Pattern Recognition}, 2015, pp. 5162--5170.

\bibitem{wang2020sdc}
Lijun Wang, Jianming Zhang, Oliver Wang, Zhe Lin, and Huchuan Lu,
\newblock ``Sdc-depth: Semantic divide-and-conquer network for monocular depth
  estimation,''
\newblock in {\em Proceedings of the IEEE Conference on Computer Vision and
  Pattern Recognition}, 2020, pp. 541--550.

\bibitem{jin2020geometric}
Lei Jin, Yanyu Xu, Jia Zheng, Junfei Zhang, Rui Tang, Shugong Xu, Jingyi Yu,
  and Shenghua Gao,
\newblock ``Geometric structure based and regularized depth estimation from 360
  indoor imagery,''
\newblock in {\em Proceedings of the IEEE Conference on Computer Vision and
  Pattern Recognition}, 2020, pp. 889--898.

\bibitem{mirza2014conditional}
Mehdi Mirza and Simon Osindero,
\newblock ``Conditional generative adversarial nets,''
\newblock {\em arXiv preprint arXiv:1411.1784}, 2014.

\bibitem{zhu2018multi}
Dandan Zhu, Lei Dai, Ye~Luo, Guokai Zhang, Xuan Shao, Laurent Itti, and Jianwei
  Lu,
\newblock ``Multi-scale adversarial feature learning for saliency detection,''
\newblock {\em Symmetry}, vol. 10, no. 10, pp. 457, 2018.

\bibitem{pan2017salgan}
Junting Pan, Cristian~Canton Ferrer, Kevin McGuinness, et~al.,
\newblock ``Salgan: Visual saliency prediction with generative adversarial
  networks,''
\newblock {\em arXiv preprint arXiv:1701.01081}, 2017.

\bibitem{vaswani2017attention}
Ashish Vaswani, Noam Shazeer, Niki Parmar, Jakob Uszkoreit, Llion Jones,
  Aidan~N Gomez, {\L}ukasz Kaiser, and Illia Polosukhin,
\newblock ``Attention is all you need,''
\newblock in {\em Proceedings of the Conference on Neural Information
  Processing Systems}, 2017, pp. 5998--6008.

\bibitem{wang2017residual}
Fei Wang, Mengqing Jiang, Chen Qian, Shuo Yang, Cheng Li, Honggang Zhang,
  Xiaogang Wang, and Xiaoou Tang,
\newblock ``Residual attention network for image classification,''
\newblock in {\em Proceedings of the IEEE Conference on Computer Vision and
  Pattern Recognition}. Springer, 2017, pp. 3156--3164.

\bibitem{fang2018pairwise}
Hao-Shu Fang, Jinkun Cao, Yu-Wing Tai, and Cewu Lu,
\newblock ``Pairwise body-part attention for recognizing human-object
  interactions,''
\newblock in {\em Proceedings of the European Conference on Computer Vision}.
  Springer, 2018, pp. 51--67.

\bibitem{wang2017deep}
Wenguan Wang and Jianbing Shen,
\newblock ``Deep visual attention prediction,''
\newblock {\em IEEE Transactions on Image Processing}, vol. 27, no. 5, pp.
  2368--2378, 2017.

\bibitem{lu2016hierarchical}
Jiasen Lu, Jianwei Yang, Dhruv Batra, and Devi Parikh,
\newblock ``Hierarchical question-image co-attention for visual question
  answering,''
\newblock in {\em Proceedings of the International Conference on Neural
  Information Processing Systems}, 2016, pp. 289--297.

\bibitem{yu2019deep}
Zhou Yu, Jun Yu, Yuhao Cui, Dacheng Tao, and Qi~Tian,
\newblock ``Deep modular co-attention networks for visual question answering,''
\newblock in {\em Proceedings of the IEEE Conference on Computer Vision and
  Pattern Recognition}, 2019, pp. 6281--6290.

\bibitem{lu2019see}
Xiankai Lu, Wenguan Wang, Chao Ma, Jianbing Shen, Ling Shao, and Fatih Porikli,
\newblock ``See more, know more: Unsupervised video object segmentation with
  co-attention siamese networks,''
\newblock in {\em Proceedings of the IEEE Conference on Computer Vision and
  Pattern Recognition}, 2019, pp. 3623--3632.

\bibitem{zeng2019multi}
Yu~Zeng, Yunzhi Zhuge, Huchuan Lu, Lihe Zhang, Mingyang Qian, and Yizhou Yu,
\newblock ``Multi-source weak supervision for saliency detection,''
\newblock in {\em Proceedings of the IEEE Conference on Computer Vision and
  Pattern Recognition}, 2019, pp. 6074--6083.

\bibitem{zhang2017bridging}
Dingwen Zhang, Deyu Meng, Long Zhao, and Junwei Han,
\newblock ``Bridging saliency detection to weakly supervised object detection
  based on self-paced curriculum learning,''
\newblock {\em arXiv preprint arXiv:1703.01290}, 2017.

\bibitem{qian2019language}
Mingyang Qian, Jinqing Qi, Lihe Zhang, Mengyang Feng, and Huchuan Lu,
\newblock ``Language-aware weak supervision for salient object detection,''
\newblock {\em Pattern Recognition}, vol. 96, pp. 106955, 2019.

\bibitem{yan2019semi}
Pengxiang Yan, Guanbin Li, Yuan Xie, Zhen Li, Chuan Wang, Tianshui Chen, and
  Liang Lin,
\newblock ``Semi-supervised video salient object detection using
  pseudo-labels,''
\newblock in {\em Proceedings of the IEEE International Conference on Computer
  Vision}, 2019, pp. 7284--7293.

\bibitem{zhou2018semi}
Yuan Zhou, Shuwei Huo, Wei Xiang, Chunping Hou, and Sun-Yuan Kung,
\newblock ``Semi-supervised salient object detection using a linear feedback
  control system model,''
\newblock {\em IEEE Transactions on Cybernetics}, vol. 49, no. 4, pp.
  1173--1185, 2018.

\bibitem{zhang2017supervision}
Dingwen Zhang, Junwei Han, and Yu~Zhang,
\newblock ``Supervision by fusion: Towards unsupervised learning of deep
  salient object detector,''
\newblock in {\em Proceedings of the IEEE International Conference on Computer
  Vision}, 2017, pp. 4048--4056.

\bibitem{chen2020adversarial}
Tianlong Chen, Sijia Liu, Shiyu Chang, Yu~Cheng, Lisa Amini, and Zhangyang
  Wang,
\newblock ``Adversarial robustness: From self-supervised pre-training to
  fine-tuning,''
\newblock in {\em Proceedings of the IEEE Conference on Computer Vision and
  Pattern Recognition}, 2020, pp. 699--708.

\bibitem{dai2020sg}
Angela Dai, Christian Diller, and Matthias Nie{\ss}ner,
\newblock ``Sg-nn: Sparse generative neural networks for self-supervised scene
  completion of rgb-d scans,''
\newblock in {\em Proceedings of the IEEE Conference on Computer Vision and
  Pattern Recognition}, 2020, pp. 849--858.

\bibitem{lai2011large}
Kevin Lai, Liefeng Bo, Xiaofeng Ren, and Dieter Fox,
\newblock ``A large-scale hierarchical multi-view rgb-d object dataset,''
\newblock in {\em Proceedings of the IEEE International Conference on Robotics
  and Automation}. IEEE, 2011, pp. 1817--1824.

\bibitem{zhang2016large}
Jing Zhang, Wanqing Li, Pichao Wang, Philip Ogunbona, Song Liu, and Chang Tang,
\newblock ``A large scale rgb-d dataset for action recognition,''
\newblock in {\em Proceedings of the International Workshop on Understanding
  Human Activities through 3D Sensors}. Springer, 2016, pp. 101--114.

\bibitem{he2018amc}
Yihui He, Ji~Lin, Zhijian Liu, Hanrui Wang, Li-Jia Li, and Song Han,
\newblock ``Amc: Automl for model compression and acceleration on mobile
  devices,''
\newblock in {\em Proceedings of the European Conference on Computer Vision}.
  Springer, 2018, pp. 784--800.

\bibitem{cheng2017survey}
Yu~Cheng, Duo Wang, Pan Zhou, and Tao Zhang,
\newblock ``A survey of model compression and acceleration for deep neural
  networks,''
\newblock {\em arXiv preprint arXiv:1710.09282}, 2017.

\bibitem{ma2017learning}
Yunpeng Ma, Dengdi Sun, Qianqian Meng, Zhuanlian Ding, and Chenglong Li,
\newblock ``Learning multiscale deep features and svm regressors for adaptive
  rgb-t saliency detection,''
\newblock in {\em Proceedings of the International Symposium on Computational
  Intelligence and Design}. IEEE, 2017, vol.~1, pp. 389--392.

\bibitem{li2017unified}
Chenglong Li, Guizhao Wang, Yunpeng Ma, Aihua Zheng, Bin Luo, and Jin Tang,
\newblock ``A unified rgb-t saliency detection benchmark: dataset, baselines,
  analysis and a novel approach,''
\newblock {\em arXiv preprint arXiv:1701.02829}, 2017.

\bibitem{wang2018rgb}
Guizhao Wang, Chenglong Li, Yunpeng Ma, Aihua Zheng, Jin Tang, and Bin Luo,
\newblock ``{Rgb-T} saliency detection benchmark: Dataset, baselines, analysis
  and a novel approach,''
\newblock in {\em Proceedings of the Chinese Conference on Image and Graphics
  Technologies}. Springer, 2018, pp. 359--369.

\bibitem{sun2019rgb}
Dengdi Sun, Sheng Li, Zhuanlian Ding, and Bin Luo,
\newblock ``Rgb-t saliency detection via robust graph learning and
  collaborative manifold ranking,''
\newblock in {\em Proceedings of the International Conference on Bio-Inspired
  Computing: Theories and Applications}. Springer, 2019, pp. 670--684.

\bibitem{tu2019m3s}
Zhengzheng Tu, Tian Xia, Chenglong Li, Yijuan Lu, and Jin Tang,
\newblock ``M3s-nir: Multi-modal multi-scale noise-insensitive ranking for
  rgb-t saliency detection,''
\newblock in {\em Proceedings of the IEEE Conference on Multimedia Information
  Processing and Retrieval}. IEEE, 2019, pp. 141--146.

\bibitem{tu2020multi}
Zhengzheng Tu, Zhun Li, Chenglong Li, Yang Lang, and Jin Tang,
\newblock ``Multi-interactive encoder-decoder network for rgbt salient object
  detection,''
\newblock {\em arXiv preprint arXiv:2005.02315}, 2020.

\bibitem{tu2019rgb}
Zhengzheng Tu, Tian Xia, Chenglong Li, Xiaoxiao Wang, Yan Ma, and Jin Tang,
\newblock ``Rgb-t image saliency detection via collaborative graph learning,''
\newblock {\em IEEE Transactions on Multimedia}, vol. 22, no. 1, pp. 160--173,
  2019.

\bibitem{zhang2019rgb}
Qiang Zhang, Nianchang Huang, Lin Yao, Dingwen Zhang, Caifeng Shan, and Jungong
  Han,
\newblock ``Rgb-t salient object detection via fusing multi-level cnn
  features,''
\newblock {\em IEEE Transactions on Image Processing}, vol. 29, pp. 3321--3335,
  2019.

\bibitem{tu2020rgbt}
Zhengzheng Tu, Yan Ma, Zhun Li, Chenglong Li, Jieming Xu, and Yongtao Liu,
\newblock ``Rgbt salient object detection: A large-scale dataset and
  benchmark,''
\newblock {\em arXiv preprint arXiv:2007.03262}, 2020.

\end{thebibliography}

\end{document}